\documentclass[letterpaper, 10 pt, conference]{ieeeconf}
\IEEEoverridecommandlockouts                

\usepackage[table, x11names]{xcolor}
\usepackage[colorlinks=true,linkcolor=blue,citecolor=blue,urlcolor=blue,pagebackref]{hyperref}
\usepackage{times} 
\usepackage{amsmath} 
\usepackage{amssymb}  
\usepackage{graphicx}
\usepackage{algorithm}
\usepackage{mathtools}
\usepackage{algorithm}
\usepackage{algorithmicx}
\usepackage[noend]{algpseudocode}
\usepackage[font=footnotesize]{subcaption}

\def\secref#1{Sec.~\ref{#1}}
\def\figref#1{Fig.~\ref{#1}}
\def\tabref#1{Tab.~\ref{#1}}
\def\eqref#1{Eq.~(\ref{#1})}
\def\algref#1{Alg.~\ref{#1}}

\usepackage{balance}

\usepackage{multirow} 

\usepackage{cite}

\usepackage{glossaries-extra}
\setabbreviationstyle[acronym]{long-short}
\glssetcategoryattribute{acronym}{nohyperfirst}{true}

\makeglossaries

\newacronym{slam}{SLAM}{Simultaneous Localization and Mapping}
\newacronym{sfm}{SfM}{Structure from Motion}
\newacronym{ba}{BA}{Bundle Adjustment}
\newacronym{pgo}{PGO}{Pose-Graph Optimization}
\newacronym{plgo}{PLGO}{Pose-Landmark Graph Optimization}
\newacronym{icp}{ICP}{Iterative Closest Point}
\newacronym{vo}{VO}{Visual Odometry}
\newacronym{vio}{VIO}{Visual-Inertial Odometry}
\newacronym{vpr}{VPR}{Visual Place Recognition}
\newacronym{sgd}{SGD}{Stochastic Gradient Descent}
\newacronym{ls}{LS}{Least-Squares}
\newacronym{ils}{ILS}{Iterative Least-Squares}
\newacronym{irls}{IRLS}{Iterative Reweighed Least-Squares}
\newacronym{gn}{GN}{Gauss-Newton}
\newacronym{lm}{LM}{Levenberg-Marquardt}
\newacronym{trm}{TRM}{Trusted-Region Method}
\newacronym{pcg}{PCG}{Preconditioned Conjugate Gradient}
\newacronym{map}{MAP}{Maximum-A-Posteriori}
\newacronym{ad}{AD}{Automatic Differentiation}
\newacronym{ndt}{NDT}{Normal Distributed Transform}
\newacronym{boss}{\texttt{BOSS}}{\texttt{Basic Object Serialization System}}
\newacronym{gf}{GF}{Gaussian Filters}
\newacronym{pf}{PF}{Particle Filters}
\newacronym{sdp}{SDP}{Semi-Definite Programming}
\newacronym{awgn}{AWGN}{Additive White Gaussian Noise}
\newacronym{ate}{ATE}{Absolute Trajectory Error}
\newacronym{dof}{DoF}{Degrees-of-Freedom}

\def\ie{\emph{i.e.}}
\def\eg{\emph{e.g.}}
\def\etal{\emph{et al.}}

\newcommand{\bv}{\mathbf{v}}

\newcommand{\bt}{\mathbf{t}}
\newcommand{\bM}{\mathbf{M}}
\newcommand{\bA}{\mathbf{A}}

\newcommand{\bK}{\mathbf{K}}

\newcommand{\bH}{\mathbf{H}}
\newcommand{\bI}{\mathbf{I}}
\newcommand{\bX}{\mathbf{X}}
\newcommand{\bZ}{\mathbf{Z}}
\newcommand{\bR}{\mathbf{R}}

\newcommand{\bT}{\mathbf{T}}

\newcommand{\bJ}{\mathbf{J}}
\newcommand{\bZero}{\mathbf{0}}

\newcommand{\cN}{\mathcal{N}}

\newcommand{\bb}{\mathbf{b}}

\newcommand{\be}{\mathbf{e}}

\newcommand{\bx}{\mathbf{x}}

\newcommand{\bz}{\mathbf{z}}

\newcommand{\bh}{\mathbf{h}}

\newcommand{\bp}{\mathbf{p}}
\newcommand{\bDelta}{\mathbf{\Delta}}

\newcommand{\bDeltar}{\mathbf{\Delta r}}
\newcommand{\bDeltam}{\mathbf{\Delta m}}

\newcommand{\bDeltax}{\mathbf{\Delta x}}
\newcommand{\bDeltaa}{\mathbf{\Delta a}}
\newcommand{\bDeltaz}{\mathbf{\Delta z}}
\newcommand{\bDeltaX}{\mathbf{\Delta X}}
\newcommand{\bDeltat}{\mathbf{\Delta t}}

\newcommand{\bDeltaZ}{\mathbf{\Delta Z}}
\newcommand{\bzero}{\mathbf{0}}
\newcommand{\tTov}{\mathrm{t2v}}
\newcommand{\vTot}{\mathrm{v2t}}

\newcommand{\bmu}{\boldsymbol{\mu}}
\newcommand{\bnu}{\boldsymbol{\nu}}
\newcommand{\bSigma}{\mathbf{\Sigma}}
\newcommand{\bOmega}{\mathbf{\Omega}}

\newcommand{\eq}{=}

\newcommand{\manifoldchart}{\mathrm{chart}}

\DeclareMathOperator*{\argmin}{argmin}

\newcommand{\Fnew}{F_\mathrm{new}}
\newcommand{\Fold}{F_\mathrm{old}}
\newcommand{\Finternal}{F_\mathrm{internal}}

\def\g2o{$g^2o$}
\def\se3{\mathrm{SE}(3)}
\def\t2v{\mathrm{t2v}}
\def\v2t{\mathrm{v2t}}

\def\fromvec{\mathrm{fromVector}}
\def\tovec{\mathrm{toVector}}
\def\chisquared{$\chi^2$~}
\def\sqrtsam{$\sqrt{\text{SAM}}$~}

\renewcommand{\bmu}{\mu}
\renewcommand{\bnu}{\nu}
\newcommand{\setalglinespace}{}
\newcommand{\MyEndFor}{\EndFor}
\newcommand{\MyEndWhile}{\EndWhile}
\newcommand{\MyEndIf}{\EndIf}

\def\g2o{$g^2o$}
\def\t2v{\mathrm{t2v}}
\def\v2t{\mathrm{v2t}}

\def\sqrtsam{$\sqrt{\text{SAM}}$~}

\def\myParagraph#1{\paragraph{\textsc{#1}}}

\usepackage{listings}
\definecolor{sbase03}{HTML}{002B36}
\definecolor{sbase02}{HTML}{073642}
\definecolor{sbase01}{HTML}{586E75}
\definecolor{sbase00}{HTML}{657B83}
\definecolor{sbase0}{HTML}{839496}
\definecolor{sbase1}{HTML}{93A1A1}
\definecolor{sbase2}{HTML}{EEE8D5}
\definecolor{sbase3}{HTML}{FDF6E3}
\definecolor{syellow}{HTML}{B58900}
\definecolor{sorange}{HTML}{CB4B16}
\definecolor{sred}{HTML}{DC322F}
\definecolor{smagenta}{HTML}{D33682}
\definecolor{sviolet}{HTML}{6C71C4}
\definecolor{sblue}{HTML}{268BD2}
\definecolor{scyan}{HTML}{2AA198}
\definecolor{sgreen}{HTML}{859900}
\title{\LARGE \bf Least Squares Optimization: from Theory to Practice}

\author{
  Giorgio Grisetti~$^{1}$ \hspace{1.7cm}%
  Tiziano Guadagnino~$^{1}$ \hspace{1.7cm}%
  Irvin Aloise~$^{1}$ \hspace{1.7cm}%
  Mirco Colosi~$^{1,2}$ \\%
  Bartolomeo Della Corte~$^{1}$ \hspace{5cm}%
  Dominik Schlegel~$^{1}$
  \thanks{
    $^{1}$~Department of Computer, Control, and Management Engineering ''Antonio Ruberti'', Sapienza University of Rome, Rome, Italy. 
    Email: {\texttt{\{grisetti, guadagnino, ialoise, colosi, dellacorte, schlegel\}@diag.uniroma1.it}}
  }%
  \thanks{$^{2}$~Robot Navigation and Perception (CR/AER1), Robert Bosch 
    Corporate Research, Stuttgart, Germany. Email: {\texttt{Mirco.Colosi@de.bosch.com}}
  }%
}

\begin{document}
\maketitle
\thispagestyle{empty}
\pagestyle{empty}

\begin{abstract}
  Nowadays, Non-Linear Least-Squares embodies the foundation
  of many Robotics and Computer Vision systems. The research community
  deeply investigated this topic in the last years, and this resulted
  in the development of several open-source solvers to approach
  constantly increasing classes of problems.  In this work, we propose
  a unified \emph{methodology} to design and develop efficient
  Least-Squares Optimization algorithms, focusing on the structures
  and patterns of each specific domain. Furthermore, we present a
  novel open-source \emph{optimization system}, that addresses
  transparently problems with a different structure and designed to be
  easy to extend.  The system is written in modern C++ and can run
  efficiently on embedded systems\footnote[3]{Source code \url{http://srrg.gitlab.io/srrg2-solver.html}}.
  We validated our approach by
  conducting comparative experiments on several problems using tandard
  datasets.  The results show that our system achieves
  state-of-the-art performances in all tested scenarios.
\end{abstract}

\section{Introduction}
\label{sec:intro}
\gls{ils} solvers are core building blocks of many robotic
applications, systems and subsystems~\cite{grisetti2010tutorial}. This
technique has been traditionally used for
calibration~\cite{kummerle2011simultaneous,censi2013simultaneous,dellacorte2019unified},
registration~\cite{newcombe2011kinectfusion,pomerleau2013comparing,serafin2017nicp}
and global optimization
\cite{kummerle2011g,dellaert2012gtsam,ceres-solver,ila2017fast}.  In
particular, modern \gls{slam} systems typically employ multiple
\gls{ils} solvers at different levels: in computing the incremental
ego motion of the sensor, in refining the localization of a robot upon
loop closure and - most notably - to obtain a globally consistent map.
Similarly, in several computer vision systems, \gls{ils} is used to
compute/refine camera parameters, estimating the structure of a scene,
the position of the camera or both.  Many inference problems in
robotics are effectively described by a factor
graph~\cite{kaess2017fg}, which is a graphical model expressing the
joint likelihood of the \emph{known} measurements with
respect to a set of unknown conditional variables.  Solving a factor
graph requires to find the values of the variables that maximize the
joint likelihood of the measurements.  If the noise affecting the
sensor data is Gaussian the solution of a factor graph can be computed
by an \gls{ils} solver implementing variants of the well known
\gls{gn} algorithm.

The relevance of the topic has been addressed by several works such as
GTSAM~\cite{dellaert2012gtsam}, \g2o~\cite{kummerle2011g},
SLAM++\cite{ila2017slam++}, or the Ceres solver~\cite{ceres-solver} by
Google.  These systems have grown over time to include comprehensive
libraries of factors and variables, that can tackle a large variety of
problems, and in most cases these systems can be used as black boxes.
Since they typically consist of an extended codebase, entailing them
to a particular application/architecture to achieve the maximum
performances is a non-trivial task.  In contrast, extending these
systems to approach new problems is typically easier than customizing:
in this case the developer has to implement some additional
functionalities/classes according to the API of the system.  Also in
this case, however, an optimal implementation might require a
reasonable knowledge of the solver internals.

We believe that at the current times, a researcher
working in robotics should possess the knowledge on how to design
factor graph solvers for specific problems.  Having this skill enables
to both effectively extend existing systems and realize custom
software that utilizes the hardware at its maximum.  Accordingly, the
primary goal of this paper is to provide the reader with a methodology
on how to mathematically define such a solver for a problem.  To this
extent in Sec.~\ref{sec:least-squares}, we start by revising the
nonlinear least squares by highlighting the connections between
inference on conditional Gaussian distributions and \gls{ils}.  In the
same section, we introduce the $\boxplus$ method introduced by
Hertzberg~\emph{et al.}~\cite{hertzberg2013integrating} to deal with
non-Euclidean domains.  Furthermore, we discuss on how to cope with
outliers in the measurements through robust cost functions and we
outline the effects of sparsity in factor graphs.  We conclude the
section by presenting a general methodology on how to design factors
and variables that describe a problem.  In
Sec.~\ref{sec:use-cases-examples} we bake up this methodology by
providing examples that approach four prominent problems in Robotics:
\gls{icp}, projective registration, \gls{ba}, \gls{pgo}.

When it comes to the implementation of a solver, several choices have
to be made in the light of the problem structure, the compute
architecture and the operating conditions (on-line or batch).  In this
work we characterize \gls{ils} problems, distinguishing between 
\emph{dense} and \emph{sparse}, \emph{batch} and \emph{incremental},
\emph{stationary} and \emph{non-stationary} based on their structure
and application domain.  In Sec.~\ref{sec:taxonomy} we provide a more detailed
description of these characteristics, while in Sec~\ref{sec:related}
we discuss how \gls{ils} has been used in the literature to approach
various problems in Robotics and by highlighting how addressing a
problem according to its traits leads to effective solutions.

\begin{figure*}[!t]
	\centering
	\begin{subfigure}{0.9\columnwidth}
	 	\centering
	 	\includegraphics[width=\columnwidth]{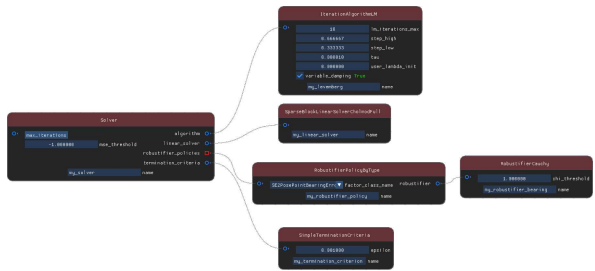}
	 	\subcaption{Graphical solver configurator.}
	 	\label{fig:motivational-config}
	\end{subfigure}
	\begin{subfigure}{0.9\columnwidth}
	 	\centering
		\includegraphics[width=\columnwidth]{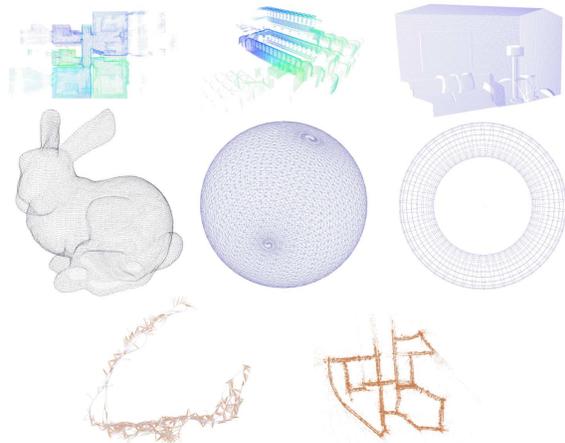}
		\subcaption{Datasets used in the evaluation.}
	 	\label{fig:motivational-datasets}
	\end{subfigure}
	\caption{Left: visual configuration manager implemented in our framework. 
		Each block represents a configurable sub-module. Right: dataset used in the 
		evaluation - dense and sparse.}
	\label{fig:motivational}
\end{figure*}

The second \emph{orthogonal} goal of this work is to propose a
\emph{unifying system that deals with dense/sparse, static/dynamic,
  batch problems}, with no apparent performance loss compared to
ad-hoc solutions.  We build on the ideas that are at the base of the
\g2o optimizer~\cite{kummerle2011g}, to address some requirements
arising from users and developers, namely: fast convergence, small
runtime per iteration, rapid prototyping, trade-off between
implementation effort and performances, and, finally, code
compactness.  In~\secref{sec:solver-design} we highlight from the
general algorithm outlined~\secref{sec:least-squares}, a set of
functionalities that results in a modular, decoupled and minimal
design.  This analysis ultimately leads to a modern compact and
efficient C++ library released under BSD3 license for \gls{ils} of
Factor Graphs that relies on a component model, presented in
Sec~\ref{sec:solver-architecture} that effectively runs on both on
\texttt{x86-64} and \texttt{ARM} platforms.
To ease prototyping we offer an interactive environment to graphically
configure the solver (Fig.~\ref{fig:motivational-config}).  The core
library of our solver consists of no more than 6000 lines of C++ code,
whereas the companion libraries implementing a large set of factors
and variables for approaching problems -\eg~2D/3D \gls{icp},
projective registration, \gls{ba}, 2D/3D \gls{pgo} and \gls{plgo} and
many others - is at the time of this writing below 4000 lines.  Our
system relies on our visual component framework, image processing and
visualization libraries that contain no optimization code and consists
of approximately 20000 lines.  To validate our claims, we conducted
extensive comparative experiments on publicly avalilable datasets
(Fig.~\ref{fig:motivational-datasets}) - in dense and sparse
scenarios.  We compared our solver with sparse approaches such as
GTSAM, \g2o and Ceres, and with dense ones, such as the well-known PCL
library~\cite{rusu2011pcl}.  The experiments presented in
Sec.~\ref{sec:experiments} confirm that our system has performances
that are on par with other state-of-the-art frameworks.
\vspace{5pt}
Summarizing, the contribution of this work is twofold:
\begin{itemize}
  \item[--] We present a methodology on how to design a solver for a generic
    class of problems, and we exemplify such a methodology by showing
    how it can be used to approach a relevant subset of problems in
    Robotics.
  \item[--] We propose an open-source, component-based \gls{ils} system
    that aims to coherently address problems having different
    structure, while providing state-of-the-art performances.
\end{itemize}

\section{Taxonomy of \gls{ils} problems} \label{sec:taxonomy}
Whereas the theory on \gls{ils} is well-known, the effectiveness of an
implementation greatly depends on the structure of the problem being
addressed and on the operating conditions.  We qualitatively
distinguish between \emph{dense} and \emph{sparse} problems, by
discriminating on the connectivity of the factor graph. A dense
problem is characterized by many measurements affected by relatively
few variables.  This occurs in typical registration problems, where
the likelihoods of the measurements (\eg~the intensities of image
pixels) depend on a single variable expressing the sensor position.
In contrast, sparse problems are characterized by measurements that
depend only on a small subset of variables. Examples of sparse
problems include \gls{pgo} or \gls{ba}.

A further orthogonal classification of the problems divides them in
\emph{stationary} and \emph{non-stationary}.  A problem is stationary when the
measurements do not change during the iterations of the optimization.  This occurs when the
data-association is known a priori with sufficient certainty.
Conversely, non-stationary problems admit measurements that might 
change during the optimization, as a result of a modification of the variables 
being estimated. A typical case of non-stationary problem is point
registration~\cite{besl92icp}, when the associations between the points
in the model and in the reference are computed at each iteration based
on an heuristic that depends on the current estimate of their displacement.

Finally, the problem might be extended over time by adding new
variables and measurements.  
Several Graph-Based \gls{slam} systems exploit this intrinsic characteristic in 
on-line applications, reusing the computation done while
solving the original problem to determine the solution for the
augmented one.
We refer to a solver with this capability as
an \emph{incremental} solver, in contrast to \emph{batch} solvers that
carry on all the computation from scratch once the factor graph is
augmented.

In this taxonomy we left out other crucial aspects that affect the
convergence basin of the solver such as the linearity of the
measurement function, or the domain of variables and
measurements. Exploiting the structure of these domains has shown to
provide even more efficient solutions~\cite{rosen2016sesync}, with the
obvious shortcoming that they are restricted to the specific problem
they are designed to address.

Using a sparse stationary solver on a dense non-stationary problem
results in carrying on useless computations that hinders the usability
of the system.  Using a dense dynamic solver to approach a sparse
stationary problem presents similar issues.  State-of-the-art
open-source solvers like the ones mentioned in Sec~\ref{sec:intro}
focus on sparse stationary or incremental problems.  Dense solvers are
usually within the application/library using it and tightly coupled to
it.  On the one hand, this allows to reduce the time per iteration,
while on the other hand it results in avoidable code replication when
multiple systems are integrated.  This might result in potential
inconsistencies among program parts and consequent bugs.

\section{Related Work}
\label{sec:related}
In this section, we revise the use of \gls{ils} in approaching several
problems in robotics, to highlight the structure and the peculiarities
that each problem presents to the solver according to the taxonomy
presented in Sec.~\ref{sec:taxonomy}.  Furthermore, we provide an
overview of generic sparse solvers that are commonly used nowadays
for factor graph optimization.

\subsection{\gls{ils} in Robotics}
In calibration, \gls{ils} has been used extensively since the
first works appeared until these
days~\cite{zhang2000calibration,censi08calib,dicicco2016calib}.
Common works in batch calibration involve relatively small state
spaces covering only the parameters to be estimated.  Since these
parameters condition directly or indirectly all measurements, these
class of problems typically requires a dense stationary solver.  When
temporal calibration is required, however, the changing time offset
might result in considering different data chunks at each
iterations thus requiring a dense, non-stationary solver, such as
the one presented in~\cite{dellacorte2019unified}.

Among the first works on pairwise shape registration relying on
\gls{ils}, we find the \gls{icp} proposed by Besl and
McKay~\cite{besl92icp}, while Chen and Medioni~\cite{chen91icra}
proposed the first \gls{ils} method for the incremental reconstruction
of a 3D model from multiple range images.  These methods constitute
the foundation of many registration algorithms appearing during the
subsequent years.  In particular Lu and Milios~\cite{lu1997icp}
specialized \gls{icp} to operate on 2D laser scans. All these works
employed dense non-stationary solvers, to estimate the robot pose that
better explain the point measurements. The non-stationary aspect
arises from the heuristic used to estimate the data association is based
on the current pose estimate.

In the context of \gls{icp}, Censi~\cite{censi2008icp} proposed an
alternative metric to compute the distance between two points and an
approach to estimate the information matrix from the set of
correspondences~\cite{censi2007accurate}. Subsequently,
Segal~\etal~\cite{segal2009rss} proposed the use of covariance matrices
that better reflect the structure \\ of the surface in computing the
error.
Registration has been addressed by Bieber~\etal~\cite{biber2003ndt}
for 2D scans and subsequently Magnusson~\etal~\cite{magnusson2007ndt}
for 3D point clouds by using a pure Newton's method relying on a
Gaussian approximation of the point clouds to be registered, called
\gls{ndt}. Serafin~\emph{et. al}~\cite{serafin2017nicp} approached the
problem of point cloud registration using a 6D error function encoding
also the normal difference in the error vector. All the approaches
mentioned so far leverage on a dense \gls{ils} solver, with the
notable exception of \gls{ndt} that is a second-order approach that
specializes the Newton's algorithm.

In the context of Computer Vision, the p2p
algorithm~\cite{wolf2000elements, fischler1981random} allows to find
the camera transformation that minimizes the reprojection error
between a set of 3D points in the scene and their 2D projections in
the image. The first stage of p2p is usually conducted according to a
consensus scheme that relies on an ad-hoc minimal solver requiring
only 3 correspondences.  The final stage, however typically uses a
dense and stationary \gls{ils} approach, since the correspondences do
not change during the iterations.  When the initial guess of the
camera is known with sufficient accuracy, like in \gls{vo}, only the
latter stage is typically used.  In contrast to these feature-based
solvers, Engels~\emph{et. al}~\cite{engel2013semi} approach \gls{vo}
by minimizing the reprojection error between two images through dense
and non-stationary \gls{ils}.  Using this method requires the system
to possess a reasonably good estimate of the depth for a subset of the
point in the scene. Such initialization is usually obtained by
estimating the transformation between two images using a combination
of RANSAC and direct solvers, and then computing the depth through
triangulation between the stereo pair.  Della
Corte~\emph{et. al}~\cite{dellacorte2018mpr} developed a registration
algorithm called MPR, which was built on this idea. As a result, MPR
is able to operate on depth images capturing different cues and
obtained with arbitrary projection functions.  To operate on-line, all
the registration works mentioned so far rely on ad-hoc dense and
non-stationary \gls{ils} solvers that leverage on the specific
problem's structure to reduce the computation.

The scan based \gls{icp} algorithm~\cite{lu1997icp} has been
subsequently employed by the same authors~\cite{lu1997globally} as a
building block for a system that estimates a globally consistent map.
The core is to determine the relative transforms between pairwise
scans through \gls{icp}. These transformations are known as
\emph{constraints}, and a global map is obtained by finding the
position of all the scans that better explain the constraints. The
process can be visualized as a graph, whose nodes are the scan
positions and whose edges are the constraints, hence this problem is
called \gls{pgo}. Constraints can exist only between spatially close
nodes, due to the limited sensor range. Hence, \gls{pgo} is inherently
sparse.  Additionally, in the on-line case the graph is incrementally
augmented as new measurements become available, rendering it
incremental.  We are unaware on these two aspects being
exploited in the design of the underlying solver
in~\cite{lu1997globally}.  The work of Lu and Milios inspired
Borrman~\etal~\cite{borrmann2008globally} to produce an effective 3D
extension.

For several years after the introduction of Graph-Based
SLAM~\cite{lu1997globally}, the community put aside \gls{ils}
approaches in favor of filtering methods relying on
Gaussian~\cite{dissanayake2001solution, davison2002simultaneous,
	leonard2003consistent, se2002mobile, castellanos2004limits,
	clemente2007mapping} or Particle~\cite{montemerlo2002fastslam,
	montemerlo2003fastslam, grisetti2007improved,
	stachniss2007analyzing} representation of the posterior. 
Filtering approaches were preferred since they were regarded as
more suitable to be run on-line on a moving robot with the available
computational resources of the era, and the sparsity of the problem
had not yet been fully exploited.

In a Graph-Based \gls{slam} problem, it is common to have a number of
variables in the order of hundreds or thousands.  Such a high number
of variables results in a large optimization problem that represented
a challenge for the computers of the time, rendering global
optimization a bottleneck of Graph-Based \gls{slam} systems.  In the
remainder of this document we will refer to the global optimization
module in Graph-SLAM as the \emph{back-end}, in contrast to the
\emph{front-end} which is responsible to construct the factor graph
based on the sensor measurements.  Gutmann and
Konolidge~\cite{gutmann1999incremental} addressed the problem of
incrementally building a map, by finding topological relations and
loop closures.  This work was one of the first on-line implementations
of Graph-Based \gls{slam}. The core idea to reduce the computation in
the back-end was to restrict the optimization to the portions of the
graph having the larger errors.  This insight has inspired several
subsequent works~\cite{kaess2007isam,ila2017slam++}.

\subsection{Stand-Alone \gls{ils} solvers}
Whereas dense solvers are typically embedded in the specific
application for performance reasons, sparse solvers are complex enough
to motivate the design of generic libraries for\gls{ils}.  The first work to
explicitly consider the sparsity of \gls{slam} in conjunction with a
direct method to solve the linear system was \sqrtsam, developed by
Dallaert~\etal~\cite{dellaert2006square}.
Kaess~\etal~\cite{kaess2007isam} exploited this aspect of the problem
in iSAM, the second iteration of \sqrtsam. Here when a new edge is
added to the graph, the system computes a new solution reusing part
of the previous one and selectively updating the vertices. In the
third iteration of the system - iSAM2 -
Kaess~\etal~\cite{kaess2012isam2} exploited the Bayes Tree to solve
the optimization problem without explicitly constructing the linear
system. This solution is in contrast with the general trend of
decoupling linearization of the problem and solution of the linear
system and highlights the connections between the elimination
algorithms used in the solution of a linear system and inference on
graphical models. This self-contained engine allows deal very
efficiently with dynamic graph that grows during time and Gaussian
densities, two typical features of the \gls{slam} problem.  The final
iteration of the system, called GTSAM~\cite{dellaert2012gtsam}, embeds
all this concepts in a single framework.

Meanwhile, Hertzberg with his thesis~\cite{hertzberg2008thesis}
introduced the $\boxplus$ method to systematically deal with
non-Euclidean spaces and sparse problems. This work has been at the
root of the framework of K\"{u}mmerle \etal~\cite{kummerle2011g} - called 
\g2o.  This system introduces a layered
architecture that allows to easily exchange sub-modules of the system
- \eg~the linear solver or optimization algorithm.  A further paper of
Hertzberg~\emph{et al.}~\cite{hertzberg2013integrating} extends the
$\boxplus$ method \gls{ils} to filtering.

Agarwal~\etal~proposed in their Ceres Solver~\cite{ceres-solver} a 
generalized framework to perform non-linear optimization. Ceres embeds 
state-of-the-art methodologies that take advantages of modern CPUs - \eg~SIMD 
instructions and multi-threading - resulting in a complete and fast framework. 
One of its most relevant feature is represented by the efficient use of 
\gls{ad}~\cite{griewank1996algorithm}, that consists 
in the algorithmic computation of derivatives starting from the error 
function. Further information on the topic of~\gls{ad} can be found 
in~\cite{griewank1989automatic,griewank2008evaluating}.

In several contexts knowing the optimal value of a solution 
is not sufficient, and also the covariance is required.  In
\gls{slam}, knowing the marginal covariances relative to a variable is
fundamental to approach data-association.  To this extent
Kaess~\emph{et. al}~\cite{kaess2009covariance} outlined the use of the
elimination tree.  Subsequently, Ila~\etal~\cite{ila2017slam++}
designed SLAM++, an optimization framework to estimate mean and
covariance of the state by performing incremental Cholesky
updates. This work takes advantage of the incremental aspect of the
problem to selectively update the approximated Hessian matrix by using
parallel computation.

\gls{ils} algorithms have several known drawbacks. Perhaps the most
investigated aspect is the sensitivity of the solution to the initial
guess, that is reflected by the convergence basin. A wrong initial
guess might lead a non-linear solver to converge to an inconsistent local minimum.
Convex optimization~\cite{boyd2004convex} is one of the possible
strategies to overcome this problem, however its use is highly domain
dependent. Rosen~\etal~\cite{rosen2016sesync} explored this topic,
proposing a system to perform optimization of generic $\mathrm{SE}(d)$
factor graphs. In their system, called SE-Sync, they use Riemannian
Truncated-Newton Trust-Region method to certifiably compute the global
optimum in a two step optimization (rotation and translation).
Briales~\etal~\cite{briales2017cartan} extended this approach to
jointly optimize rotation and translation using the same concepts.
Bai~\etal~\cite{bai2018robust} provided a formulation of the
\gls{slam} problem based on \emph{constrained optimization}, where
constraints are represented by loop-closure cycles. Still, those
approaches are bounded to $\mathrm{SE}(d)$ sparse optimization.
In contrast, Ni~\etal~\cite{ni2010multi} and
Grisetti~\etal~\cite{grisetti2012robust} exploited respectively nested
dissection and hierarchical local sub-graphs devising divide and
conquer strategies to both increase the convergence basin and speed up
the computation.

\section{Least Squares Minimization}
\label{sec:least-squares}
\begin{figure*}[!t]
  \centering
  \includegraphics[width=0.8\linewidth]{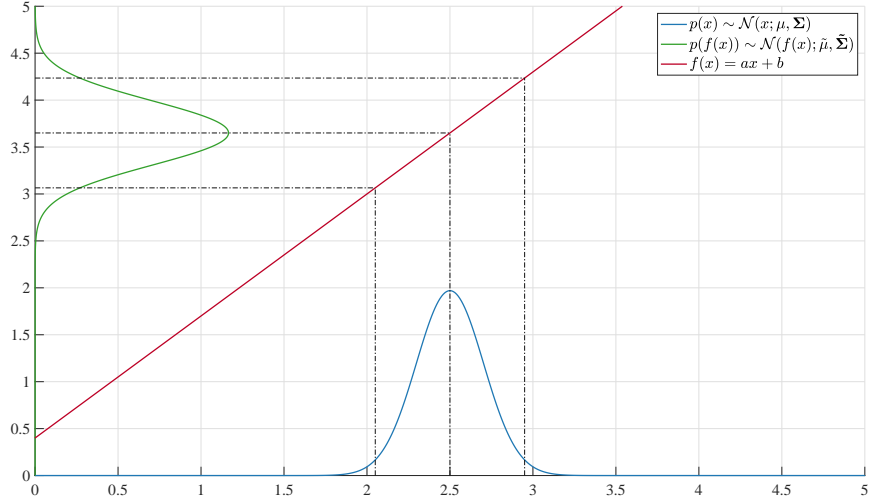}
  \caption{Affine transformation of a uni-variate Gaussian distribution. The 
    blue curve represents the source PDF while in green we show the output PDF. 
    The red line represents the affine transformation.}
  \label{fig:gaussian-transform}
\end{figure*}
      
This section describes the foundations of \gls{ils} minimization.  We
first present a formulation of the problem, that highlights its
probabilistic aspects (Section~\ref{sec:problem-formulation}).  In
Section~\ref{sec:linear-measurements} we review some basic rules for
manipulating the Normal distribution and we apply these rules to the
definition presented in Section~\ref{sec:problem-formulation}, leading
to the initial definition of linear \gls{ls}.  In
Section~\ref{sec:non-linear-ls} we discuss the effects of non-linear
observation model, assuming that both the state space and the
measurements space are Euclidean. Subsequently,
we relax this assumption on the structure of state and measurement
spaces, proposing a solution that uses smooth manifold
encapsulation.  
In Section~\ref{sec:robustifiers} we introduce effects of outliers in the 
optimization and we show commonly used methodologies to reject them.
Finally in Section~\ref{sec:sparsity}, we address the
case of large, sparse problem characterized by measurement functions
depending only on small subsets of the state. Classical problems such
as \gls{slam} or \gls{ba} fall in this category and are
characterized by a rather sparse structure.  

\subsection{Problem Formulation}\label{sec:problem-formulation}
Let~$\mathcal{W}$ be a stationary system whose non-observable state
variable is represented by~$\bx$ and let~$\bz$ be a measurement, \ie~a
perception of the environment. The state is distributed according to a
prior distribution $p(\bx)$, while the conditional distribution of the
measurement given the state $p(\bz|\bx)$ is known. $p(\bz|\bx)$ is
commonly referred to as the \textit{observation model}. Our goal is to
find the most likely distribution of states, given the measurements -
\ie~$p(\bx|\bz)$.  A straightforward application of the Bayes rule
results in the following:
\begin{equation}
p(\bx|\bz) = \frac{p(\bx)p(\bz|\bx)}{p(\bz)} \propto p(\bx)\,p(\bz|\bx).
\label{eq:bayes}
\end{equation}
The proportionality is a consequence of the normalization
factor~$p(\bz)$, which is constant.

In the reminder of this work, we will considers two key assumptions:
\begin{itemize}
\item[--] the prior about the states is uniform, \ie
  \begin{equation}
    p(\bx) = \cN(\bx;\bmu_x, \bSigma_x = \inf) = \cN(\bx; \bnu_x, \bOmega_x =0),
    \label{eq:prob_x}
  \end{equation}
  \item[--] the observation model is Gaussian, \ie
  \begin{equation}
    p(\bz|\bx) = \cN(\bz; \bmu_{z|x}, \bOmega_{z|x}^{-1}) 
    \quad \text{where} \quad \bmu_{z|x}=\bh(\bx) 
    \label{eq:prob_meas}.
  \end{equation} 
\end{itemize}
\eqref{eq:prob_x} expresses the uniform prior about the states using
the canonical parameterization of the Gaussian.  Alternatively, the
moment parameterization characterizes the Gaussian by the information
matrix - \ie~the inverse of the covariance matrix $\bOmega_x =
\bSigma_x^{-1}$ - and the information vector $\nu_x = \bOmega_x\mu_x$.
The canonical parameterization is better suited to represent
non-informative prior, since $\bOmega_x = 0$ does not lead to
numerical instabilities while implementing the algorithm.  In
contrast, the moment parameterization can express in a stable manner
situations of absolute certainty by setting $\bSigma_x = 0$.  In the
remainder, we will use both representations upon convenience, being
their relation clear.

In \eqref{eq:prob_meas} the mean $\mu_{z|x}$ of the predicted
measurement distribution is controlled by a generic non-linear
function of the state $\bh(\bx)$, commonly referred to as
\textit{measurement function}. In the next section we will derive the
solution for~\eqref{eq:bayes}, imposing that the measurement function
is an \textit{affine transformation} of the state - i.e. $\bh(\bx) =
\bA \bx + \bb$ - illustrated in~\figref{fig:gaussian-transform}. In 
Section~\ref{sec:non-linear-ls} we address the
more general non-linear case.

\subsection{Linear Measurement Function}\label{sec:linear-measurements}
In case of linear measurement function, expressed as $\bh(\bx) = \bA
(\bx-\bmu_x) + \hat\bz$, the prediction model has the following form:
\begin{align}
  p(\bz|\bx) &= \cN(\bz; \mu_{z|x}=\bA (\overbracket{\bx-\mu_x}^{\bDeltax}) +
  \hat \bz,\, \bOmega_{z|x}^{-1}) \nonumber\\
  p(\bz|\bDeltax) &= \cN(\bz; \mu_{z|\Delta x}=\bA \bDeltax + \hat \bz,
  \bOmega_{z|x}^{-1}),
  \label{eq:prob_linear_meas}
\end{align}
with $\hat \bz$ constant and $\bmu_x$ being the mean of the prior.
For convenience, we express the stochastic variable $\bDeltax=\bx-\mu_x$ as
\begin{equation}
p(\bDeltax) = \cN(\bDeltax;0, \bSigma_x = \inf) = \cN(\bx; \nu_{\Delta x},
\bOmega_x = 0)
\label{eq:prob_dx}.
\end{equation}
Switching between $\bDeltax$ and $\bx$ is achieved by summing or subtracting 
the mean.
To retrieve a solution for \eqref{eq:bayes} we first compute the joint
probability of $p(\bz,\bDeltax)$ using the chain rule and,
subsequently, we condition this joint distribution with respect to the
known measurement $\bz$. For further details on the Gaussian
manipulation, we refer the reader to~\cite{schon2011manipulating}.

\myParagraph{Chain Rule} \label{par:chain_rule}
\begin{figure*}[!t]
  \centering
  \includegraphics[width=0.8\linewidth]{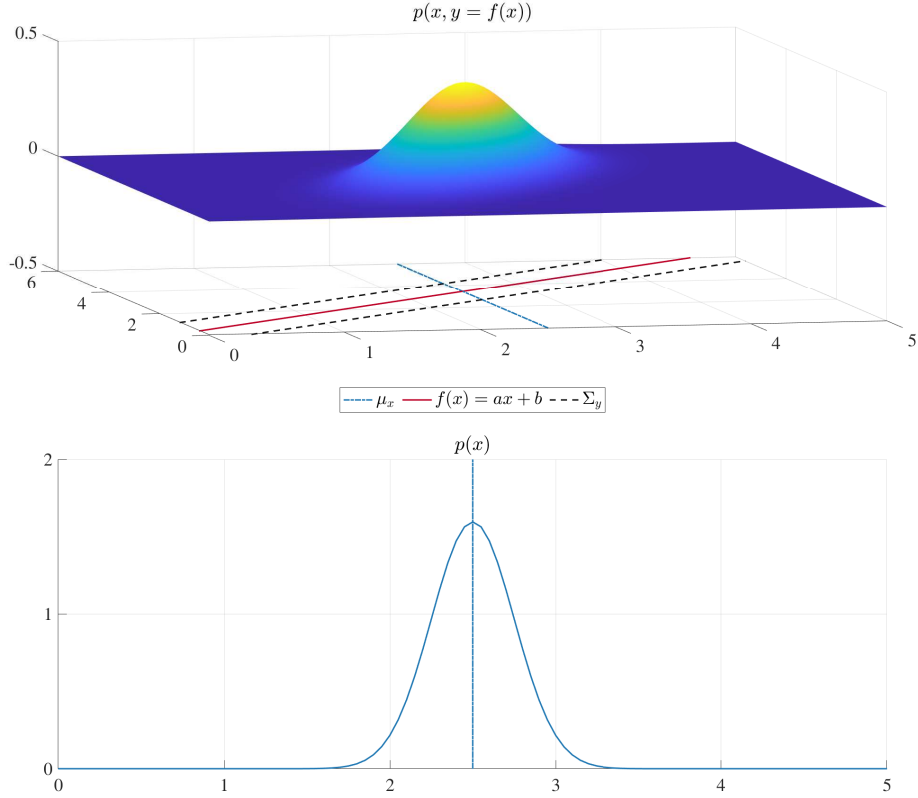}
  \caption{Given a uni-variate Gaussian PDF $p(x)$ and an affine 
  trasformation $f(x)$ - indicated in red - we computed the joint distribution 
  $p(x, y = f(x))$ through the chain rule.}
  \label{fig:gaussian-chain-rule}
\end{figure*}

Under the Gaussian assumptions made in the previous section,
the parameters of the joint distribution
over states and measurements $p(\bDeltax,\bz)$ have the following block structure:
\begin{align}
  p(\bDeltax,\bz) &= \cN\left(\bDeltax, \bz; \bmu_{\Delta x,z},\bOmega_{\Delta x,z}^{-1}\right)
  \label{eq:chain_rule}\\
  \bmu_{\Delta x,z}  =
   \begin{pmatrix}
     \bzero \\
     \hat\bz
   \end{pmatrix}
  &\qquad
  \bOmega_{\Delta x,z}  =
  \begin{pmatrix}
    \bOmega_{xx}  & \bOmega_{xz} \\
    \bOmega_{xz}^T  & \bOmega_{zz}
  \end{pmatrix}.
  \label{eq:chain_rule_things} 
\end{align}
The value of the terms in \eqref{eq:chain_rule_things} are obtained by applying the chain rule
to multivariate Gaussians to \eqref{eq:prob_linear_meas} and~\eqref{eq:prob_dx}, according to~\cite{schon2011manipulating},
and they result in the following:
\begin{align*}
  \bOmega_{xx} &= \bA^T \bOmega_{z|x} \bA + \bOmega_x \\
  \bOmega_{xz} &= - \bA^T \bOmega_{z|\Delta x} \\
  \bOmega_{zz} &= \bOmega_{z|x}. 
\end{align*}
Since we assumed the prior to be non-informative, we can set
$\bOmega_x = 0$. As a result, the information vector
$\bnu_{\Delta x,z}$ of the joint distribution is computed as:
\begin{equation}
  \bnu_{\Delta x,z} = 
  \begin{pmatrix}
    \bnu_{\Delta x} \\ \bnu_z
  \end{pmatrix} =
  \bOmega_{\Delta x,z} \,\bmu_{\Delta x,z} =
  \begin{pmatrix}
    -\bA^T \bOmega_{z|x} \hat\bz \\
    \bOmega_{z|x} \hat\bz
  \end{pmatrix}.
  \label{eq:info_vec_chain}
\end{equation}
\figref{fig:gaussian-chain-rule} shows visually the outcome distribution.

\myParagraph{Conditioning}\label{par:conditioning}
\begin{figure*}[!t]
  \centering
  \includegraphics[width=0.8\linewidth]{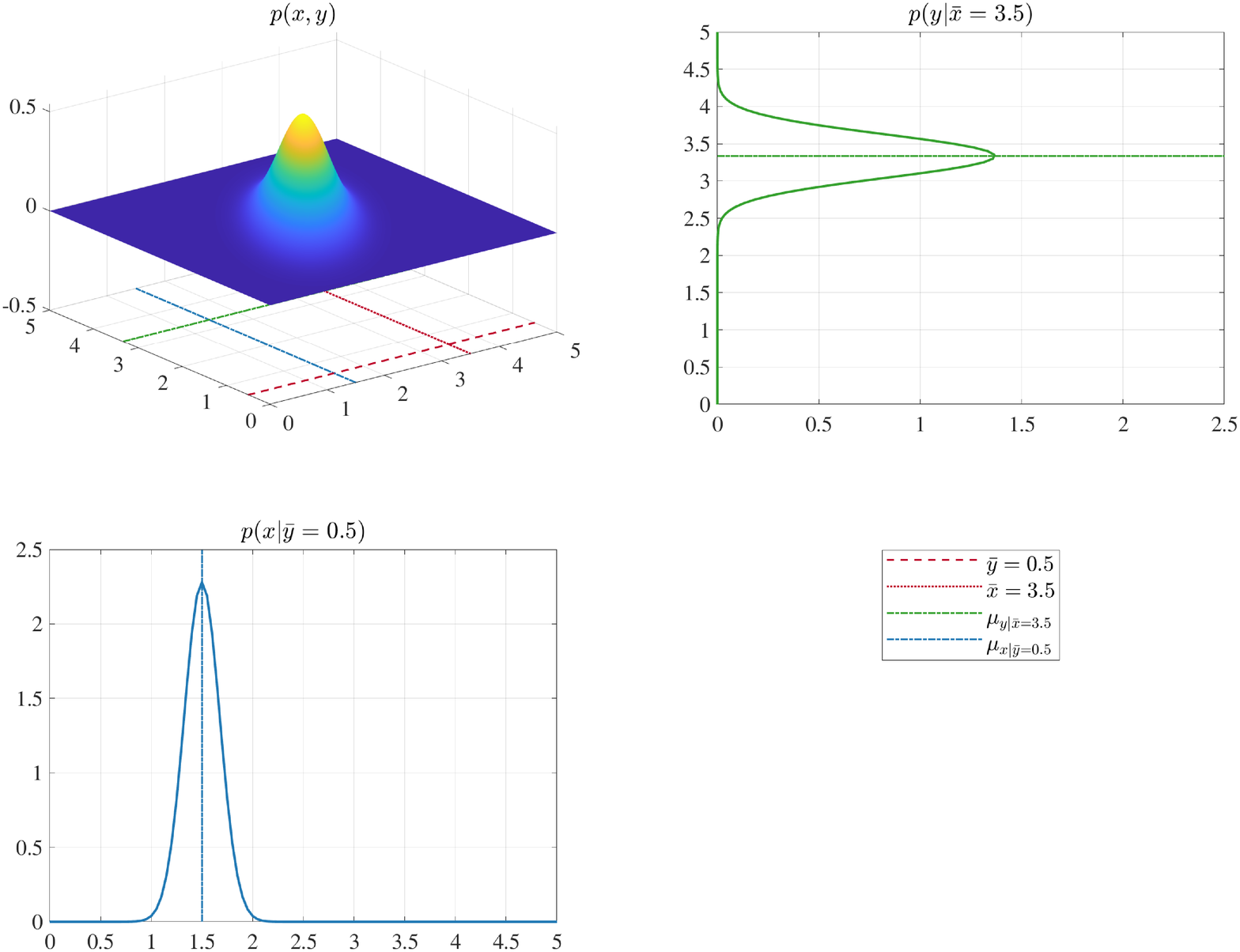}
  \caption{Conditioning of a bi-variate Gaussian. Top left: the source PDF; top 
  right and bottom left indicate the conditioning over $x$ and $y$ 
  respectively. }
  \label{fig:gaussian-conditioning}
\end{figure*}

Integrating the known measurement $\bz$ in the joint distribution $p(\bDeltax,\bz)$ of
\eqref{eq:chain_rule} results in a new distribution $p(\bDeltax|\bz)$. This can be done by
conditioning in the Gaussian domain. Once again we refer the reader to~\cite{schon2011manipulating}
for the proofs, while we report here the results that the conditioning has on the Gaussian parameters:
\begin{equation}
  \label{eq:conditioning}
  p(\bDeltax|\bz) \sim \cN\left(\bDeltax; \nu_{\Delta x|z},
  \bOmega_{\Delta x|z}\right)
\end{equation}
where
\begin{align}
  \bnu_{\Delta x|z} &=\bnu_{\Delta x} - \bOmega_{xz} \bz \nonumber \\
  &=\bnu_{\Delta x} - (-\bA^\top \bOmega_{z|x})\bz 
  =\bA^\top \bOmega_{z|x}(\underbracket{\bz - \hat\bz}_{-\be})
  \label{eq:conditioning_mu} \\
  \bOmega_{\Delta x|z} &= \bOmega_{zz} = \underbracket{\bA^\top \bOmega_{z|x} \bA}_{\bH}.
  \label{eq:conditioning_sigma}
\end{align}
The conditioned mean $\bmu_{\Delta x|z}$ is retrieved from the
information matrix and the information vector as:
\begin{align}
  \bmu_{\Delta x|z} &= \bOmega_{\Delta x|z}^{-1}\bnu_{x|z} 
  = -{\bH}^{-1}
  \underbracket{\bA^\top \bOmega_{z|x} \be}_{\bb} = -\bH^{-1} \bb.
  \label{eq:linear-solve}
\end{align}
Remembering that $\bDeltax = \bx -\mu_x$, the Gaussian distribution over the 
conditioned states has the same information matrix, while the mean is obtained 
by summing the increment's mean $\bmu_{\Delta x|z}$ as
\begin{equation}
  \bmu_{x|z} = \bmu_x + \bmu_{\Delta x|z}.
  \label{eq:linear-increment}
\end{equation}
An important result in this derivation is that the matrix
$\bH=\bOmega_{\Delta x|z}$ is the information matrix of the the estimate,
therefore, we can estimate not only the optimal solution $\bmu_{x|z}$, but
also its uncertainty $\bSigma_{x\mid z}=\bOmega_{\Delta x|z}^{-1}$.
\figref{fig:gaussian-conditioning} illustrates visually the conditioning of a 
bi-variate Gaussian distribution.

\myParagraph{Integrating Multiple Measurements}\label{par:multiple-meas}
Integrating multiple independent measurements $\bz_{1:K}$ requires to
stack them in a single vector. As a result, the observation model
becomes
\begin{align}
  \label{eq:multiple_meas}
        {} & p(\bz_{1:K}|\bDeltax) = \prod_{k = 1}^{K} p(\bz_k | \bDeltax) \sim
        \cN(\bz; \bmu_{z|x}, \bOmega_{z|x}) = \\
        &= \left(
        \begin{bmatrix} \bz_1 \\ \vdots \\ \bz_K\end{bmatrix}; \,
          \begin{bmatrix} \bA_1 \\ \vdots \\ \bA_K \end{bmatrix} \bDeltax +
          \begin{bmatrix} \hat \bz_1 \\ \vdots \\ \hat \bz_K \end{bmatrix} ,
          \begin{bmatrix} \bOmega_{z_1|x} & & \\ & \ddots & \\ & &
            \bOmega_{z_K|x} \end{bmatrix}
          \right).
          \nonumber
\end{align}
Hence, matrix $\bH$ and vector $\mathbf{b}$ are composed by the sum of
each measurement's contribution; setting $\be_k=\hat\bz_k-\bz_k$, we compute 
them as follows:
\begin{equation}
  \label{eq:total-hessian}
  \bH = \sum_{k = 1}^{K} \underbracket{\bA_k^\top \bOmega_{z_k|x} \bA_k}_{\bH_k}
  \qquad
  \mathbf{b} = \sum_{k = 1}^{K} \underbracket{ \bA_k^\top \bOmega_{z_k|x}
    \be_k}_{\mathbf{b}_k}.
\end{equation}

\subsection{Non-Linear Measurement Function}\label{sec:non-linear-ls}
Equations~(\ref{eq:linear-solve})~(\ref{eq:linear-increment})
and~(\ref{eq:total-hessian}) allow us to find the exact mean of the
conditional distribution, under the assumptions that i) the
measurement noise is Gaussian, ii) the measurement function is an
affine transform of the state and iii) both measurement and state
spaces are Euclidean.  In this section we first relax the assumption
on the affinity of the measurement function, leading to the common
derivation of the \gls{gn} algorithm. Subsequently, we address the
case of non-Euclidean state and measurement spaces.

If the measurement model mean $\bmu_{z|x}$ is controlled by
a non-linear but \textit{smooth} function $\bh(\bx)$, and that the
prior mean $\mu_x=\breve \bx$ is reasonably close to the optimum, we
can approximate the behavior of $\bmu_{x|z}$ through the
first-order Taylor expansion of $\bh(\bx)$ around the mean, namely:
\begin{align}
  \bh(\breve \bx + \bDeltax) \approx
  \underbracket{\bh(\breve \bx)}_{\hat \bz} +
  \underbracket{\frac{\partial \bh(\bx)}{\partial\bx}\bigg\rvert_
    {\bx=\breve \bx}}_{\bJ}\bDeltax = \bJ \bDeltax + \hat \bz.
  \label{eq:linearization}
\end{align}
The Taylor expansion reduces conditional mean to an affine transform in
$\bDeltax$.  Whereas the conditional distribution will not be in
general Gaussian, the parameters of a Gaussian approximation can still
be obtained around the optimum through \eqref{eq:conditioning_sigma}
and \eqref{eq:linear-increment}.  Thus, we can use the same algorithm
described in \secref{sec:linear-measurements}, but we have to compute
the linearization at each step.  
Summarizing, at each iteration, the \gls{gn} algorithm:
\begin{itemize}
  \item[--] processes each measurement $\bz_k$ by evaluating error
  $\be_k(\bx) = \bh_k(\bx) - \bz_k$ and Jacobian $\bJ_k$
  at the current solution $\breve \bx$:
  \begin{align}
    \label {eq:nonlinear-error}
    \be_k &= \bh(\breve \bx) - \bz\\
    \label {eq:nonlinear-jacobian}
    \bJ_k &= \frac{\partial \bh_k(\bx)}{\partial\bx}\bigg\rvert_{\bx = \breve 
    \bx}.
  \end{align}
  \item[--] builds a coefficient matrix and coefficient vector for the
  linear system in \eqref{eq:linear-solve}, and computes the optimal
  perturbation $\bDeltax$ by solving a linear system:
  \begin{align}
    \label {eq:nonlinear-solution}
    \bDeltax = &-\bH^{-1} \bb \\
    \bH = \sum_{k = 1}^{K} \bJ_k^\top \bOmega_{z_k|x} \bJ_k &\quad
    \bb = \sum_{k = 1}^{K} \bJ_k^\top \bOmega_{z_k|x} \be_k 
    \nonumber
  \end{align}
  \item[--] applies the computed perturbation to the current state as in
  \eqref{eq:linear-increment} to get an improved estimate
  \begin{equation}
    \breve \bx \leftarrow \breve \bx + \bDeltax.
    \label {eq:nonlinear-increments}
  \end{equation}    
\end{itemize}
A smooth prediction function has lower-magnitude higher order terms in
its Taylor expansion.  The smaller these terms are, the better its
linear approximation will be. This leads to the situations close to
the ideal affine case.  In general, the \emph{smoother} the
measurement function $\bh(\bx)$ is and the closer the initial guess is
to the optimum, 
the better the convergence properties of the problem.

\myParagraph{Non-Euclidean Spaces}\label{par:non-euclidean-spaces}
\begin{figure}[!t]
  \centering
  \includegraphics[width=0.99\columnwidth]{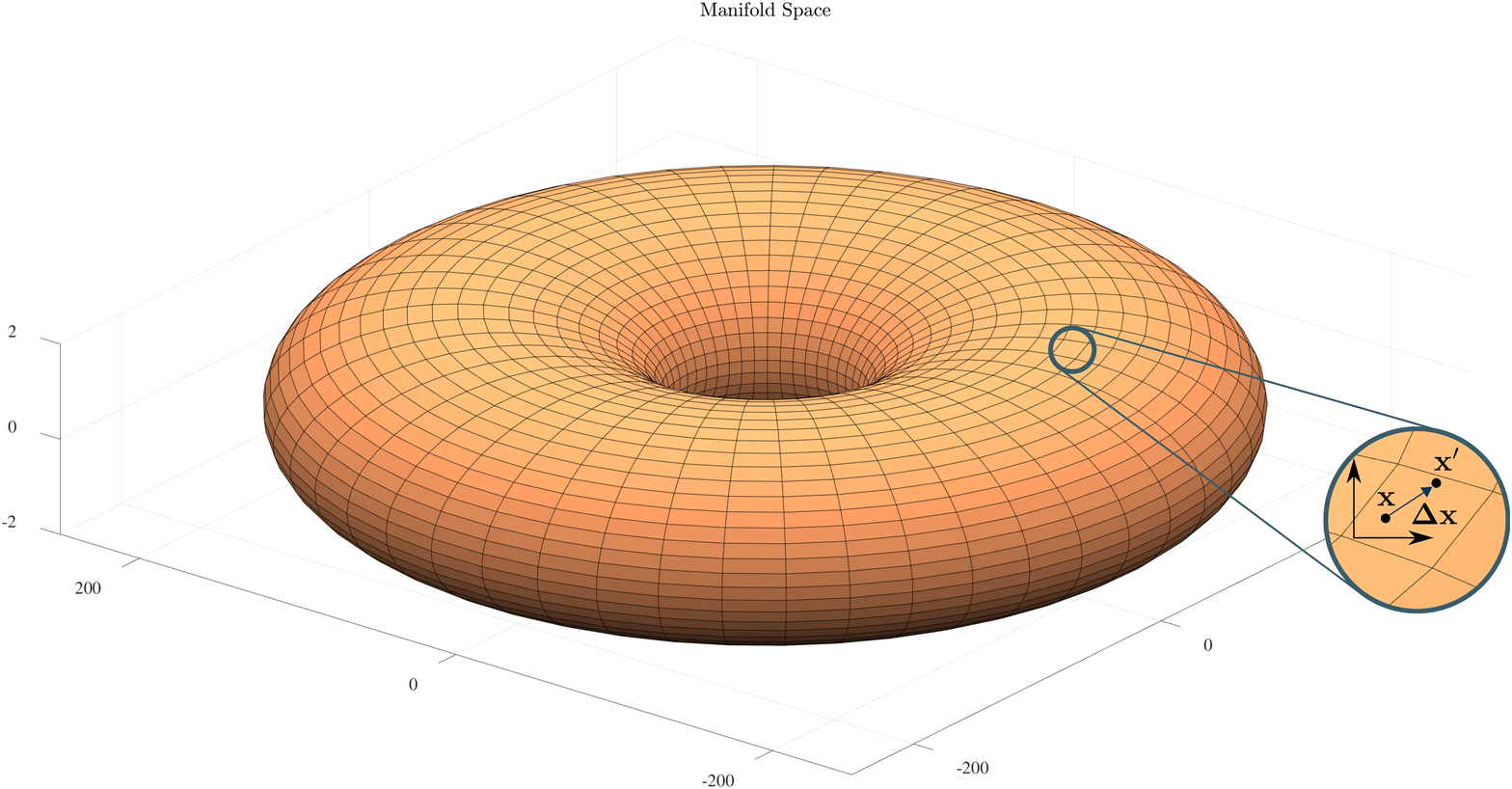}
  \caption{Illustration of a Manifold space. Since the manifold is smooth, 
  \emph{local} perturbations - \ie~$\bDeltax$ in the illustration - can be 
  expressed with a suitable Euclidean vector.}
  \label{fig:manifolds}
\end{figure}
The
previous formulation of \gls{gn} algorithm uses vector addition and
subtraction to compute the error $\be_k$ in \eqref{eq:nonlinear-error}
and to apply the increments in \eqref{eq:nonlinear-increments}. 
However, these two operations are only defined in Euclidean spaces. When this 
assumption is violated -
as it usually happens in Robotics and Computer Vision applications - the 
straightforward implementation does not generally provide satisfactory results.
Rotation matrices or angles cannot be directly added or subtracted
without performing subsequent non-trivial normalization. Still,
typical continuous states involving rotational or similarity transformations are
known to lie on a smooth manifold~\cite{lee2013smooth-manifolds}.

A smooth manifold $\mathbb{M}$ is a space that, albeit not homeomorphic to
$\mathbb{R}^n$, admits a locally Euclidean parameterization around each element
$\bM$ of the domain, commonly referred to as \emph{chart} - as illustrated 
in~\figref{fig:manifolds}. A chart computed in
a manifold point $\bM$ is a function from $\mathbb{R}^n$ to a new point
$\bM^\prime$
on the manifold:
\begin{equation}
\manifoldchart_\bM(\bDeltam):\mathbb{R}^n \rightarrow \mathbb{M}.
\label{eq:chart}
\end{equation}
Intuitively, $\bM^\prime$ is obtained by ``walking'' along the
perturbation $\bDeltam$ on the chart, starting from the origin. A null motion
($\bDeltam=\bzero$) on the chart leaves us at the point where the chart is
constructed - \ie~$\manifoldchart_\bM(\bzero)=\bM$.

Similarly, given two points $\bM$ and $\bM^\prime$ on the manifold, we can
determine the motion $\bDeltam$ on the chart constructed on $\bM$ that would
bring us to $\bM^\prime$. Let this operation be the inverse chart, denoted as 
$\manifoldchart_{\bM}^{-1}(\bM^\prime)$.  The direct and inverse charts allow us to
define operators on the manifold that are analogous to the sum and subtraction.
Those operators, referred to as $\boxplus$ and $\boxminus$, are, thence,
defined as:
\begin{align}
  \label{eq:generic-box-plus}
  \bM] &= \bM \boxplus \bDeltam \triangleq \manifoldchart_\bM(\bDeltam)\\
  \label{eq:generic-box-minus}
  \bDeltam &= \bM^\prime \boxminus \bM \triangleq \manifoldchart_{\bM}^{-1}(\bM^\prime).
\end{align}
This notation - firstly introduced by
Smith~\etal~\cite{smith1986representation} and then generalized by
Hertzberg~\etal~\cite{hertzberg2013integrating,hertzberg2012tutorial} - 
allows us to straightforwardly adapt the Euclidean version of \gls{ils} to 
operate on manifold spaces. The dimension of the chart is chosen to be the
minimal needed to represent a generic perturbation on tha manifold.  On
the contrary, the manifold representation can be chosen
arbitrarily.

A typical example of smooth manifold is the $SO(3)$ domain of 3D
rotations.  We represent an element $SO(3)$ on the manifold as a
rotation matrix $\bR$. In contrast, the for perturbation, we pick a
minimal representation consisting on the three Euler angles $\bDeltar
= (\Delta \phi, \Delta\theta, \Delta\psi)^\top$. Accordingly, the
operators become:
\begin{align}
  \label{eq:rot-box-plus}
  \bR_A \boxplus \bDeltar &= \fromvec(\bDeltar)\,\bR_A\\
  \label{eq:rot-box-minus}
  \bR_A \boxminus \bR_B &= \tovec(\bR_B^{-1}\,\bR_A).
\end{align}
The function~$\fromvec(\cdot)$ computes a rotation matrix as
the composition of the rotation matrices relative to each Euler
angle. In formul\ae:
\begin{equation}
  \bR = \fromvec(\bDeltar) = \bR_x(\Delta \phi) \, \bR_y(\Delta
  \theta)\, \bR_z(\Delta \psi).
  \label{eq:from-vector}
\end{equation}
The function~$\tovec(\cdot)$ does the opposite by computing the value
of each Euler angle starting from the matrix~$\bR$. It operates by
equating each of its element to the corresponding one in the matrix
product~$\bR_x(\Delta \phi) \, \bR_y(\Delta \theta)\, \bR_z(\Delta
\psi)$, and by solving the resulting set of trigonometric equations.
As a result, this operation is quite articulated.  Around the origin
the chart constructed in this manner is immune to singularities.

\begin{figure*}[!t]
  \centering
  \includegraphics[width=0.8\linewidth]{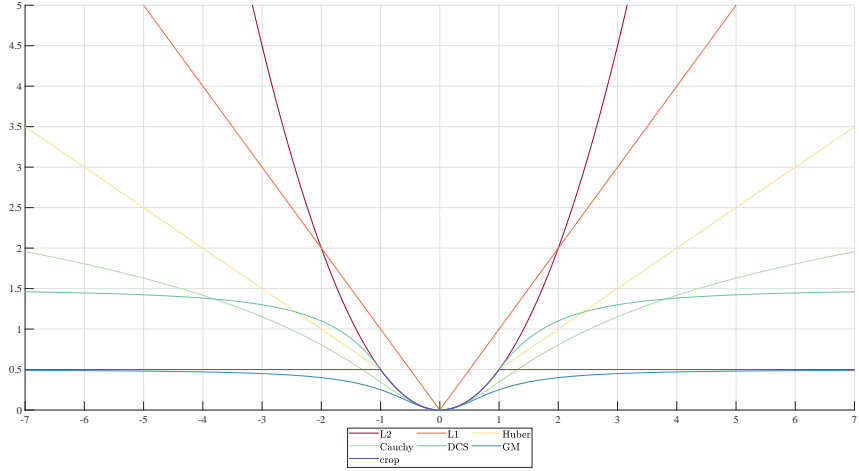}
  \caption{Commonly used robust kernel functions. The kernel threshold is set 
    to 1 in all cases.}
  \label{fig:robust-kernels}
\end{figure*}

Once defined proper $\boxplus$ and $\boxminus$ operators,
we can reformulate our minimization problem in the manifold
domain. To this extent we can simply  replace the $+$ with a $\boxplus$ in the 
computation of the Taylor expansion of \eqref{eq:linearization}. Since we will 
compute an increment on the chart, we need to compute the expansion on the 
chart $\bDeltax$ at the local optimum, that is at the origin
of the chart itself $\bDeltax=0$, in formul\ae:
\begin{equation}
  \label{eq:linearization-manifold}
  \bh_k(\breve \bX \boxplus \bDeltax) \approx \bh_k(\breve \bX) +
  \underbracket{\frac{\partial\bh_k(\breve \bX \boxplus
      \bDeltax)}{\partial\bDeltax}\bigg\rvert_{\bDeltax =
      0}}_{\breve{\bJ}_k}\bDeltax.
\end{equation}
The same holds when applying the increments in \eqref{eq:nonlinear-increments}, 
leading to:
\begin{equation}
  \label{eq:nonlinear-increments-manifold}
  \breve \bX \leftarrow \breve \bX \boxplus \bDeltax.
\end{equation}
Here we denoted with capital letters the manifold representation of
the state $\bX$, and with $\bDeltax$ the Euclidean perturbation. Since
the optimization \emph{within one iteration} is conducted on the chart, the
origin of the chart $\breve\bX$ on the manifold stays constant during
this iteration. 
If the measurements lie on a manifold too, a local $\boxminus$
operator is required to compute the error, namely:
\begin{equation}
\be_k(\bX) = \hat\bZ_k \boxminus \bZ_k = \bh_k(\bX) \boxminus \bZ_k.
\label{eq:manifold-error}
\end{equation}
To apply the previously defined optimization algorithm we should linearize the
error around the current estimate through its first-order Taylor expansion.
Posing $\breve\be_k = \be_k(\breve{\bX})$, we have the following relation:
\begin{equation}
  \be_k(\breve \bX \boxplus\bDeltax) = \bh_k(\breve \bX) \boxminus \bZ_k 
  \approx \breve\be_k +\frac{\partial \be_k  
  (\breve\bX\boxplus\bDeltax)}{\partial\bDeltax}\bigg\rvert_{\bDeltax=0}\bDeltax
  = \breve\be_k + \tilde\bJ_k \bDeltax.
  \label{eq:manifold-jac}
\end{equation}
The reader might notice that in~\eqref{eq:manifold-jac} the error space may 
differ from the increments one, due to the $\boxminus$ operator. As reported 
in~\cite{aloise2019chordal}, having a different parametrization might enhance 
the convergence properties of the optimization in specific scenarios. 
Still, to avoid any inconsistencies, the information matrix $\bOmega_k$ should 
be expressed on a chart around the current measurement $\bZ_k$.

\subsection{Handling Outliers: Robust Cost Functions}\label{sec:robustifiers}
In~\secref{sec:non-linear-ls} we described a methodology to compute the 
parameters of the Gaussian distribution over the state $\bx$ which minimizes 
the Omega-norm of the error between prediction and observation. More concisely,
we compute the optimal state $\bx^\star$ such that:
\begin{equation}
  \bx^* = \argmin_\bx \sum_{k=1}^{K} \lVert \be_k(\bx) \rVert_{\bOmega_k}^2.
  \label{eq:minimize}
\end{equation}
The mean of our estimate $\mu_{x|z}=\bx^*$ is the local optimum of the
\gls{gn} algorithm, and the information matrix $\bOmega^*_{x|z}=\bH^*$, is he 
coefficient matrix of the system at convergence.
The procedure reported in the previous section assumes all
measurements \emph{correct}, albeit affected by noise. 
Still, in many real cases this is not the case. This is mainly due to aspects 
that are hard to model - \ie~multi-path phenomena or incorrect data
associations. These \emph{wrong} measurements are commonly referred to
as \emph{outliers}. On the contrary, \emph{inliers} represent the \emph{good} 
measurements.

A common assumption made by several techniques to reject outliers is
that the inliers tend to agree towards a common solution, while
outliers do not. This fact is at the root of consensus schemes such as
RANSAC~\cite{fischler1981random}.  In the context of \gls{ils}, the
quadratic nature of the error terms leads to over-accounting for
measurement whose error is large, albeit those measurements are
typically outliers. However, there are circumstances where all errors
are quite large even if no outliers are present. A typical case occurs
when we start the optimization from an initial guess which is far from the
optimum.

A possible solution to this issue consists
in carrying on the optimization under a cost
function that grows sub-quadratically with the error.
Indicating with $u_k(\bx)$ the L1 Omega-norm of the error term 
in~\eqref{eq:nonlinear-error}, its derivatives with respect to the state 
variable $\bx$ can be computed as follows:
\begin{align}
  u_k(\bx) &= \sqrt{\be_k(\bx)^T \bOmega_k \be_k(\bx)} \label{eq:l1-norm}\\
  \frac{ \partial u_k(\bx)} {\partial \bx} &=
  \frac{1}{u_k(\bx)} \be_k(\bx)^T \bOmega_k \frac{ \partial\be_k(\bx) } 
  {\partial \bx } .
  \label{eq:l1-norm-derivative}
\end{align}
We can generalize \eqref{eq:minimize} by introducing a scalar function
$\rho(u)$ that computes a new error term as a function of the L1-norm.  
\eqref{eq:minimize} is a special case for $\rho(u)=\frac{1}{2}u^2$.
Thence, our new problem will consist in minimizing the following function:
\begin{equation}
  \bx^* = \argmin_\bx \sum_{k=1}^{K} \rho(u_k(\bx))
  \label{eq:minimize-robust}
\end{equation}
Going more in detail and analyzing the gradients of \eqref{eq:minimize-robust} 
we have the following relation:
\begin{align}
  \frac{ \partial \rho(u_k(\bx))} {\partial \bx} &=
  \left. \frac{\partial \rho(u)}{\partial u} \right|_{u=u_k(\bx)} 
  \frac{\partial u_k(\bx)} {\partial \bx} \nonumber\\
  &=
  \left. \frac{\partial \rho(u)}{\partial u} \right|_{u=u_k(\bx)}
  \frac{1}{u_k(\bx)} \be_k(\bx)^T \bOmega_k \frac{ \partial\be_k(\bx) } 
  {\partial \bx } \nonumber\\
  &=\gamma_k(\bx) \be_k(\bx)^T \bOmega_k \frac{ \partial\be_k(\bx) } {\partial 
  \bx }  \label{eq:minimize-robust-gradient}
\end{align}
where 
\begin{equation}
  \gamma_k(\bx) = \left. \frac{\partial \rho(u)}{\partial u} 
  \right|_{u=u_k(\bx)} \frac{1}{u_k(\bx)}.
\end{equation}
The \emph{robustifier} function $\rho(\cdot)$ acts on the gradient, 
modulating the 
magnitude of the error term through a scalar function $\gamma_k(\bx)$.
Still, we can also compute the gradient of the \eqref{eq:minimize} as follows:
\begin{equation}
  \frac{\partial \lVert \be_k(\bx) \rVert_{\bOmega_k}^2}{\partial \bx} = 
  2 \be_k(\bx)^T \bOmega_k \frac{\partial \be_k(\bx)}{\partial 
  \bx}.\label{eq:minimize-gradient}
\end{equation}
We notice that \eqref{eq:minimize-gradient} and
\eqref{eq:minimize-robust-gradient} differ by a scalar term $\gamma(\bx)$ that
depends on $\bx$.  By absorbing this scalar term at each iteration in
a new information matrix $\bar\bOmega_k(\bx)=\gamma_k(\bx) \bOmega_k$, we
can rely on the iterative algorithm illustrated in the previous
sections to implement a robust estimator. In this sense, at each iteration we 
compute $\gamma_k(\bx)$ based on the result of the previous iteration. 
Note that, upon convergence the \eqref{eq:minimize-gradient} and
\eqref{eq:minimize-robust-gradient} will be the same, therefore, they lead to 
the same optimum. This formalization of the problem is called \gls{irls}.

The use of robust cost functions biases the information matrix of the system 
$\bH$. Accordingly, if we want to recover an estimate of the solution 
uncertainty when using robust cost functions, we need to ``undo'' the
effect of function $\rho(\cdot)$. 
This can be easily achieved recomputing $\bH$ after convergence considering 
only inliers and disabling the robustifier - \ie~setting 
$\rho(u)=\frac{1}{2}u^2$.~\figref{fig:robust-kernels} illustrates some of the 
most common cost function used in Robotics and Computer Vision.
Further information on modern robust cost function can be found in the work of 
MacTavish~\etal~\cite{mactavish2015all}.

\subsection{Sparsity}\label{sec:sparsity}
\begin{figure}[!t]
  \centering
  \begin{subfigure}{0.489\columnwidth}
    \centering
    \includegraphics[width=\linewidth]{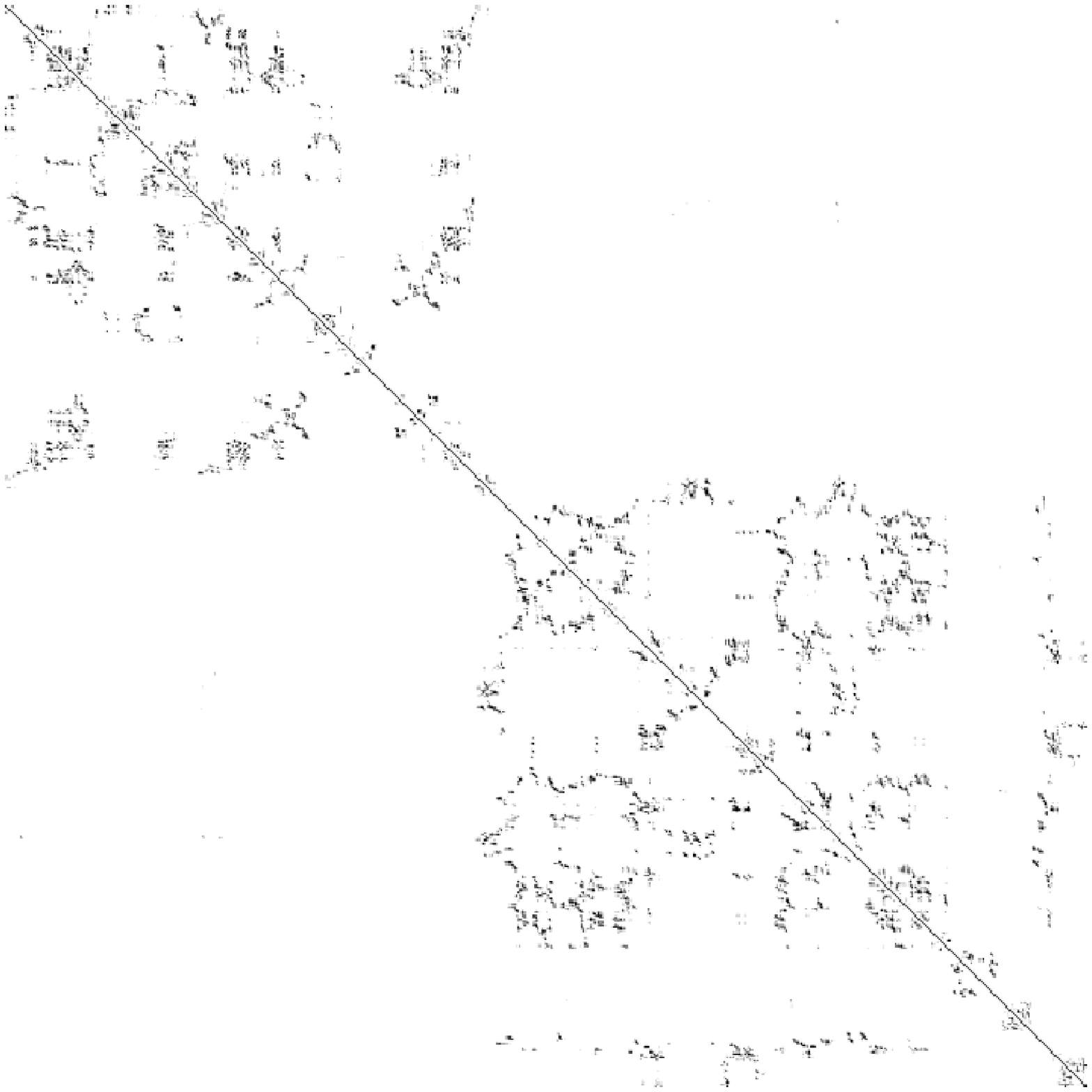}
    \subcaption{$\bH$ matrix before reordering.}
    \label{fig:reordering-raw-H}
  \end{subfigure}
  \begin{subfigure}{0.489\columnwidth}
    \centering
    \includegraphics[width=\linewidth]{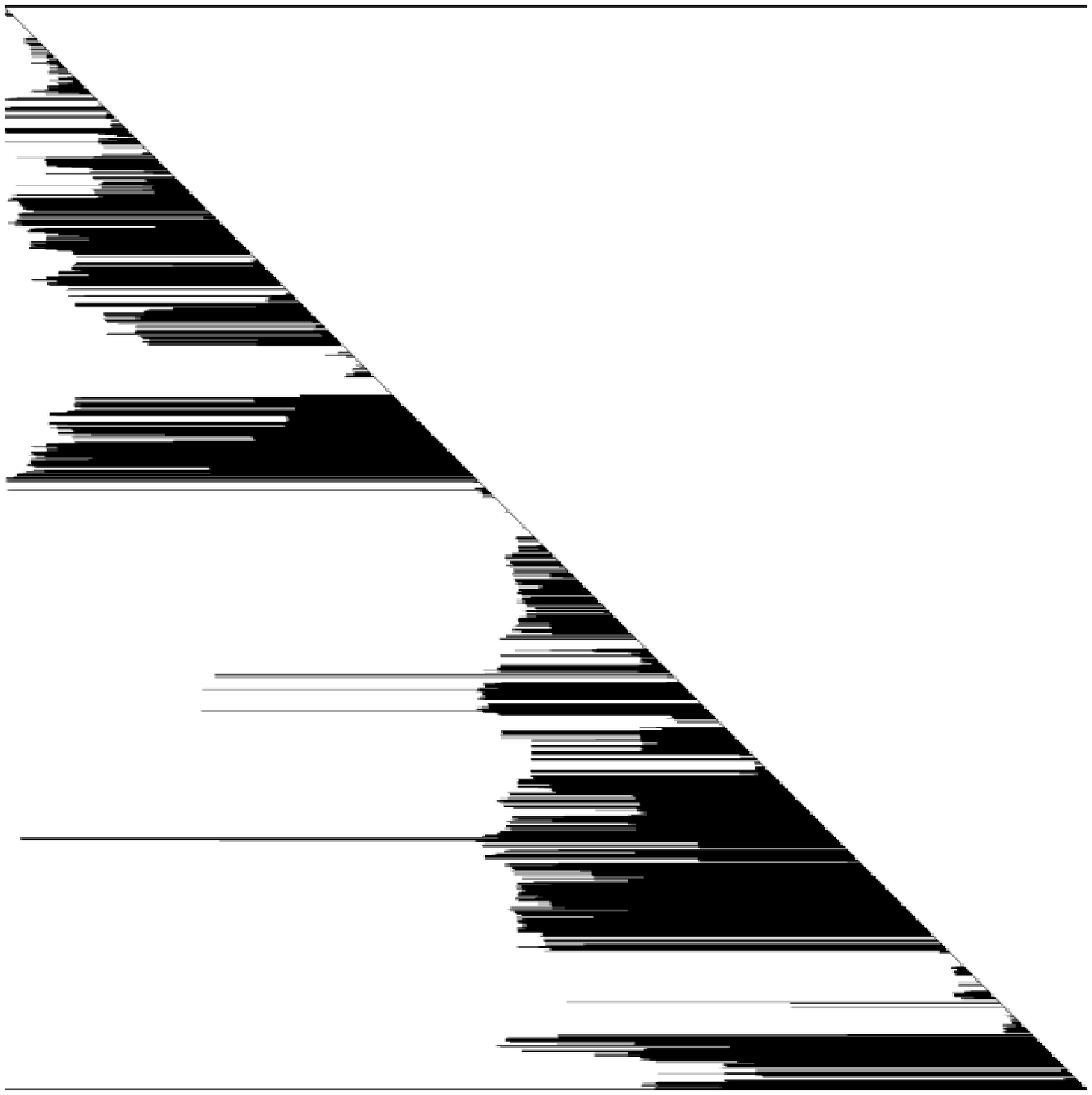}
    \subcaption{Cholesky decomposition before reordering.}
    \label{fig:reordering-raw-L}
  \end{subfigure} \\ \vspace{10pt}
  \begin{subfigure}{0.489\columnwidth}
    \centering
    \includegraphics[width=\linewidth]{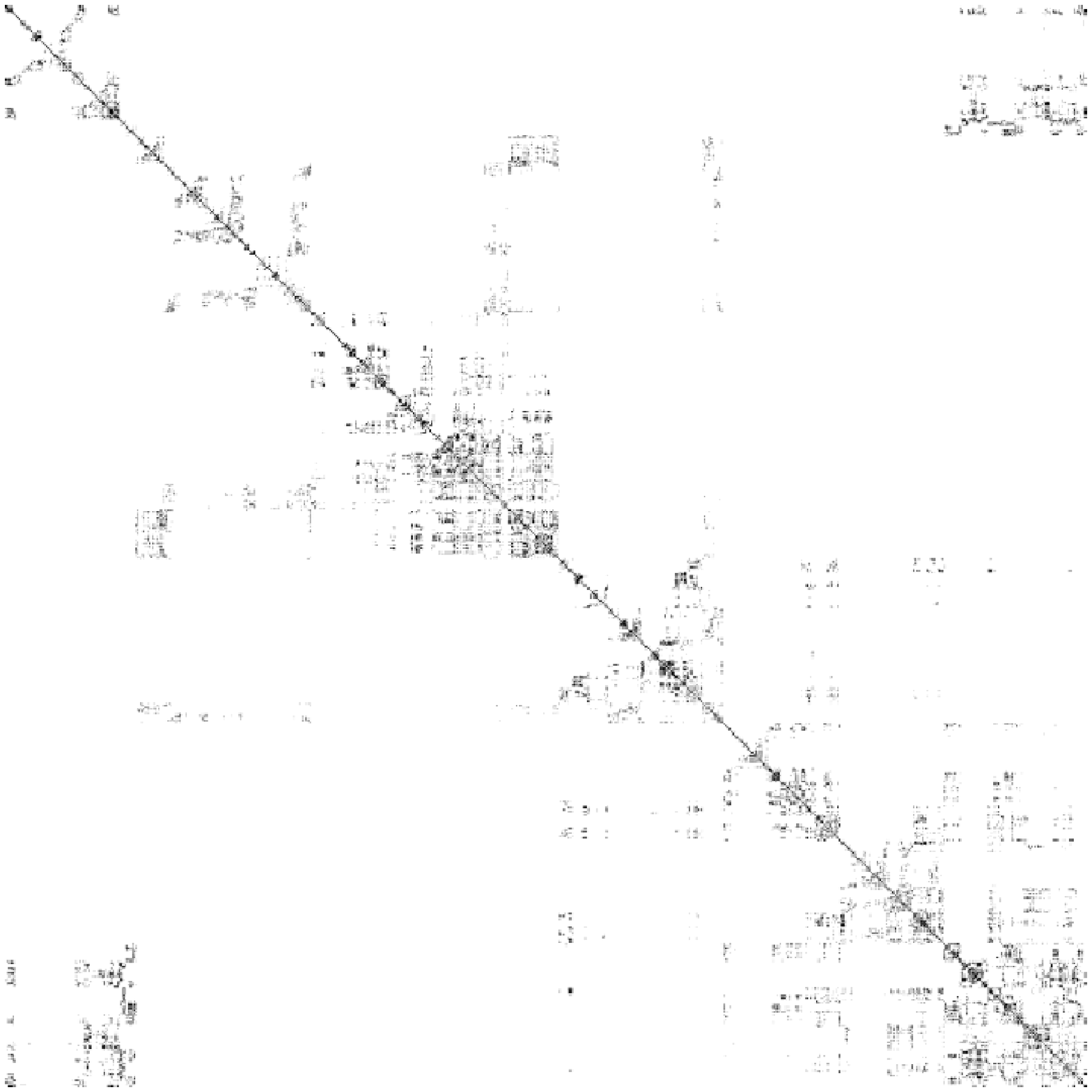}
    \subcaption{$\bH$ matrix after reordering.}
    \label{fig:reordering-amd-H}
  \end{subfigure}
  \begin{subfigure}{0.489\columnwidth}
    \centering
    \includegraphics[width=\linewidth]{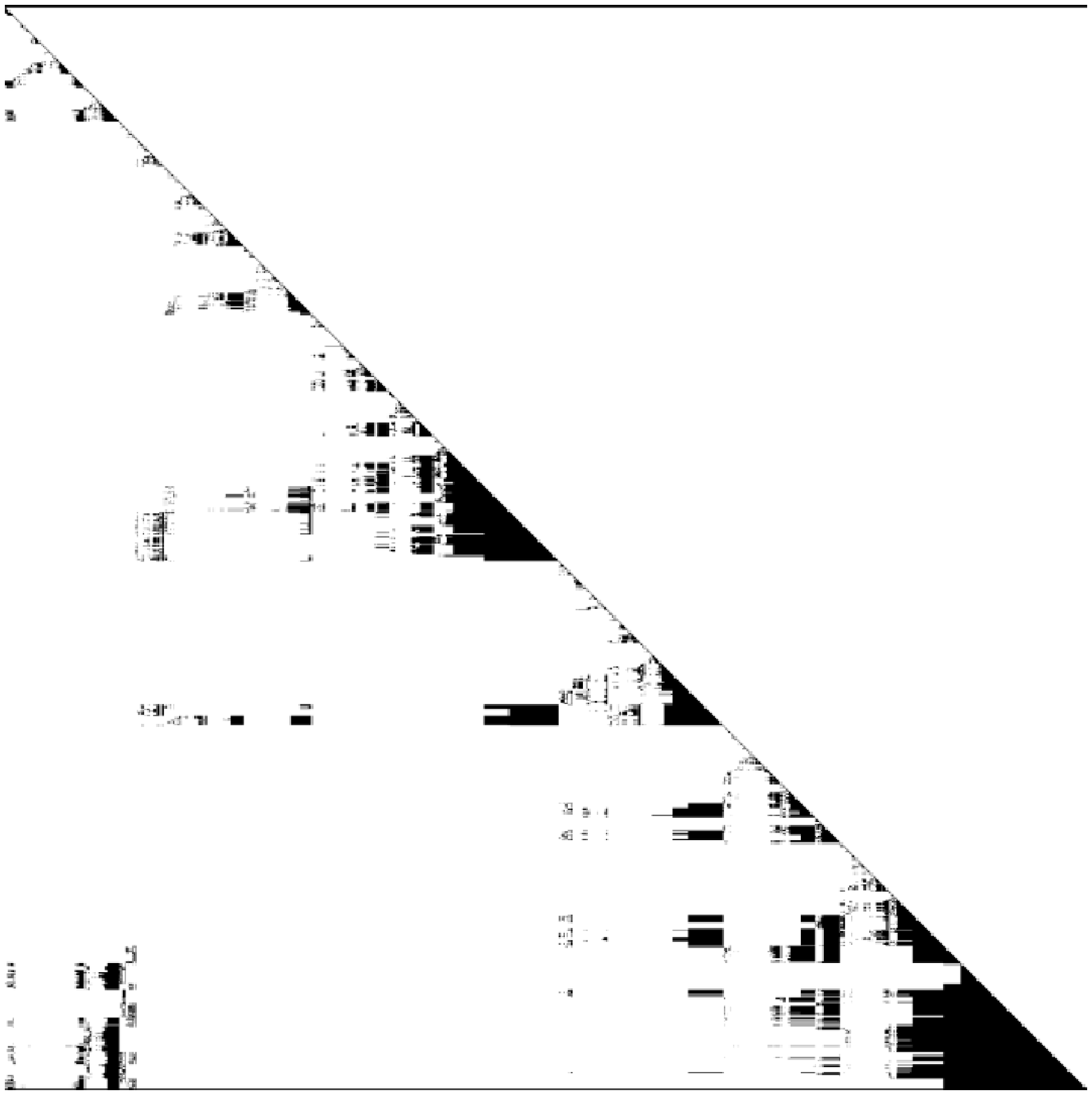}
    \subcaption{Cholesky decomposition after reordering.}
    \label{fig:reordering-amd-L}
  \end{subfigure} 
  \caption{Effects of AMD variable reordering on the fill-in of the Cholesky 
    decomposition of matrix $\bH$. Black pixels indicate non-zero blocks. As 
    illustrated in~\figref{fig:reordering-raw-L} 
    and~\figref{fig:reordering-amd-L}, variable reordering dramatically reduces 
    the fill-in of the decomposed matrix.}
  \label{fig:reordering}
\end{figure}
Minimization algorithms like \gls{gn} or \gls{lm} lead to the repeated 
construction and solution of the linear system $\bH\bDeltax = -\mathbf{b}$.  
In many cases, each measurement $\bz_k$ only involves a
small subset of state variables, namely:
\begin{equation}
  \bh_k(\bx) = \bh_k(\bx_k) \quad \mbox{where} \quad \bx_k= \{ \bx_{k_1}, 
  \dots, \bx_{k_q} \} \in \bx.
\end{equation}
Therefore, the Jacobian for the error term $k$ has the following structure:
\begin{equation}
  \bJ_k = \left[\bzero \cdots \bzero \, \bJ_{k_1} \, \bzero \cdots \bzero
  \, \bJ_{k_h} \, \bzero \cdots \bzero \, \bJ_{k_q} \, \bzero \cdots
  \bzero \right].
\end{equation}
According to this, the contribution $\bH_k = \bJ_k^T \bOmega_k \bJ_k$ of the 
$k^\mathrm{th}$ measurement to the system matrix $\bH$ exhibits the following 
pattern:
\begin{equation*}
  \bH_k =
  \begin{footnotesize}
  \begin{pmatrix}
  \cdot &  &  &  &  &  &\\
  & \bJ_{k_1}^T\bOmega_k\bJ_{k_1} & \cdots & \bJ_{k_1}^T\bOmega_k\bJ_{k_h} &
  \cdots & \bJ_{k_1}^T\bOmega_k\bJ_{k_q} & \\
  & \vdots & & \vdots & & \vdots & \\
  & \bJ_{k_h}^T\bOmega_k\bJ_{k_1} & \cdots & \bJ_{k_h}^T\bOmega_k\bJ_{k_h} &
  \cdots & \bJ_{k_h}^T\bOmega_k\bJ_{k_q} & \\
  & \vdots & & \vdots & & \vdots & \\
  & \bJ_{k_q}^T\bOmega_k\bJ_{k_1} & \cdots & \bJ_{k_q}^T\bOmega_k\bJ_{k_h} &
  \cdots & \bJ_{k_q}^T\bOmega_k\bJ_{k_q} & \\
  &  &  &  &  &  & \cdot \\
  \end{pmatrix}.
  \end{footnotesize}
\end{equation*}
Each measurement introduces a finite number of non-diagonal
components that depends quadratically on the number of variables that
influence the measurement.
Therefore, in the many cases when the number of measurements is proportional to the number of variables such as \gls{slam} or \gls{ba}, the system matrix $\bH$ is sparse, symmetric and positive 
semi-definite \emph{by construction}.
Exploiting these intrinsic
properties, we can efficiently solve the linear system
in~\eqref{eq:linear-solve}. In fact, the literature provides many solutions to
this kind of problem, which can be arranged in two main groups:
(i) iterative methods and
(ii) direct methods.  The former computes an iterative solution to the
linear system by following the gradient of the quadratic form.  These
techniques often use a \textit{pre-conditioner} \eg~\gls{pcg}, which
scales the matrix to achieve quicker convergence to take steps along
the steepest directions.  Further informations about these approaches
can be found in~\cite{saad2003iterative}. Iterative methods might
require a quadratic time in computing the exact solution of a linear system,
but they might be the only option when the dimension of the problem is
very large due to their limited memory requirements.
Direct methods, instead, compute return the exact solution of the
linear system, usually leveraging on some matrix decomposition
followed by backsubstitution.  Typical methods include: the
\textit{Cholesky} factorization~$\bH = \mathbf{L}\mathbf{L}^T$ or the
\textit{QR-decomposition}.  A crucial parameter controlling the
performances of a sparse direct linear solver is the fill-in, that is
the number of new non-zero elements introduced by the specific
factorization. A key aspect in reducing the fill in is the ordering of
the variables. Since computing the optimal reordering is NP-hard,
approximated
techniques~\cite{amestoy1996approximate,davis2004column,karypis1998multilevelk}
are generally employed. Fig.~\ref{fig:reordering} shows the effect of
different variable ordering on the \emph{same} system matrix. A larger
fill-in results in more demanding computations.  We refer
to~\cite{davis2006direct} and \cite{agarwal2012variable} for a more
detailed analysis about this topic.

\subsection{A unifying formalism: Factor Graphs}\label{sec:factor-graphs}
\begin{figure}[!t]
  \centering
  \includegraphics[width=\columnwidth]{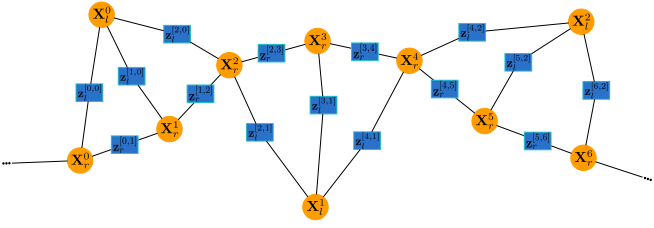}
  \caption{
    Illustration of a generic factor graph. Orange round nodes depict state 
    variables $\bx_{0:N}$. Each blue squared nodes, instead, represents a 
    factor $p(\bz|\bx_{k_s}, \dots, \bx_{k_q})$; edges denote the
    variables in conditionals of a factor $\{\bx_{k_s}, \dots, \bx_{k_q}\}$.
  }
  \label{fig:factor-graph}
\end{figure}

In this section, we introduce a formalism to represent a super-class
of the minimization problems discussed so far: \emph{factor graphs}.
We recall that 
i) our state $\bx=\{ \bx_{1:N} \}$ is composed by $N$ variables,
ii) the conditional probabilities $p(\bz_k|\bx)$ might depend only
by a subset of the state variables $\bx_k=\{ \bx_{k_1}, \bx_{k_2}, \dots, \bx_{k_q} 
\} \in \bx$ and
iii) we have no prior about the state $p(\bx)=\mathcal U (\bx)$.
Given this, we can expand~\eqref{eq:bayes} as follows:
\begin{equation}
  p(\bx|\bz) \propto \prod_{k=1}^K p(\bz_k|\bx_k)
  = \prod_{k=1}^K p(\bz_k|\bx_{k_1}, \bx_{k_2}, \dots, \bx_{k_q}).
\label{eq:factor-graph-vars}
\end{equation}
\eqref{eq:factor-graph-vars} expresses the likelihood of the
measurements as a product of factors. 
This concept is elegantly captured in the \textit{factor graph} formalism, 
which provides a graphical representation for
this kind of problems.
%
A factor graph is a bipartite graph where each node represents either
a variable $\bx_i \in \bx$, or a \emph{factor}
$p(\bz_k|\bx_k)$.  
\figref{fig:factor-graph} illustrates an example of factor graph.
Edges connect a factor $p(\bz_k|\bx_k)$ with each of
its variables $\bx_k = \{\bx_{k_1}, \bx_{k_2}, \dots, \bx_{k_q}\}$.
In the remainder of this document, we will
stick to the factor graph notation and we will refer to the
measurement likelihoods $p(\bz_k|\bx_k)$ as factors.
Note that, the main difference between~\eqref{eq:factor-graph-vars}
and~\eqref{eq:multiple_meas} is that the former highlights the subset of
state variables $\bx_k$ from which the observation depends, while the latter 
considers \textit{all} state variables - also those that have a null 
contribution.

The aim of this section is to use the factor graph formulation to formalize the 
\gls{ils} minimization exposed so far. In this sense,~\algref{alg:gn-manifolds} 
reports a step-by-step expansion of the vanilla \gls{gn} algorithm exploiting 
the factor graph formalism - supposing that both states and measurement belong 
to a smooth manifold. 
In the remainder of this document, we indicate with bold uppercase symbols 
elements lying on a manifold space - \eg~$\bX \in \mathrm{SE}(3)$; lowercase 
bold symbols specify their corresponding vector perturbations 
- \eg~$\bDeltax \in \mathbb{R}^n$.
%
%
Going more in detail, at each iteration the algorithm re-initializes its 
workspace to store the current linear system (lines  
\ref{line:initialize-start}-\ref{line:initialize-end}).
Subsequently, it processes each measurement $\bZ_k$ (line
\ref{line:measurement-loop}), computing 
i) the prediction (line~\ref{line:prediction}),
ii) the error vector (line~\ref{line:error}) and 
iii) the coefficients to apply the robustifier 
(lines~\ref{line:l1}-\ref{line:adapt}).
While processing a measurement, it also computes the blocks $\bJ_{k,i}$ of the 
Jacobians with respect to the variables $\bX_i \in \bX_k$ involved in the 
factor.  
We denote with $\bH_{i,j}$ the block $i,j$ of the $\bH$ matrix
corresponding to the variables $\bX_i$ and $\bX_j$; similarly we indicate with 
$\mathbf{b}_{i}$ the block of the coefficient vector for the
variable $\bX_i$.
This operation is carried on in the
lines~\ref{line:variable-nested-loop}-\ref{line:b-update}. 
The contribution of each measurement to the linear system is added 
in a block fashion. 
Further efficiency can be achieved exploiting the symmetry of the system matrix 
$\bH$, computing only its lower triangular part.
Finally, once the linear system $\bH\bDeltax = -\bb$ has been built, it is 
solved using a general sparse linear solver (line~\ref{line:solve-system}) and 
the perturbation $\bDeltax$ is applied to the current state in a block-wise 
fashion (line~\ref{line:apply-perturbation}).
The algorithm proceeds until convergence is reached, namely when the delta of 
the cost-function $F$ between two consecutive iteration is lower than a 
threshold $\epsilon$ - line~\ref{line:convergence-loop}.
\begin{algorithm}[!t]
  \caption{Gauss-Newton minimization algorithm for manifold
    measurements and state spaces}
  \setalglinespace
  \begin{algorithmic}[1]
  \Require{
    Initial guess $\breve\bX$;
    Measurements $\mathcal{C} = \{\langle\bZ_k, \bOmega_k\rangle\}$
  }
  \Ensure{Optimal solution $\bX^\star$}
  \State $\Fold \leftarrow \inf $
  \State $\Fnew \leftarrow 0 $
  \While{$\Fold - \Fnew > \epsilon$} \label{line:convergence-loop}
    \State $\Fold \leftarrow \Fnew$
    \State $\Fnew \leftarrow 0$
    \State $\bb \leftarrow 0$ \label{line:initialize-start}
    \State $\bH \leftarrow 0$ \label{line:initialize-end}
    
    \ForAll{$k \in \{1\, \dots\, K\}$} \label{line:measurement-loop}
      \State $\hat \bZ_k \leftarrow \bh_k(\breve\bX_k)$         \label{line:prediction} 
      \State $\be_k \leftarrow  \hat \bZ_k \boxminus \bZ_k$     \label{line:error}
      \State $\chi_k \leftarrow  \be_k^T \bOmega_k \be_k$
      \State $\Fnew \leftarrow  \Fnew + \chi_k$
      \State $ u_k \leftarrow  \sqrt {\chi_k}$ \label{line:l1}
      \State $\gamma_k = \frac{1}{u_k} \left. \frac{\partial \rho_k(u)}{\partial u} \right|_{u=u_k}.$ \label{line:robust}
      \State $\tilde \bOmega_k = \gamma_k \bOmega_k $ \label{line:adapt}
      
      \ForAll{$ \bX_i \in \{\bX_{k_1}\, ...\, \bX_{k_q}\}$} \label{line:variable-loop}
      \State $\tilde{\bJ}_{k,i} \leftarrow 
       \left. \frac{\partial \bh_k(\bX \boxplus \bDeltax) \boxminus 
       \bZ_k}{\partial \bDeltax_i} \right|_{\bDeltax=\bZero}$ 
       \label{line:jacobian}
      
      \ForAll{$ \bX_j \in \{ \bX_{k_1}\, ...\, \bX_{k_q}\} \mbox{ and } j<=i$} \label{line:variable-nested-loop} 
        \State $\bH_{i,j} \leftarrow \bH_{i,j} + \bJ_{k,i}^\top \tilde \bOmega 
        _k\bJ_{k,j}$ \label{line:h-update}
        \State $\mathbf{b}_i \leftarrow \mathbf{b}_i +  \bJ_{k,i}^\top \tilde 
        \bOmega_i \be_k$ \label{line:b-update}
      \MyEndFor
      \MyEndFor
    \MyEndFor
    
    \State $\bDeltax \leftarrow \mathrm{solve}(\bH\bDeltax = -\mathbf{b})$ 
    \label{line:solve-system}
    
    \ForAll{$\bX_i \in \bX$}
      \State $\breve\bX_i \leftarrow \breve\bX_i \boxplus \bDeltax_i $    
      \label{line:apply-perturbation}
    \MyEndFor
  \MyEndWhile
  \State
  \Return $\breve\bX$
  \end{algorithmic}
  \label{alg:gn-manifolds}
\end{algorithm}

Summarizing, instantiating \algref{alg:gn-manifolds} on a specific problem 
requires to:
\begin{itemize}
\item[--] Define for each type of variable $\bx_i \in \bx$ 
  i) an extended parametrization $\bX_i$, 
  ii) a vector perturbation $\bDeltax_i$ and 
  iii) a $\boxplus$ operator that computes a new point on the manifold
  $\bX_i'=\bX_i \boxplus \bDeltax_i$. 
  If the variable is Euclidean, the extended and the increment parametrization 
  match and, thus, $\boxplus$ degenerates to vector addition.
\item[--] For each type of factor $p(\bz_k|\bx_k)$, specify
  i) an extended parametrization $\bZ_i$,
  ii) an Euclidean representation $\bDeltaz_i$ for the error vector and
  ii) a $\boxminus$ operator such that, given two points on the manifold 
  $\bZ_i$ and $\bZ'_i$, $\bDeltaz_i=\bZ'_i \boxminus \bZ_i$ represents the 
  motion on the chart that moves $\bZ_i$ onto $\bZ'_i$.
  If the measurement is Euclidean, the extended and perturbation 
  parametrizations match and, thus, $\boxminus$ becomes a simple vector 
  difference. Finally, it is necessary to define the measurement function 
  $\bh_k(\bX_k)$, that given a subset of state variables 
  $\bX_k$, computes the expected measurement $\hat \bZ_k$.
\item[--] Choose a robustifier function $\rho_k(u)$ for each type of
  factor. The non-robust case is captured by choosing 
  $\rho_k(u)=\frac{1}{2}u^2$.
\end{itemize}
%
%
Note that, depending on the choices and on the application, not all
these steps indicated here are required. 
Furthermore, the value of some variables might be known beforehand - \eg~the 
initial position of the robot in \gls{slam} is typically set at the origin.
Hence, these variables do not need to be estimated in the optimization process, 
since they are \emph{constants} in this context.  
In~\algref{alg:gn-manifolds}, fixed variables can be handled in the solution 
step - \ie~line~\ref{line:solve-system} - suppressing all block rows and 
columns in the linear system that arise from these special nodes.
In the next section, we present how to formalize several common \gls{slam} 
problems through the factor graph formalization introduced so far.

\section{Examples}
\label{sec:use-cases-examples}
In this section we present examples on how to apply the methodology
illustrated in \secref{sec:factor-graphs} to typical problems in
robotics, namely: Point-Cloud Registration, Projective Registration,
\gls{ba} and \gls{pgo}.

\subsection{ICP}\label{sec:icp}
\begin{figure}[!t]
  \centering
  \includegraphics[width=0.9\linewidth]{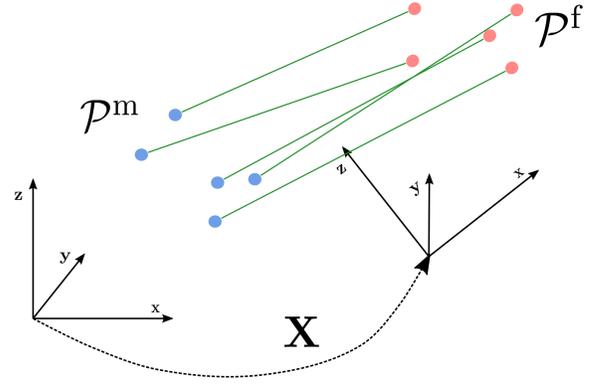}
  \caption{Registration of two point cloud through \gls{icp}. The red points 
  represent entries of the \emph{fixed} cloud, while the blue points belong to 
  the \emph{moving} one. Green lines emphasize the associations between points 
  belonging to the two clouds.}
  \label{fig:icp}
\end{figure}

\gls{icp} represents a family of algorithms
used to compute a transform that maximizes the overlap between two 
point clouds. Let $\mathcal{P}^\mathrm{f}$ be the 
cloud that stays \emph{fixed} and $\mathcal{P}^\mathrm{m}$ be the one that is 
\emph{moved} to maximize the overlap. 
\gls{icp} algorithms achieve this goal progressively refining an 
initial guess of the target transformation $\bX$ by alternating two phases: 
data-association, and optimization. 
The aim of the data-association is to find a point
$\bp^\mathrm{f}_i \in \mathcal P^\mathrm{f}$
that is likely to be the same as the point $\bp^\mathrm{m}_j \in \mathcal 
P^\mathrm{m}$ being transformed according to $\bX$, see \figref{fig:icp}. 
%
Note that, several heuristics to determine the data association have been
proposed by the research community, depending on the context of the problem. 
The most common one are either geometry based - \ie~nearest neighbor, normal 
shooting, projective association - or rely on appearance-based evaluations.
Discussing data-association strategies is
out of the scope fo this work, still, we can generalize the
\emph{outcome} of data association by a selector function
$\mathrm{j}(k) \in \{1,\dots,|\mathcal P^\mathrm{f}| \}$ that maps a point
index $k$ in the moving cloud to an index $j$ in the fixed cloud. In
this way, we indicate a pair of corresponding points as 
$\langle \bp^\mathrm{m}_k, \bp^\mathrm{f}_{\mathrm{j}(k)}\rangle$.  
In contrast to data-association, the optimization step is naturally described 
as an \gls{ils} problem.  
The variable to be estimated is a transform $\bX$ whose domain depends on the 
specific scenario - \eg~$\mathrm{SE}(2)$, $\mathrm{SE}(3)$ or even a Similarity 
if the two clouds are at different scales. In the remaining of this section, we 
will use $\bX \in \se3$ to instantiate our factor-graph-based \gls{ils} problem.

\subsubsection{Variables}\label{sec:se3-var}
Since the transformation we should estimate is a 3D Isometry 
$\bX \in \se3$, our state lies on a smooth manifold. Therefore we should define 
all the entities specified in~\secref{sec:factor-graphs}, namely:
\begin{itemize}
\item[--] Extended Parameterization: 
  we conveniently define a transformation 
  $\bX=[\bR\;|\;\bt] \in \se3$ as a rotation matrix 
  $\bR$ and a translation vector $\bt$.
  Using this notation, the following relations hold:
  \begin{align}
    \bX_{12} = \bX_1 \bX_2 &\triangleq 
      \begin{bmatrix} \bR_1 \bR_2 & \bt_1 + \bR_1\bt_2 \end{bmatrix} 
      \label{eq-icp-se3-multiplication} \\
    \bX^{-1} &\triangleq \begin{bmatrix} \bR^\top & -\bR^\top\bt \end{bmatrix}.
    \label{eq-icp-se3-inversion}
  \end{align}
\item[--] Perturbation Vector: 
  a commonly used vector parametrization is 
  $\bDeltax^\top =[ \bDeltat^\top \; \bDeltaa^\top] \in \mathbb{R}^6$, where 
  $\bDeltat \in \mathbb{R}^3$ represents a translation, while $\bDeltaa \in 
  \mathbb{R}^3$ is a minimal representation for the rotation. 
  The latter might use Euler angles, unit-quaternion or the logarithm of the 
  rotation matrix.
\item[--] $\bX \boxplus \bDeltax$ Operator: 
  this is straightforwardly implemented by
  first computing the transformation $\bDeltaX = [ \bDelta \bR \; \bDeltat ] 
  \in \se3$ from the perturbation, and then applying such a perturbation to the 
  previous transform. In formul\ae:
  \begin{equation}
    \bX \boxplus \bDeltax = \vTot(\bDeltax) \bX
    \label{eq-icp-boxplus-se3}
  \end{equation}
  where $\vTot(\bDeltax)$ computes a transform $\bDeltaX$ from a
  perturbation vector $\bDeltax$. Its implementation depends on the
  parameters chosen for the rotation part $\bDeltaa$.
  Note that, the perturbation might be applied to the left or to the right of 
  the initial transformation. 
  In this document we will consistently apply it to the left. Finally, we 
  define also the inverse function 
  $\bDeltax = \tTov(\bDeltaX)$, that computes perturbation vector from the 
  transformation matrix such that $\bDeltax = \tTov(\vTot(\bDeltax))$.
\end{itemize}

\subsubsection{Factors}\label{sec:icp-factor}
In this problem, we have just one type of factor, which depends on
the relative position between a pair of corresponding points, after applying 
the current transformation $\bX$ to the moving cloud. Given a set 
of associations 
$\{\langle \bp^\mathrm{m}_s, \bp^\mathrm{f}_{\mathrm{j}(s)}\rangle, \dots, 
\langle \bp^\mathrm{m}_K, \bp^\mathrm{f}_{\mathrm{j}(K)}\rangle\}$, each fixed 
point $\bp^\mathrm{f}_{\mathrm{j}(k)}$ constitutes a measurement $\bz_k$ - 
since its value does not change during optimization. On the contrary, each 
moving point $\bp^\mathrm{m}_k$ will be used to generate the prediction 
$\hat{\bz}$. Note that, the measurement space is Euclidean in ths scenario - 
\ie~$\mathbb{R}^3$. Therefore, we only need to define the following entities:
%
%
%
\begin{itemize}
  \item[--] Measurement Function:
  it computes the position of a point $\bp^\mathrm{m}_{k}$ that corresponds to 
  the point $\bp^\mathrm{f}_{\mathrm{j}(k)}$ in fixed scene by applying the 
  transformation $\bX$, namely:
  \begin{equation}
    \label{eq:icp-measurement-se3}
    \bh^\mathrm{icp}_k(\bX) \triangleq 
      \bX^{-1} \bp^\mathrm{m}_{k} = \bR^\top (\bp^\mathrm{m}_{\mathrm{j}(k)} - 
      \bt)
  \end{equation}
  \item[--] Error Function: since both prediction and measurement are 
  Euclidean, the $\boxminus$ operator boils down to simple vector difference. 
  The error, thus, is a 3-dimensional vector computed as:
  \begin{equation}
    \label{eq:icp-error-se3}
    \be^\mathrm{icp}_k(\bX) = \bh_k(\bX) - \bp^\mathrm{f}_{\mathrm{j}(k)}.
  \end{equation}
\end{itemize}
The Jacobians can be computed analytically very straightforwardly 
from~\eqref{eq:icp-error-se3} as:
\begin{equation}
  \label{eq:icp-jacobians-se3}
  \bJ^\mathrm{icp}(\bX, \bp) =
  \frac
  {\partial \left(\bX \boxplus \bDeltax \right)^{-1} \bp}
  {\partial \bDeltax}
  \Big\rvert_{\bDeltax = \bZero}.
\end{equation}
%
With this in place, we can now fully instantiate
\algref{alg:gn-manifolds}. For completeness, in the appendix we report
the functions $\vTot(\cdot)$ and $\tTov(\cdot)$ for $\se3$ objects, together 
with the analytical derivation of the Jacobians.
Since in this case the measurement is Euclidean, the Jacobians of
error function and measurement function are the same.

\subsection{Projective Registration}\label{sec:pr}
\begin{figure}[!t]
  \centering
  \includegraphics[width=\linewidth]{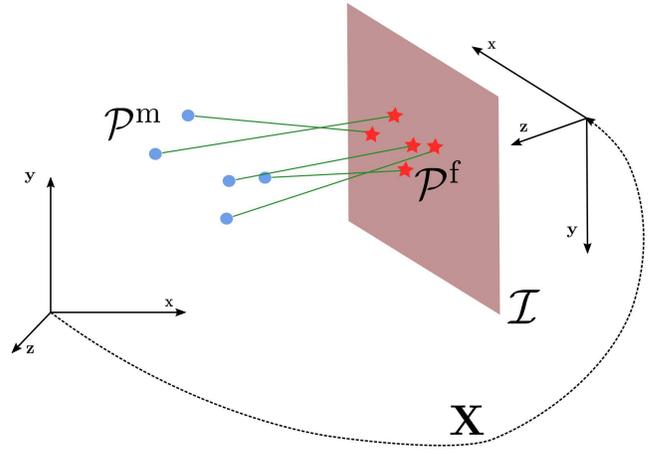}
  \caption{Projective Registration scenario: $\mathcal{I}$ represents the image
    plane; blue points represent 3D entries of the \emph{moving} cloud, while 
    red stars indicate the projection of each corresponding 3D point onto 
    $\mathcal{I}$. Finally, green lines emphasize the associations between 
    moving cloud and fixed image projections.}
  \label{fig:picp}
\end{figure}

Projective Registration consists in determining the pose $\bX$ of a
camera in a known 3D scene from a set of 2D projections of
these points on the image plane. In this case our \emph{fixed} point cloud 
$\mathcal{P}^{\mathrm{f}}$ will be consisting of the image projections, while 
the \emph{moving} one $\mathcal{P}^{\mathrm{m}}$ will be composed by the known 
location of the 3D points, see \figref{fig:picp}.  
We use the notation for data-association defined in~\secref{sec:icp}, in 
which the function $\mathrm{j}(i)$ retrieves the index of a 2D measurement on 
the image that corresponds to the 3D point 
$\bp_i^{\mathrm{m}} \in \mathcal{P}^{\mathrm{m}}$.  
Also in this scenario, the only variable to estimate is the transformation $\bX 
\in \se3$, therefore we will simply re-use the entities defined 
in~\secref{sec:se3-var} and focus only on the factors.

\subsubsection{Factors}\label{sec:proj-factors}
Given a set of 2D-3D associations
$\{\langle \bp^\mathrm{m}_s, \bp^\mathrm{f}_{\mathrm{j}(s)}\rangle, \dots, 
\langle \bp^\mathrm{m}_K, \bp^\mathrm{f}_{\mathrm{j}(K)}\rangle\}$, each
fixed point $\bp^\mathrm{m}_k \in \mathbb{R}^2$ will represent a measurement 
$\bz_k$, each moving point will contribute to the prediction $\hat\bz_k$. 
Therefore we can define:
\begin{itemize}
\item[--] Measurement Function: 
  it is the projection on the image plane of a
  scene point $\bp^\mathrm{m}_k$, assuming the camera is at $\bX$. Such
  a prediction is obtained by first mapping the point in the camera
  reference frame to get a new point $\bp^\mathrm{icp}$, and then
  projecting this point on the image plane, in formlu\ae:
  \begin{align}
    \bp^\mathrm{icp} &\triangleq \bX^{-1} \bp^\mathrm{m} \label{eq:pr-pt-se3}\\
    \bp^\mathrm{cam} &\triangleq \bK \bp^\mathrm{icp} \label{eq:pr-pp-se3}\\
    \bp^\mathrm{img} &\triangleq \mathrm{hom}(\bp^\mathrm{cam}) = 
      \begin{pmatrix} 
        p^{\mathrm{cam}}_x / p^{\mathrm{cam}}_z \\  
        p^{\mathrm{cam}}_y / p^{\mathrm{cam}}_z
      \end{pmatrix}.
    \label{eq:pr-pcam-se3}
  \end{align}
  Note that, $\bp^\mathrm{cam}$ is the point in homogeneous image
  coordinates, while $\bp^\mathrm{img}$ is the 2D point in pixel coordinates 
  obtained normalizing the point through homogeneous division. Finally, the 
  complete measurement function is defined as:
  \begin{equation}
    \bh^\mathrm{reg}_k(\bX) \triangleq \mathrm{hom}(\bK \bX^{-1}     
    \bp^\mathrm{m}_{\mathrm{j}(k)}) = 
    \mathrm{hom}(\bK [\bh^{icp}(\bp^\mathrm{m}_{\mathrm{j}(k)})]).
    \label{eq:pr-prediction}
  \end{equation}
\item[--] Error Function: 
also in this case, both measurement and prediction are Euclidean vectors and, 
thus, we can use the vector difference to compute the 2-dimensional error as 
follows:
  \begin{equation}
    \be^\mathrm{reg}_{k}(\bX) = \bh^\mathrm{reg}_{k}(\bX)- \bz_k
    \label{eq:pr-error}
  \end{equation}
\end{itemize}
Note that, we can exploit the work done in~\secref{sec:icp-factor} to easily 
compute Jacobians using the chain-rule, namely:
\begin{align}
  \bJ^\mathrm{reg}(\bX, \bp) &=
  \overbrace{
    \left.
    \frac
        {\partial \mathrm{hom}(\bv)}
        {\partial \bv}
        \right|_{\bv = \bp^\mathrm{cam}}
  }^{\bJ^\mathrm{hom} (\bp^\mathrm{cam})}
  \bK \bJ^{ICP} (\bX, \bp) \nonumber \\
  &=\bJ^\mathrm{hom} (\bp^\mathrm{cam}) \bK \bJ^{ICP} (\bX, \bp).
  \label{eq:pr-jacobians-chain-rule}
\end{align}

\subsection{Structure from Motion and Bundle Adjustment}\label{sec:sfm-ba}
\begin{figure} [!t]
  \centering
  \includegraphics[width=\linewidth]{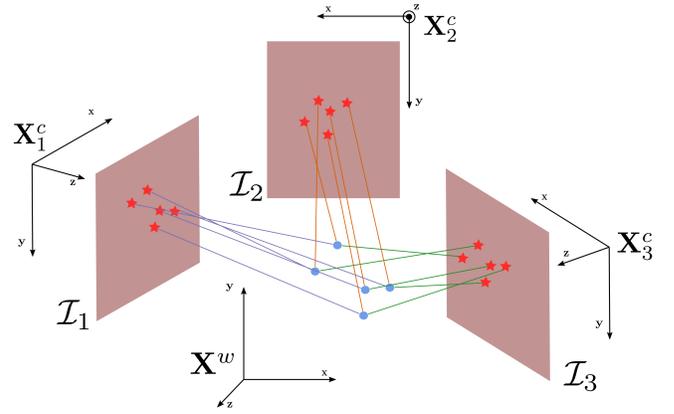}
  \caption{\gls{sfm} scenario: blue dots represent 3D point in the world, while 
    red dots indicate their projection onto a specific image plane 
    $\mathcal{I}_k$. Colored lines emphasize the data association between 3D 
    points and their corresponding image projections.}
  \label{fig:bundle-adj}
\end{figure}

\gls{sfm} is the problem of determining the
pose of $N$ cameras and the position of $M$ 3D points on a scene,
from their image projections. The scenario is shown in ~\figref{fig:bundle-adj}. This problem is highly non-convex, and
tackling it with \gls{ils} requires to start from an initial guess not too
far from the optimum.  Such a guess is usually obtained by using
Projective Geometry techniques to determine an initial layout of the
camera poses. Subsequently, the points are triangulated to
initialize all variables.  A final step of the algorithm consists in
performing a non-linear refinement of such an initial guess - known as \gls{ba} 
- which is traditionally approached as an \gls{ils} problem. Since typically 
each camera observes only a subset of
points, and a point projection depends only on the relative pose
between the observed point and the observing camers, \gls{ba} is a good
example of a sparse problem.

\subsubsection{Variables}\label{sec:ba-vars}
We want to estimate the pose of each camera $\bX^\mathrm{c}_{1:N}$,
\emph{and} the position of each point $\bx^\mathrm{p}_{1:M}$.
The state will thus be a data structure
$\bX = \langle \bX^\mathrm{c}_{1:N}, \bx^\mathrm{p}_{1:M} \rangle$ storing
all camera poses and all points. Given this, for the camera poses 
$\bX^\mathrm{c}_{1:N}$, the definitions in~\secref{sec:se3-var} will be used 
again. As for the points, we do not need a specific extended parametrization, 
since they lie on $\Re^3$. Therefore we should define only:
\begin{itemize}
  \item[--] Perturbation Vector: the total perturbation vector is defined as
  \begin{equation*}
    \bDeltax^\top = 
    \begin{pmatrix}
      \bDeltax^{\mathrm{c}\top}_1 & \cdot\cdot & \bDeltax^{\mathrm{c}\top}_N & 
      | & 
      \bDeltax^{\mathrm{p}\top}_1 & \cdot\cdot & \bDeltax^{\mathrm{p}\top}_M
    \end{pmatrix}.
  \end{equation*}
  Otherwise speaking, it is a $6N+3M$ vector obtained by stacking the 
  individual perturbations.
  \item[--] $\bX \boxplus \bDeltax$ Operator: the operator will use the same 
  machinery introduced in~\secref{sec:se3-var} for the poses and the standard 
  Euclidean addition for the point positions.
\end{itemize}

\subsubsection{Factors}\label{sec:ba-factors}
Similar to Projective Registration, in \gls{ba} a measurement $\bz_k$ is a
projection of a point on the 2D image plane. However, in this specific 
scenario, such a projection depends not only on the estimate of a camera pose 
but also on the estimate of the point. Note that, this information was known in the
case of Projective Registration, while now it becomes part of the state.  
For consistency with \algref{alg:gn-manifolds}, if the $k^\mathrm{th}$
measurement arises from observing the point $\bx^\mathrm{p}_m$ with
the camera $\bX^\mathrm{c}_n$, we will denote these indices with two
selector functions $n=\mathrm{n}(k)$ and $m=\mathrm{m}(k)$, that map
the factor index $k$ respectively to the indices of the observing
camera and the observed point.  For the $k^\mathrm{th}$ factor, the
camera and point variables will then be
$\bx^\mathrm{p}_{\mathrm{m}(k)}$ and $\bX^\mathrm{c}_{\mathrm{n}(k)}$.
Note that, also in this case, a measurement $\bz_k$ lies in an Euclidean space 
- \ie~$\mathbb{R}^2$. Given this, to instantiate a factor we define:
\begin{itemize}
\item[--] Measurement Function: 
  the prediction $\hat\bz_k$ can be easily obtained from 
  \eqref{eq:pr-prediction}, namely:
  \begin{align}
    \bh^\mathrm{ba}_{k}(\bX) &\triangleq 
    \bh^{ba}_{k}(\bX^\mathrm{c}_{\mathrm{n}(k)}, \bx^\mathrm{p}_{\mathrm{m}(k)})
    \nonumber \\
    &=\mathrm{hom}\left(\bK ({\bX^\mathrm{c}_{\mathrm{n}(k)}}^{-1}) 
    \bx^\mathrm{p}_{\mathrm{m}(k)}\right).
    \label{eq:ba-measurement-function}
  \end{align}
\item[--] Error Function: 
it is the Euclidean  difference between prediction and measurement:
  \begin{equation}
    \label{eq:ba-error}
    \be^{ba}_{k}(\bX^\mathrm{c}_{\mathrm{n}(k)}, 
    \bx^\mathrm{p}_{\mathrm{m}(k)}) \triangleq
    \bh^{ba}_{k}\left(\bX^\mathrm{c}_{\mathrm{n}(k)}, 
    \bx^\mathrm{p}_{\mathrm{m}(k)}\right)-\bz_k .
  \end{equation}
\end{itemize}
In this context, the Jacobian $\bJ_k^\mathrm{ba}(\bX)$ will be consisting of two
blocks, corresponding to the perturbation of the camera pose and to
the perturbation of the point position, in formul\ae:
\begin{align}
  \bJ^\mathrm{ba}_k &= 
  \begin{pmatrix}  
    \bZero & \cdot & \bZero &
    \bJ^\mathrm{ba}_{k,\mathrm{n}(k)} & 
    \bZero & \cdot & \bZero &
    \bJ^\mathrm{ba}_{k,\mathrm{m}(k)} & 
    \bZero & \cdot & \bZero
  \end{pmatrix} 
  \label{eq:ba-jac-generic}
\end{align}
where
\begin{align}
  \bJ^\mathrm{ba}_{k,\mathrm{n}(k)}(\bX) &=
  \left.
  \frac {\partial \be^\mathrm{ba}_k(\bX^\mathrm{c}_{\mathrm{n}(k)} \boxplus 
  \bDeltax^\mathrm{r}, \bx^\mathrm{p}_{\mathrm{m}(k)})}
        {\partial \bDeltax^\mathrm{c}}
        \right|_{\bDeltax^\mathrm{c}=\bZero}
        \\
  \bJ^\mathrm{ba}_{k,\mathrm{m}(k)}(\bX) &=
  \left.
  \frac {\partial \be^\mathrm{ba}_k(\bX^\mathrm{r}_{\mathrm{n}(k)}, 
  \bx^\mathrm{p}_{\mathrm{m}(k)} \boxplus \bDeltax^\mathrm{p})}
        {\partial \bDeltax^\mathrm{p}}
        \right|_{\bDeltax^\mathrm{p}=\bZero}.
\end{align}
%
Again, the measurement domain is Euclidean, thus the Jacobians of the error
function and the measurement function match.
For completeness, in the Appendix of this document we report a more in-depth 
derivation of the Jacobians.

Still, since all measurements are relative, given a particular
solution $\bX^\star$ all solutions $\bX'= \bT \bX^\star$ obtained by applying
a transformation $\bT \in \se3$ to \emph{all} the variables in $\bX^\star$
have the same residual \chisquared and, thus, are equivalent.  Furthermore, all
solutions $\bX' = s \bX^\star$ obtained by scaling all poses and landmarks
by a constant $s$ are equivalent too. This reflects the fact that
observing an object that is twice as big from twice the distance
results in the same projection.  Thence, the problem of \gls{ba} is 
under-determined by 7 \gls{dof} and, thus, the vanilla \gls{gn} algorithm 
requires to fix at least 7 \gls{dof} - typically a camera pose (6 \gls{dof}), 
and the distance between two points or two camera poses (1 \gls{dof}).

\subsection{Pose Graphs}\label{sec:examples-pgo}

A pose graph is a factor graph whose variables $\bX^\mathrm{r}_{1:N}$
are poses and whose measurements $\bZ_{1:K}$ are relative measurements
between pairs of poses. 
Optimizing a pose graph means determining the configuration of poses that is 
maximally consistent with the measurements.
\gls{pgo} is very common in the \gls{slam} community, 
and several ad-hoc approaches have been proposed.
Similar to \gls{ba}, \gls{pgo} is highly non-convex, and its solution with 
\gls{ils} requires a reasonably good initial guess.

\subsubsection{Variables}\label{sec:examples-pgo-variables}
Also in \gls{pgo}, each variable $\bX^\mathrm{r}_k$ lies on the smooth 
manifold $\se3$. Once again, we will make use of the formulation used 
in~\secref{sec:se3-var} to characterize the state $\bX=\bX^\mathrm{r}_{1:N}$ 
and the perturbation vector
$\bDeltax^{\mathrm{r}T} =
(\bDeltax^{\mathrm{r}T}_1 \; \dots \; \bDeltax^{\mathrm{r}T}_N)$.
%

\subsubsection{Factors}\label{sec:examples-pgo-factors}
Using the same index notation in \secref{sec:ba-factors}, let $\bZ_k$
be the $k^\mathrm{th}$ relative pose measurement expressing the pose
$\bX_m$ in the reference frame of the pose $\bX_n$. We denote the pair
of poses as $\bX_n = \bX_{\mathrm{n}(k)}$, and $\bX_m =
\bX_{\mathrm{m}(k)}$ using the two selector functions $\mathrm{n}(k)$ and
$\mathrm{m}(k)$.
In this scenario, a measurement $\bZ_k$ expresses a relative pose between two 
variables and, consequently, also $\bZ_k$ lies on the smooth manifold $\se3$. 
Considering this, we define the following entities:
\begin{itemize}
\item[--] Measurement Function:
  this is straightforwardly obtained by expressing the observed pose 
  $\bX^\mathrm{r}_{\mathrm{m}(k)}$ in the
  reference frame of the observing pose $\bX^\mathrm{r}_{\mathrm{n}(k)}$, 
  namely:
  \begin{align}
    \bh^\mathrm{pgo}_{k}(\bX) &\triangleq 
    \bh^\mathrm{pgo}_{k}\left(\bX^\mathrm{r}_{\mathrm{n}(k)}, 
    \bX^\mathrm{r}_{\mathrm{m}(k)}\right) \nonumber \\
    &=
    ({\bX^\mathrm{r}_{\mathrm{n}(k)}})^{-1} \; \bX^\mathrm{r}_{\mathrm{m}(k)}.
    \label{eq:pgo-prediction}
  \end{align}
\item[--] Error Function: 
in this case, since the measurements are non-Euclidean too, we are required to 
specify a suitable parametrization for the error vector $\be_k$. In literature, 
many error vectorization are available~\cite{aloise2019chordal}, each one with 
different properties. Still, in this document, we will make use of the same 
6-dimensional parametrization used for the increments - 
\ie~$\be^\top = (\be^{xyz\top} \; \be^{rpy\top})$. Furthermore, we need 
to define a proper $\boxminus$ operator that expresses on a chart the
relative pose between two $\se3$ objects $\bDeltaz=\hat\bZ \boxminus \bZ$. To 
achieve this goal, we i) express $\hat\bZ$ in the reference system of 
$\bZ$ obtaining the relative transformation $\bDeltaZ$ and then ii) compute 
the chart coordinates of $\bDeltaZ$ around the origin using the 
$\tTov(\cdot)$ function. In formlu\ae:
\begin{align}
  \bDeltaz \triangleq \hat\bZ \boxminus \bZ 
    = \tTov(\bDeltaZ) = \tTov\left( \bZ^{-1} \; \hat\bZ \right).
  \label{eq:se3-boxminus}
\end{align}
Note that, since $\bDeltaZ_k$ expresses a relative 
motion between prediction and measurement, its rotational component will by 
away from singularities. With this in place, the error vector is computed as 
the pose of the prediction  
$\hat\bZ_k=\bh^\mathrm{pgo}_{k}(\bX)$ on a chart centered in $\bZ_k$; namely:
  \begin{align}
    \be^\mathrm{pgo}_{k}(\bX) &\triangleq 
    \bh^\mathrm{pgo}_{k}(\bX^\mathrm{r}_{\mathrm{n}(k)}, 
    \bX^\mathrm{r}_{\mathrm{m}(k)}) \boxminus \bZ_k \nonumber \\  
    &= 
    \left( 
      ({\bX^\mathrm{r}_{\mathrm{n}(k)}})^{-1} \; \bX^\mathrm{r}_{\mathrm{m}(k)} 
    \right)
    \boxminus \bZ_k.
    \label{eq:pgo-error}
  \end{align}
\end{itemize}
Similar to the \gls{ba} case, in \gls{pgo} the Jacobian 
$\bJ^\mathrm{pgo}_k(\bX)$
will be consisting of two blocks, corresponding to the
perturbation of observed and the observing poses. 
The measurements in this case are non-Euclidean, and, thus, we need to compute 
the Jacobians on the error function - as specified in~\eqref{eq:pgo-error}:
\begin{align}
  \bJ^\mathrm{pgo}_k &= 
  \begin{pmatrix}
    \bZero & \cdot & \bZero & 
    \bJ^\mathrm{pgo}_{k,\mathrm{n}(k)} & 
    \bZero & \cdot & \bZero & 
    \bJ^\mathrm{pgo}_{k,\mathrm{m}(k)} & 
    \bZero & \cdot & \bZero
  \end{pmatrix}
  \label{eq:pgo-jacobians-general}
\end{align}
where
\begin{align}
  \bJ^\mathrm{pgo}_{k,\mathrm{n}(k)}(\bX) &=
  \left.
  \frac {\partial \be^\mathrm{pgo}_k(\bX^\mathrm{r}_{\mathrm{n}(k)} \boxplus 
  \bDeltax^\mathrm{r}, \bX^\mathrm{r}_{\mathrm{m}(k)})}
        {\partial \bDeltax^\mathrm{r}}
        \right|_{\bDeltax^\mathrm{r}=\bZero}
        \label{eq:pgo-jacobians-i} \\
  \bJ^\mathrm{pgo}_{k,\mathrm{m}(k)}(\bX) &=
  \left.
  \frac {\partial \be^\mathrm{pgo}_k(\bX^\mathrm{r}_{\mathrm{n}(k)}, 
  \bX^\mathrm{r}_{\mathrm{m}(k)} \boxplus \bDeltax^\mathrm{r})}
        {\partial \bDeltax^\mathrm{r}}
        \right|_{\bDeltax^\mathrm{r}=\bZero}
  \label{eq:pgo-jacobians-j}
\end{align}
Analogous to the \gls{ba} case, also in
\gls{pgo} all measurements are relative, and, hence, all solutions that are 
related by a single transformation are equivalent. 
The scale invariance, however, does not apply in this context. As a result, 
\gls{pgo} is under-determined by 6 \gls{dof} and
using \gls{gn} requires to fix at least one of the poses.

\subsection{Considerations}\label{sec:examples-considerations}
In general, one can carry on the estimation by using an arbitrary
number of heterogeneous factors and variables. As an instance, if we
want to augment a \gls{ba} problem with odometry, we can model the
additional measurements with \gls{pgo} factors connecting subsequent
poses. Similarly, if we want to solve a Projective Registration
problem where the world is observed with two cameras, and we have
guess of the orientation from an inertial sensor, we can extend the
approach presented in \secref{sec:pr} by conducting the optimization
on a common origin of the rigid sensor system, instead of the camera
position.  We will have three types of factors, one for each camera,
and one modeling the inertial measurements.

As a final remark, 
common presentations of \gls{icp}, Projective Registration and 
\gls{ba} conduct the optimization by estimating \emph{world-to-sensor} frame, 
rather than the \emph{sensor-to-world}, as we have done in this
document. This leads to a more compact formulation. This avoids
inverting the transform to compute the prediction, and results in Jacobians
are easier to compute in close form. We preferred to provide the solution for
\emph{sensor-to-world} to be consistent with the \gls{pgo} formulation.  

\section{A generic Sparse/Dense Modular Least Squares Solver} 
\label{sec:solver-design}
The methodology presented in \secref{sec:factor-graphs} outlines a
straight path to the design of an \gls{ils} optimization algorithm.
Robotic applications often require to run the system on-line, and,
thus, they need efficient implementations. When extreme performances
are needed, the ultimate strategy is to \textit{overfit} the solution
to the specific problem.  This can be done both at an algorithmic
level and at an implementation level.  To improve the algorithm, one
can leverage on the a-priori known structure of the problem, by
removing parts of the algorithm that are non needed or by exploiting
domain-specific knowledge. A typical example is when the structure of
the linear system is known in advance - \eg~in \gls{ba} - where it is
common to use specialized methods to solve the linear
system~\cite{lourakis2009sba}.


Focusing on the implementation, we reported two main
bottlenecks: the computation of the linear system $\bH \bDeltax = \bb$
and its solution.
Dense problems such as \gls{icp}, Sensor Calibration or Projective
Registration, are typically characterized by a small state space and
many factors of the same type.  In \gls{icp}, for instance, the state
contains just a single $\se3$ object - \ie~the robot pose.  Still,
this variable might be connected to hundreds of thousands of factors,
one for each point correspondence.  Between iterations, the \gls{icp}
mechanism results in these factors to change, depending on the current
status of the data association.  As a consequence, these systems spend
most of their time in \emph{constructing} the linear system, while the time
required solve it is negligible. 
Notably, applications such as Position Tracking or \gls{vo} require the system
to run at the sensor frame-rate, and each new frame might take several
\gls{ils} iterations to perform the registration.
On the contrary, sparse problems like , \gls{pgo} or large scale \gls{ba} are 
characterized by thousands of variables, and a number of factors which is 
typical in the same order of magnitude. In this context, a factor is connected 
to very few variables. As an example, in case of \gls{pgo}, a single measurement
depends only two variables that express mutually observable robot poses,
whereas the complete problem might contain a number of variables proportional 
to the length of the trajectory.
This results in a large-scale linear system,
albeit most of its coefficients are null. In these scenarios, the time spent
to \emph{solve} the linear system dominates over the time required to build it.

A typical aspect that hinders the implementation of a \gls{ils}
algorithm by a person approaching this task for the first time is
the calculation of the Jacobians.  The labor-intensive solution is to
compute them analytically, potentially with the aid of some
symbolic-manipulation package. An alternative solution is to evaluate
them numerically, by calculating the Jacobians column-by-column with
repeated evaluation of the error function around the linearization
point.  Whereas this practice might work in many situations, numerical
issues can arise when the derivation interval is not properly chosen.
A third solution is to delegate the task of evaluating the analytic
solution directly to the program, starting from the error
function. This approach is called \gls{ad} and Ceres Solver~\cite{ceres-solver} 
is the most representative system to embed this feature - later also adopted by 
other optimization frameworks.

In the remainder of this section, we first revisit and generalize
\algref{alg:gn-manifolds} to support multiple solution
strategies. Subsequently, we outline some design requirements that will
finally lead to the presentation of the overall design of our approach - 
proposed in \secref{sec:solver-architecture}.

\subsection{Revisiting the Algorithm}
\label{sec:modularization}
In the previous section, we presented the implementation of a
vanilla \gls{gn} algorithm for generic factor graphs. This simplistic
scheme suffers under high non linearities, or when the cost function
is under-determined. Over time, alternatives to \gls{gn} have been
proposed, to address these issues, such as~\gls{lm} or 
\gls{trm}~\cite{conn2000trust}. All
these algorithms present some common aspects or patterns that can be
exploited when designing an optimization system. Therefore, in this
section, we reformulate \algref{alg:gn-manifolds} to isolate different
independent sub-modules. Finally we present both the \gls{gn} and the
\gls{lm} algorithms rewritten by using these sub-modules.

In \algref{alg:robustify} we isolate the operations needed to compute the 
scaling factor $\gamma_k$ for the information matrix $\bOmega_k$, knowing the 
current $\chi^2_k$. \algref{alg:linearize} performs the calculation of the
error $\be_k$ and the Jacobian $\bJ_k$ for a factor $\langle \bZ_k,
\bOmega_k \rangle$ at the current linearization
point. \algref{alg:updateHb} applies the robustifier to a factor, and
updates the linear system. \algref{alg:updateSolution} applies the
perturbation $\bDelta x$ to the current solution $\breve \bX$ to
obtain an updated estimate.  Finally, in \algref{alg:gn-small} we
present a revised version of \algref{alg:gn-manifolds} that relies on
the modules described so far.  
In~\algref{alg:lm-small}, we provide an implementation of the \gls{lm} 
algorithm that makes use of the same core sub-algorithms used 
in~\algref{alg:gn-small}.
The \gls{lm} algorithm solves a
damped version of the system, namely $(\bH + \lambda \cdot \mathrm{diag}(\bH)) 
\bDeltax = \bb$. The magnitude of the damping factor $\lambda$ is adjusted 
depending on the current variation of the $\chi^2$. If the $\chi^2$ increases 
upon an iteration, $\lambda$ increases too.
In contrast, if the solution improves, $\lambda$ is decreased.
Variants of these two algorithms - \eg~damped \gls{gn}, that solves $(\bH
+ \lambda \bI) \bDeltax = \bb$ - can be straightforwardly
implemented by slight modification to the algorithm presented here.

\begin{algorithm}[!t]
  \caption{$\mathrm{robustify}(\chi^2_k)$ -- computes the robustification 
  coefficient  $\gamma_k$}
  \setalglinespace
  \begin{algorithmic}[1]
  \Require{
    Current $\chi_k^2$.
  }
  \Ensure{
    $\gamma_k$ computed from the actual error,
  }
  \State $ u_k \leftarrow  \sqrt {\chi_k}$ 
  \State $\gamma_k = \frac{1}{u_k} \left. \frac{\partial \rho_k(u)}{\partial u} \right|_{u=u_k}$

  \State \Return $\gamma_k$ 
  \end{algorithmic}
  \label{alg:robustify}
\end{algorithm}
\begin{algorithm}[!t]
  \caption{$\mathrm{linearize}(\breve\bX_k, \breve\bZ_k)$ -- computes the error 
  $\be_k$  and the Jacobians $\bJ_k$ at the current linearization point $\breve 
  \bX$}
  \setalglinespace
  \begin{algorithmic}[1]
  \Require{
    Initial guess $\breve\bX_k$;
    Current measurement $\breve\bZ_k$;
  }
  \Ensure{
    Error: $\be_k$;
    Jacobians $\tilde J_k$;
  }
  \State $\hat \bZ_k \leftarrow \bh_k(\breve\bX_k)$          
  \State $\be_k \leftarrow  \hat \bZ_k \boxminus \bZ_k$
  \State $\tilde \bJ_k = \{\}$
  \ForAll{$ \bX_i \in \{\bX_{k_1}\, ...\, \bX_{k_q}\}$} 
    \State $\tilde{\bJ}_{k,i} \leftarrow 
    \left. \frac{\partial \bh_k(\bX \boxplus \bDeltax) \boxminus \bZ_k}{\partial \bDeltax_i} \right|_{\bDeltax=\bZero}$
    \State $\tilde \bJ_k \leftarrow \tilde \bJ_k \cup \{ \tilde{\bJ}_{k,i} \}$  
  \MyEndFor
  
  \State \Return $<\be_k, \bJ_k>$
  \end{algorithmic}
  \label{alg:linearize}
\end{algorithm}

\begin{algorithm}[!t]
  \caption{$\mathrm{updateHb}(\bH,\bb,\breve\bX_k,\bZ_k, \bOmega_k)$ -- updates 
  linear system with a factor current linearization point $\breve \bX$, and 
  returns the $\chi^2_k$ of the factor}
  \setalglinespace
  \begin{algorithmic}[1]
  \Require{
    Initial guess $\breve\bX_k$;
    Coefficients of the linear system $\bH$ and $\bb$;
    Measurement $\langle\bZ_k, \bOmega_k\rangle$
  }
  \Ensure{
    Coefficients of the linear system after the update $\bH$ and $\bb$;
    Value of the cost function for this factor $\chi^2_k$
  }
  \State $<\be_k, \bJ_k>=\mathrm{linearize}(\breve \bX_k, \bZ_k)$
  \State $\chi^2_k \leftarrow  \be_k^T \bOmega_k \be_k$
  \State $\gamma_k = \mathrm{robustify}(\chi^2_k)$ 
  \State $\tilde \bOmega_k = \gamma_k \bOmega_k $ 
  \ForAll{$ \bX_i \in \{\bX_{k_1}\, ...\, \bX_{k_q}\}$} 
     \ForAll{$ \bX_j \in \{ \bX_{k_1}\, ...\, \bX_{k_q}\} \mbox{ and } j<=i$}  
       \State $\bH_{i,j} \leftarrow \bH_{i,j} + \bJ_{k,i}^\top \tilde \bOmega 
       _k\bJ_{k,j}$ 
       \State $\mathbf{b}_i \leftarrow \mathbf{b}_i +  \bJ_{k,i}^\top \tilde 
       \bOmega_i \be_k$ 
     \MyEndFor
   \MyEndFor
   
   \State \Return $<\chi^2_k, \bH, \bb>$
  \end{algorithmic}
  \label{alg:updateHb}
\end{algorithm}
\begin{algorithm}[!t]
  \caption{$\mathrm{updateSolution}(\breve\bX, \bDeltax)$ -- applies a 
  perturbation to the current system solution}
  \setalglinespace
  \begin{algorithmic}[1]
  \Require{Current solution $\breve\bX$; Perturbation $\bDeltax$}
  \Ensure{New solution $\breve \bX$, moved according to $\bDeltax$}
  \ForAll{$\bX_i \in \bX$}
    \State $\breve\bX_i \leftarrow \breve\bX_i \boxplus \bDeltax_i $
  \MyEndFor

  \State \Return $\breve\bX$
  \end{algorithmic}
  \label{alg:updateSolution}
\end{algorithm}
\begin{algorithm}[!t]
  \caption{$\mathrm{gaussN}(\breve\bX, \mathcal{C})$ -- Gauss-Newton 
  minimization algorithm for manifold measurements and state spaces}
  \setalglinespace
  \begin{algorithmic}[1]
  \Require{Initial guess $\breve\bX$; Measurements $\mathcal{C} = 
  \{\langle\bZ_k, \bOmega_k\rangle\}$}
  \Ensure{Optimal solution $\bX^\star$}
  \State $\Fold \leftarrow \inf$, $\Fnew \leftarrow 0 $
  \While{$\Fold - \Fnew > \epsilon$}
    \State $\Fold \leftarrow \Fnew, \;\Fnew \leftarrow 0, \; \bb \leftarrow 0, \; \bH \leftarrow 0$ 
    \ForAll{$k \in \{1\, ...\, K\}$} 
    \State $\langle \chi_k, \bH_k, \bb_k \rangle \leftarrow \mathrm{updateHb}(\bH_k, \bb_k, \bX_k, \bZ_k, \bOmega_k) $
    \State $\Fnew \leftarrow \chi_k$
    \MyEndFor
    \State $\bDeltax \leftarrow \mathrm{solve}(\bH\bDeltax = -\mathbf{b})$
    \State $\breve \bX \leftarrow \mathrm{updateSolution}(\breve \bX, \bDeltax)$
  \MyEndWhile

  \State \Return $\breve\bX$
  \end{algorithmic}
  \label{alg:gn-small}
\end{algorithm}
\begin{algorithm}[!t]
  \caption{$\mathrm{levenbergM}(\breve\bX, \mathcal{C})$ -- 
    Levenberg-Marquardt minimization algorithm for manifold measurements and 
    state spaces}
  \setalglinespace
  \begin{algorithmic}[1]
    \Require{Initial guess $\breve\bX$; Measurements $\mathcal{C} = 
    \{\langle\bZ_k, \bOmega_k\rangle\}$; Maximum number of internal iteration 
    $t_{max}$}
    \Ensure{Optimal solution $\bX^\star$}
    
    \State $\Fold \leftarrow \inf$, $\Fnew \leftarrow 0$, $\Finternal 
    \leftarrow 0$
    \State $\breve \bX_{backup} \leftarrow \breve \bX$
    \State $\lambda \leftarrow \mathrm{initializeLambda}(\breve \bX, 
    \mathcal{C})$
    
    \While{$\Fold - \Fnew < \epsilon$}
      \State $\Fold \leftarrow \Fnew$, $\Fnew \leftarrow 0$, $\bb \leftarrow 
      0$, $\bH \leftarrow 0$ 
      
      \ForAll {$k \in \{1 \dots K\}$}
        \State $\langle\chi, \bH, \bb\rangle \leftarrow \mathrm{updateHb}(\bH, 
        \bb, \bX_k, \bZ_k, \bOmega_k)$
        \State $\Finternal \leftarrow \chi$
      \MyEndFor
      
      \State $t \leftarrow 0$
      \While {$t < t_{max} \wedge t > 0$}
        \State $\bDeltax \leftarrow \mathrm{solve}((\bH + \lambda 
        \bI)\bDeltax=-\bb)$
        \State $\breve \bX \leftarrow \mathrm{updateSolution}(\breve\bX, 
        \bDeltax)$
        \State $\langle\chi, \bH, \bb\rangle \leftarrow \mathrm{updateHb}(\bH, 
\bb, \bX_k, \bZ_k, \bOmega_k)$
        \State $\Fnew \leftarrow \chi$
        
        \If {$\Fnew - \Finternal < 0$}
          \State $\lambda \leftarrow \lambda / 2$
          \State $\breve \bX_{backup} \leftarrow \breve \bX$
          \State $t \leftarrow t-1$
        \Else
          \State $\lambda \leftarrow \lambda \cdot 2$
          \State $\breve \bX \leftarrow \breve \bX_{backup}$
          \State $t \leftarrow t+1$
        \MyEndIf
      \MyEndWhile
    \MyEndWhile
    
    \State \Return $\breve\bX$
  \end{algorithmic}
  \label{alg:lm-small}
\end{algorithm}

\subsection{Design Requirements}\label{sec:solver-requirements}
While designing our system, we devised a set of requirements stemming
from our experience both as developers and as users. Subsequently, we
turned these requirements in some design choices that lead to our
proposed optimization framework. Although most of these requirements indicate
good practices to be followed in potentially any new development, we
highlight here their role in the context of a solver design.
\myParagraph{Easy to use and Symmetric API} 
As users we want to configure, instantiate and run a solver in the same manner, 
regardless to the specific problem to which is applied.
Ideally, we do not want the user to care if the problem is dense or sparse.
Furthermore, in several practical scenarios, one wants to change aspects of the 
solver while it runs - \eg~the minimization algorithm chosen, the 
robust kernel or the termination criterion.
Finally we want to save/retrieve the configuration of a solver and all of
its sub-modules to/from disk.
Thence, the expected usage pattern should be: i) load the specific solver 
configuration from disk ed eventually tune it, ii) assign a problem to the 
solver or load it from file, iii) compute a solution and eventually iv) provide 
statistics about the evolution of optimization.
Note that, many current state-of-the-art \gls{ils} solver allow to easily 
perform the last 3 steps of this process, however, they do not provide the 
ability of permanently write/read their configuration on/from disk - as our 
system does.

\myParagraph{Isolating parameters, working variables and problem
  description/solution} 
When a user is presented to a new potentially
large code-base, having a clear distinction between what the
variables represent and how they are used, substantially reduces the
learning curve. In particular we distinguish between parameters,
working variables and the input/output.  Parameters are those
objects controlling the behavior of the algorithm, such as number
of iterations or the thresholds in a robustifier.
Parameters might include also processing sub-modules, such as the
algorithm to use or the algebraic solver of the linear system.
Summarizing, parameters characterize the behavior of the optimizer, 
independently from the input, and represent the configuration that can be 
stored/retrieved from disk.

In contrast to parameters, working variables are altered during the
computation, and are not directly accessible to the end user.
Finally, we have the description of the problem - \ie~the factor
graph, where the factors and the variables expose an interface
agnostic to the approach that will be used to solve the problem.
  
\myParagraph{Trade-off Development Effort / Performance} 
Quickly developing a proof of concept is a valuable feature to have while 
designing a novel system.
At the same time, once a way to approach the problem has been found, it
becomes perfectly reasonable to invest more effort to enhance its
efficiency. 

Upon instantiation, the system should provide an off-the shelf generic
and fair configuration. Obviously, this might be tweaked later for
enhancing the performances on the specific class of problems. A
possible way to enhance performances is by exploiting the special
structure of a specific class of problems, overriding the general APIs
to perform ad-hoc computations.  This results in layered APIs, where
the functionalities of a level rely only on those of the level below.
As an example, in our architecture the user can either specify the
error function and let the system to compute the Jacobians using
\gls{ad} or provide the analytical expression of the Jacobians if more
performances are needed.  Finally, the user might intervene at a lower
level, providing directly the contribution to the linear system $\bH_k
= \bJ_k^T \bOmega_k \bJ_k$ given by the factor.  In a certain class of
problems also computing this product represents a performance penalty.
An additional benefit provided by this design is the direct support
for \emph{Newton's method}, which can be straightforwardly achieved by
substituting the approximated Hessian $\bH_k = \bJ_k^T \bOmega_k
\bJ_k$ with the analytic one $\bH_k = \frac{ \partial^2 \be_k}{\partial \bDeltax_k^{2}}$.  This feature captures second order approaches such as
\gls{ndt}~\cite{biber2003normal} in the language of our API.
  
\myParagraph{Minimize Codebase} 
The likelihood of bugs in the implementation grows with the size of the 
code-base. For small teams
characterized by a high turnover - like the ones found in academic
environments - maintaining the code becomes an issue.  In this
context we choose to favor the code reuse in spite of a small
performance gain. The same class used to implement an algorithm, a
variable type or a robustifier should be used in all circumstances
- namely sparse and dense problems - where it is needed.

\section{Implementation}\label{sec:solver-architecture}
As support material for this tutorial, we offer an own implementation
of a modular \gls{ils} optimization framework, that has been designed
around the methodology illustrated in Section~\ref{sec:factor-graphs}.
Our system is written in modern C++17 and provides
static type checking, \gls{ad} and a straightforward
interface.  The core of our framework fits in less than 6000 lines of code, 
while the companion libraries to support the most common problems - \eg~ 2D
and 3D \gls{slam}, \gls{icp}, Projective and Dense Registration, Sensor 
Calibration - are contained in 6500 lines of code.
Albeit originally designed as a tool for rapid prototyping, our system
achieves  a high  degree of  customization and competes with other
state-of-the-art systems in terms of performances.

Based on the requirements outlined in \secref{sec:solver-requirements}, we
designed a component model, where the processing objects
(named {\footnotesize\verb|Configurable|}) can possess parameters, support 
dynamic
loading and can be transparently serialized.  Our framework relies on
a custom-built serialization infrastructure that supports format
independent serialization of arbitrary data structures,
named \gls{boss}.  Furthermore, thanks to this foundation, we can
provide both a graphical configurator - that allows to assemble and easily tune 
the modules of a solver - and a command-line utility to edit and run
configurations on the go.

\begin{figure*}[!t]
  \centering
  \includegraphics[width=\linewidth]{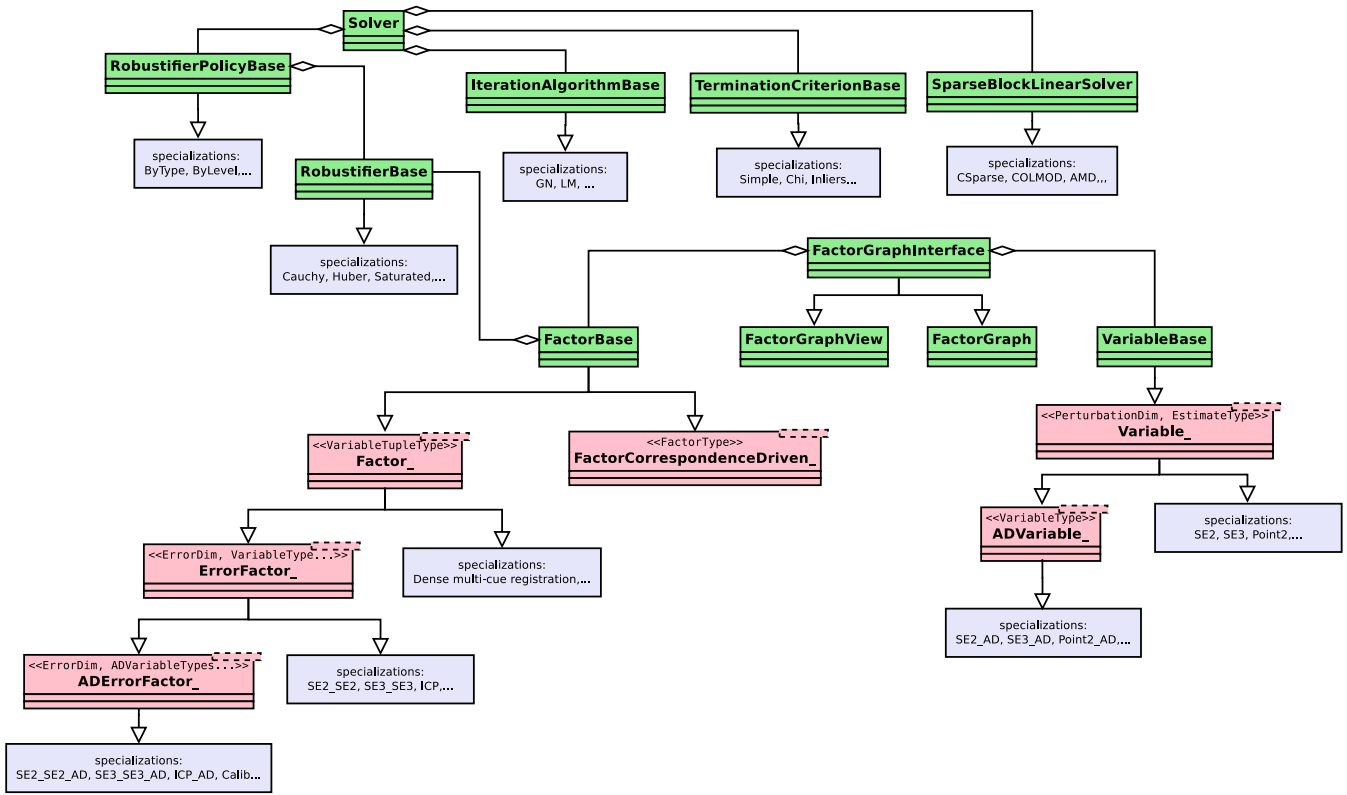}
  \caption{ 
    UML class diagram of our architecture. In green we show the 
    type-independent classes, in pink the type-dependent classes and in pale 
    blue we outline potential specializations. The template class names end 
    with an underscore and the arguments are highlighted above the name, 
    between angular brackets. Arrow lines denote inheritance, and diamonds 
    aggregation/ownership.
  }
  \label{fig:classes}
\end{figure*}

The goal of this section is to provide the reader with a quick overview of the 
proposed system, focusing on how the user interacts with it. Given the 
class-diagram illustrated in~\figref{fig:classes}, in the remainder we will 
first analyze the core modules of the solver and then provide two practical 
examples on how to use it.

\subsection{Solver Core Classes}\label{sec:solver-classes}
Our framework has been designed to satisfy the requirements stated 
in~\secref{sec:solver-design}, embedding unified APIs to cover both dense and 
sparse problems symmetrically. Furthermore, thanks to the \gls{boss} 
serialization library, the user can generate permanent configuration of the 
solver, to be later read and reused on the go. 
The configuration of a solver generally embeds the following parameters:
\begin{itemize}
  \item[--] \textit{Optimization Algorithm}: the algorithm that performs the 
  minimization; currently, only \gls{gn} and \gls{lm} are supported, still, we 
  plan to add also \gls{trm} approaches
  \item[--] \textit{Linear Solver}: the algebraic solver that computes the 
  solution of the linear system $\bH \bDeltax = -\bb$; we embed a naive 
  AMD-based~\cite{amestoy1996approximate} linear solver together with other 
  approaches based on well-known highly-optimized linear algebra libraries - 
  \eg~SuiteSparse~\footnote{{\footnotesize\url{http://faculty.cse.tamu.edu/davis/suitesparse.html}}}
  \item[--] \textit{Robustifier}: the robust kernel function applied to a 
  specific factor; we provide several commonly used instances of robustifier, 
  together with a modular mechanism to assign specific robustifier to different 
  types of factor - called robustifier \textit{policy}
  \item[--] \textit{Termination Criterion}: a simple modules that, based on 
  optimization statistics, checks whether convergence has been reached.
\end{itemize}
Note that, in our architecture there is a clear separation between solver and 
problem. In the next section we will describe how we formalized the latter.
In the remaining of this section, instead, we will focus on the solver classes, 
which are in charge of \textit{computing} the problem solution.

The class {\footnotesize\verb|Solver|} implements a unified interface for our
optimization framework. It presents itself to the user with an unified
data-structure to configure, control, run and monitor the
optimization.  This class allows to select the type of algorithm used
within one iteration, which algorithm to use to solve the linear
system, or which termination criterion to use.  This mechanism is
achieved by delegating the execution of these functions to specific
interfaces.  More in detail, the linear system is stored in a
sparse-block-matrix structure, that effectively separates the solution
of the linear system from the rest of the optimization machinery.
Furthermore, our solver supports incremental updates, and can provide an estimate of partial covariance blocks.
Finally, our system supports hierarchical approaches. In this
sense the problem can be represented at different resolutions
(levels), by using different factors. When the solution at a coarse
level is computed, the optimization starts from the next denser level,
enabling a new set of factors. In the new step, the initial guess is
computed from the solution of the coarser level.

The {\footnotesize\verb|IterationAlgorithmBase|} class defines an interface for 
the outer 
optimization algorithm - \ie~\gls{gn} or \gls{lm}. To carry on its operations, 
it relies on the interface exposed by the {\footnotesize\verb|Solver|} class. 
The latter is 
in charge to invoke the {\footnotesize\verb|IterationAlgorithmBase|}, which 
will run a 
\emph{single} iteration of its algorithm.

Class {\footnotesize\verb|RobustifierBase|} defines an interface for using 
arbitrary
$\rho(u)$ functions - as illustrated in~\secref{sec:robustifiers}. Robust 
kernels can be directly assigned to factors or, alternatively, the user might 
define a \emph{policy}, that based on the status of the actual factor decides 
which robustifier to use. The definition of a policy is done by implementing
the {\footnotesize\verb|RobustifierPolicyBase|} interface. 

Finally, {\footnotesize\verb|TerminationCriterionBase|} defines an interface 
for a predicate
that, exploiting the optimization statistics, detects whether the system has 
converged to a solution or a fatal error has occurred.

\subsection{Factor Graph Classes}\label{sec:factor-graph-classes}
In this section we provide an overview of the top-level classes
constituting a factor graph - \ie~the optimization \emph{problem} - in
our framework.  In specifying new variables or factors, the user can
interact with the system through a layered interface.  More
specifically, factors can be defined using \gls{ad} and, thus,
contained in few lines of code for rapid prototyping or the user can
directly provide how to compute analytic Jacobians if more speed is
required. Furthermore, to achieve extreme efficiency, the user can
choose to compute its own routines to update the quadratic form
directly, consistently in line with our design requirement of
\emph{more-work/more-performance}.  Note that, we observed in our
experiments that in large sparse problems the time required to
linearize the system is marginal compared to the time required to
solve it. Therefore, in most of these cases \gls{ad} can be used
without significant performance losses.

\subsubsection{Variables}
The {\footnotesize\verb|VariableBase|} implements a base abstract interface the 
variables in
a factor graph, whereas 
{\footnotesize\verb|Variable_<PerturbationDim,EstimateType>|}
specializes the base interface on a specific type.
The definition of a 
new variable extending the {\footnotesize\verb|Variable_|} template requires 
the user to 
specify 
i) the type {\footnotesize\verb|EstimateType|} used to store the value of the 
variable 
$\bX_i$, 
ii) the dimension {\footnotesize\verb|PerturbationDim|} of the perturbation 
$\bDeltax_i$ and 
iii) the $\boxplus$ operator.  This is coherent with the methodology
provided in~\secref{sec:factor-graphs}.  In addition to these fields,
variable has an integer key, to be uniquely identified within a factor
graph. Furthermore, a variable can be in either one of these three
states:
\begin{itemize}
  \item[--] {\bf Active}: the variable will be estimated
  \item[--] {\bf Fixed}: the variable stays constant through the optimization
  \item[--] {\bf Disabled}: the variable is ignored and all factors that depend 
  on it are ignored as well.
\end{itemize}
To provide roll-back operations - such as those required
by \gls{lm} - a variable also stores a stack of values.

To support \gls{ad}, we introduce the {\footnotesize\verb|ADVariable_|} 
template,
that is instantiated on a variable \textit{without} \gls{ad}.
Instantiating a variable with \gls{ad} requires to define the
$\boxplus$ operator by using the \gls{ad} scalar type instead of the
usual {\footnotesize\verb|float|} or {\footnotesize\verb|double|}.  This 
mechanism allows us to mix in a problem factors that require \gls{ad} with 
factors that do not.

\subsubsection{Factors}
The base level of the hierarchy is the {\footnotesize\verb|FactorBase|}.
It defines a common interface for this type of graph objects.  It is
responsible of 
i) computing the error - and, thus, the $\chi^2$ -
ii) updating the quadratic form $\bH$ and the right-hand side vector $\bb$ and
iii) invoking the robustifier function (if required).
When an update is requested, a factor is provided with a
structure on which to write the outcome of the operation.  A
factor can be \emph{enabled} or \emph{disabled}. In the latter case, it
will be ignored during the computation. Besides, upon update a
factor might become \emph{invalid}, if the result of the computation
is meaningless.  This occurs for instance in \gls{ba}, when a a point
is projected outside the image plane.

The {\footnotesize\verb|Factor_<VariableTupleType>|} class implements a typed
interface for the factor class. The user willing to extend the class
at this level is responsible of implementing the entire 
{\footnotesize\verb|FactorBase|}
interface, relying on functions for typed access to the
blocks of the system matrix $\bH$ and of the coefficient vector $\bb$. In this 
case, the block size is determined from the dimension of the perturbation
vector of the variables in the template argument list.  We extended
the factors at this level to implement approaches such as dense
multi-cue registration~\cite{dellacorte2018mpr}.
Special structures in the Jacobians can be exploited to speed up the
calculation of $\bH_k$ whose computation has a non negligible
cost.

The {\footnotesize\verb|ErrorFactor_<ErrorDim, VariableTypes...>|} class 
specializes a
typed interface for the factor class, where the user has to implement
both the error function $\be_k$ and the Jacobian blocks
$\bJ_{k,i}$. The calculation of the $\bH$ and the $\bb$ blocks is done
through loops unrolled at compile time since the types and the
dimensions of the variables/errors are part of the type.

The {\footnotesize\verb|ADErrorFactor_<Dim, VariableTypes...>|} class further 
specializes the 
{\footnotesize\verb|ErrorFactor_|}. Extending the class at this level only 
required to specify 
\emph{only} the error function.
The Jacobians are computed through \gls{ad}, and the updates of $\bH$ and the 
$\bb$ are done according to the base class.

Finally, the {\footnotesize\verb|FactorCorrespondenceDriven_<FactorType>|} 
implements a 
mechanism
that allows the solver to iterate over multiple factors of the same
type and connecting the same set of variables, without the need of
explicitly storing them in the graph. A 
{\footnotesize\verb|FactorCorrespondenceDriven_|} is
instantiated on a base type of factor, and it is specialized by
defining which actions should be carried on as a consequence of the
selection of the ``next'' factor in the pool by the solver.
The solver sees this type of factor as multiple ones, albeit a 
{\footnotesize\verb|FactorCorrespondenceDriven_|} is stored just
once in memory. Each time a {\footnotesize\verb|FactorCorrespondenceDriven_|} 
is accessed by the solver a
callback changing the internal parameters is called. In its basic implementation this class takes a container of
corresponding indices, and two data containers: {\footnotesize\verb|Fixed|} and 
{\footnotesize\verb|Moving|}.
Each time a new factor within
the {\footnotesize\verb|FactorCorrespondenceDriven_|} is requested, the factor 
is
configured by: selecting the next pair of corresponding indices from
the set, and by picking the elements in {\footnotesize\verb|Fixed|} and 
{\footnotesize\verb|Moving|} at those
indices.  As an instance, to use our solver within an \gls{icp} algorithm,
the user has to configure the factor by setting the {\footnotesize\verb|Fixed|} 
and 
{\footnotesize\verb|Moving|} point clouds.  The correspondence vector can be 
changed anytime to reflect a new data association. This results in different
correspondences to be considered at each iteration.

\subsubsection{FactorGraph}
To carry on an iteration, the solver has to
\emph{iterate} over the factors and, hence, it requires to randomly access the 
variables.
Restricting the solver to access a graph through an interface of random access iterators enables us to decouple the way the graph is accessed from the way it is stored.  This would allow us to
support transparent off-core storage that can be useful on very large
problems.

A {\footnotesize\verb|FactorGraphInterface|} defines the way to access a 
graph.  In
our case we use integer values as key for variables and factors.  The
solver accesses a graph only through the 
{\footnotesize\verb|FactorGraphInterface|}
and, thence, it can read/write the value of state variables, read the
factors, but it is not allowed to modify the graph structure.

A heap-based concrete implementation of a factor graph is provided by
the {\footnotesize\verb|FactorGraph|} class, that specializes the interface. 
The {\footnotesize\verb|FactorGraph|} supports transparent 
serialization/deserialization.
Our framework makes use of the open-source math library Eigen~\cite{eigenweb}, 
which provides fast and easy matrix operation. The 
serialization/deserialization of variable and factors that are constructed on 
Eigen types is automatically handled by our \gls{boss} library.

In sparse optimization it is common to operate on a local portion of
the entire problem. Instrumenting the solver with methods to specify
the local portions would bloat the implementation.  Alternatively we
rely on the concept of  {\footnotesize\verb|FactorGraphView|}  that exposes an 
interface on a 
local portion of a {\footnotesize\verb|FactorGraph|} - or of any other object 
implementing the
{\footnotesize\verb|FactorGraphInterface|}.

\section{Experiments} \label{sec:experiments}
\begin{figure*}[!t]
  \centering
  \begin{subfigure}{0.48\linewidth}
    \centering
    \includegraphics[width=0.8\columnwidth]{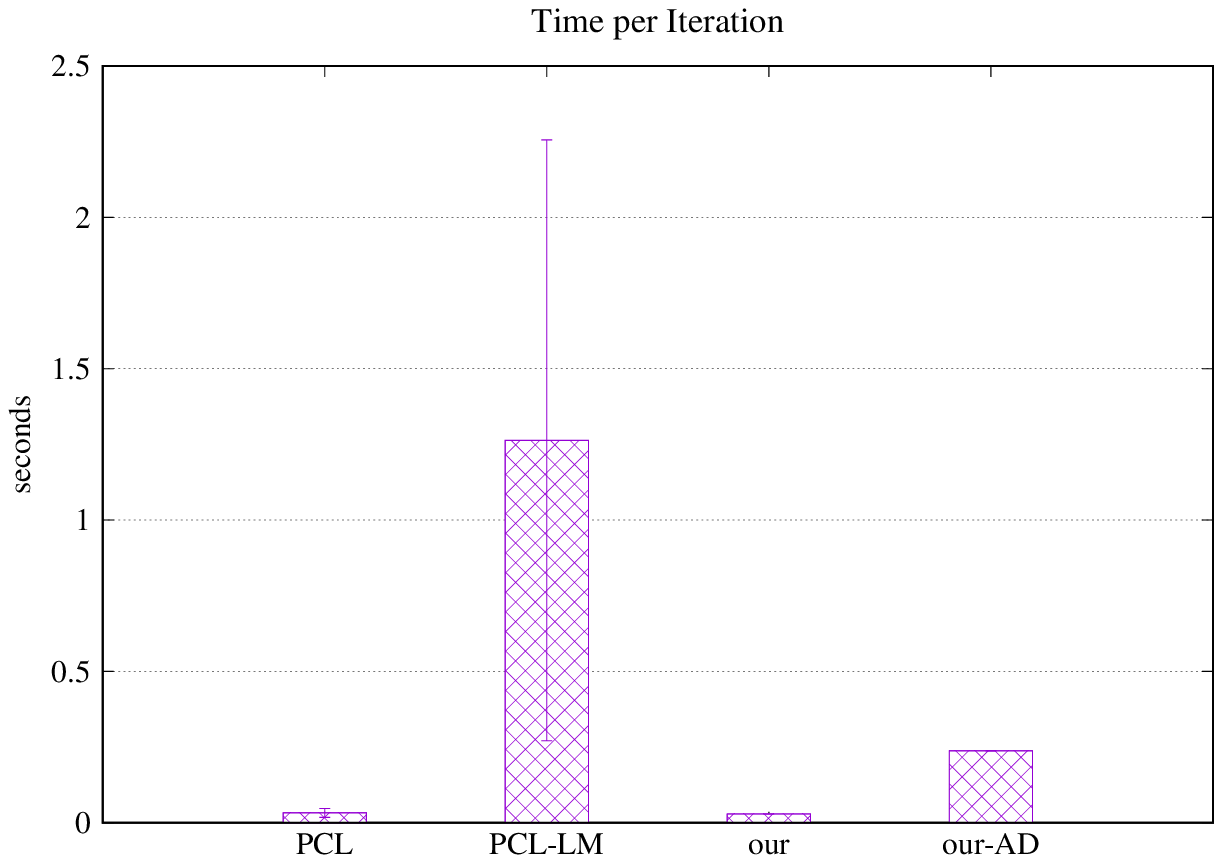}
    \subcaption{Iteration time: \texttt{ICL-NUIM} - \texttt{lr-0}}
    \label{fig:dense-timings-it-icl}
  \end{subfigure}
  \begin{subfigure}{0.48\linewidth}
    \centering
    \includegraphics[width=0.8\columnwidth]{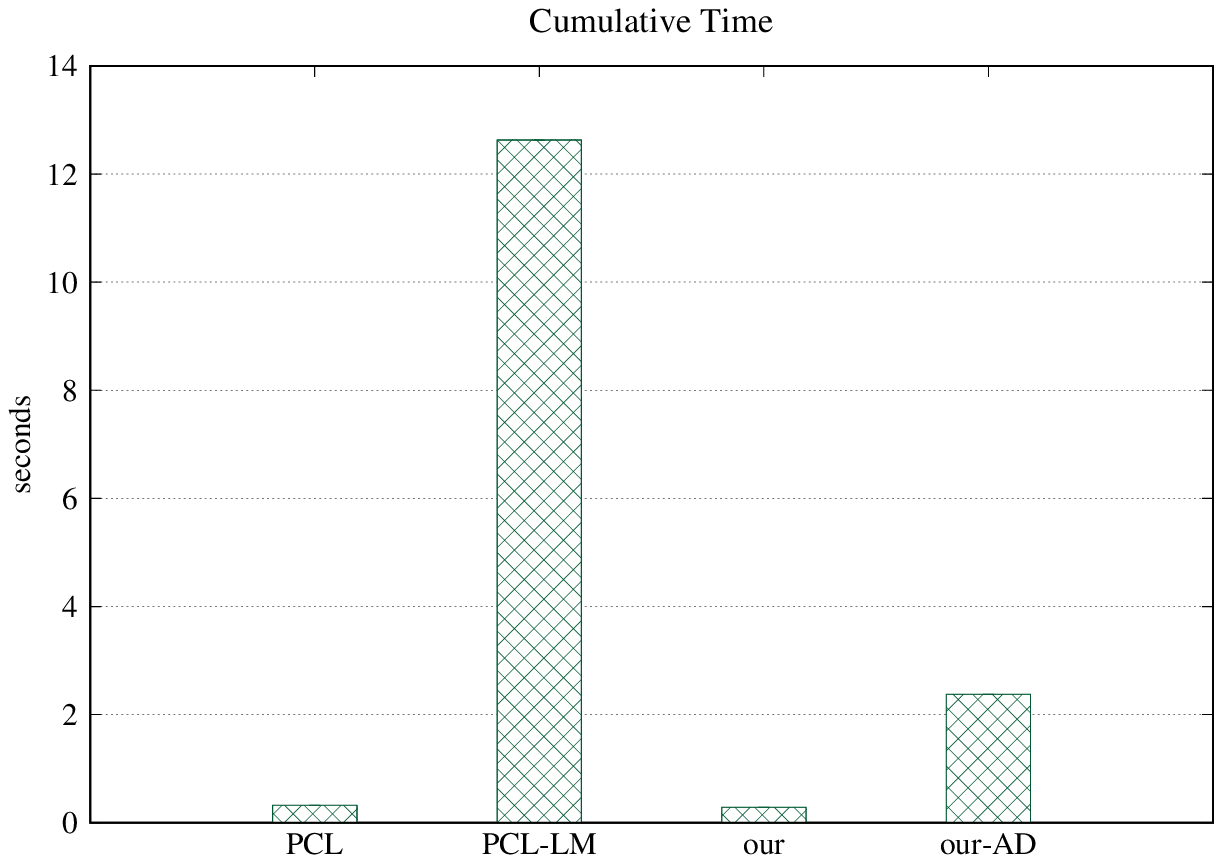}
    \subcaption{Cumulative time: \texttt{ICL-NUIM} - \texttt{lr-0}}
    \label{fig:dense-timings-cum-icl}
  \end{subfigure} \\ \vspace{5pt}
  \begin{subfigure}{0.48\linewidth}
    \centering
    \includegraphics[width=0.8\columnwidth]{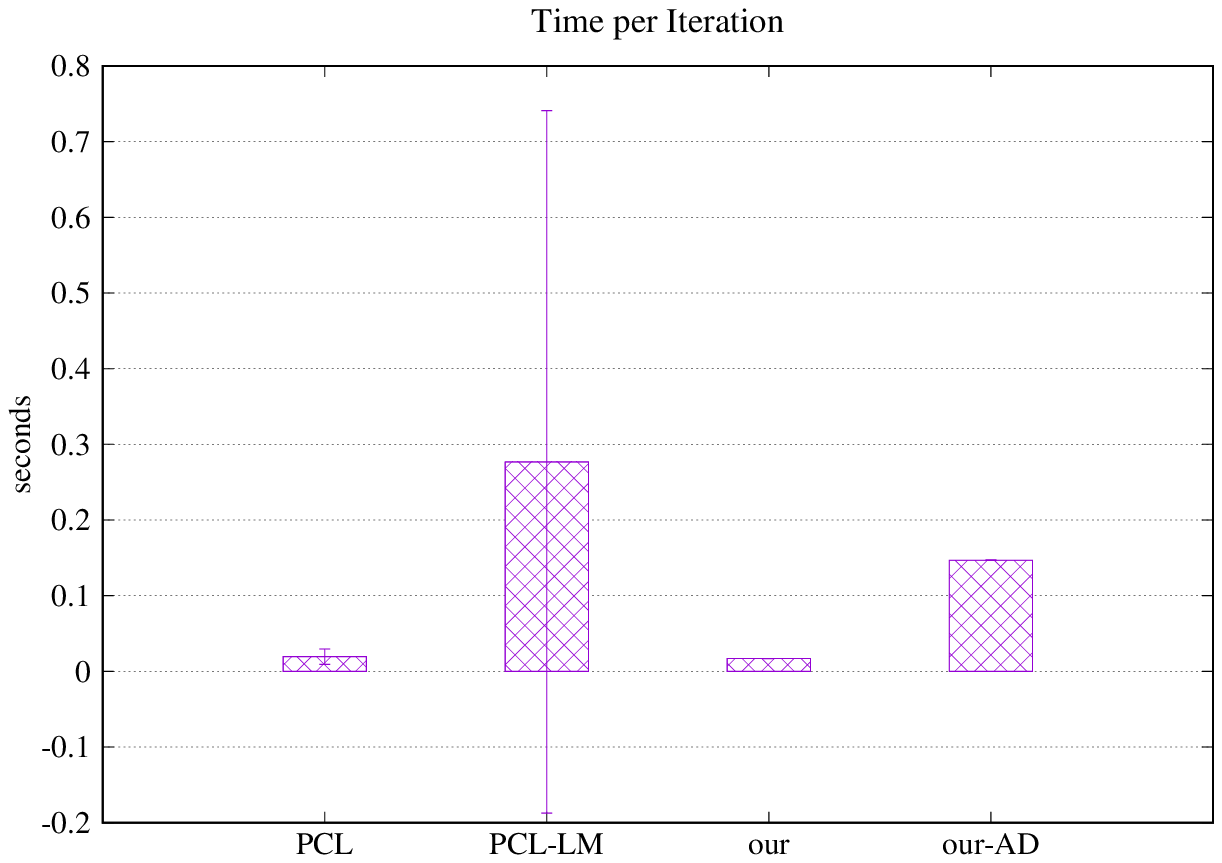}
    \subcaption{Iteration time: \texttt{ETH-Hauptgebaude}}
    \label{fig:dense-timings-it-eth-haupt}
  \end{subfigure}
  \begin{subfigure}{0.48\linewidth}
    \centering
    \includegraphics[width=0.8\columnwidth]{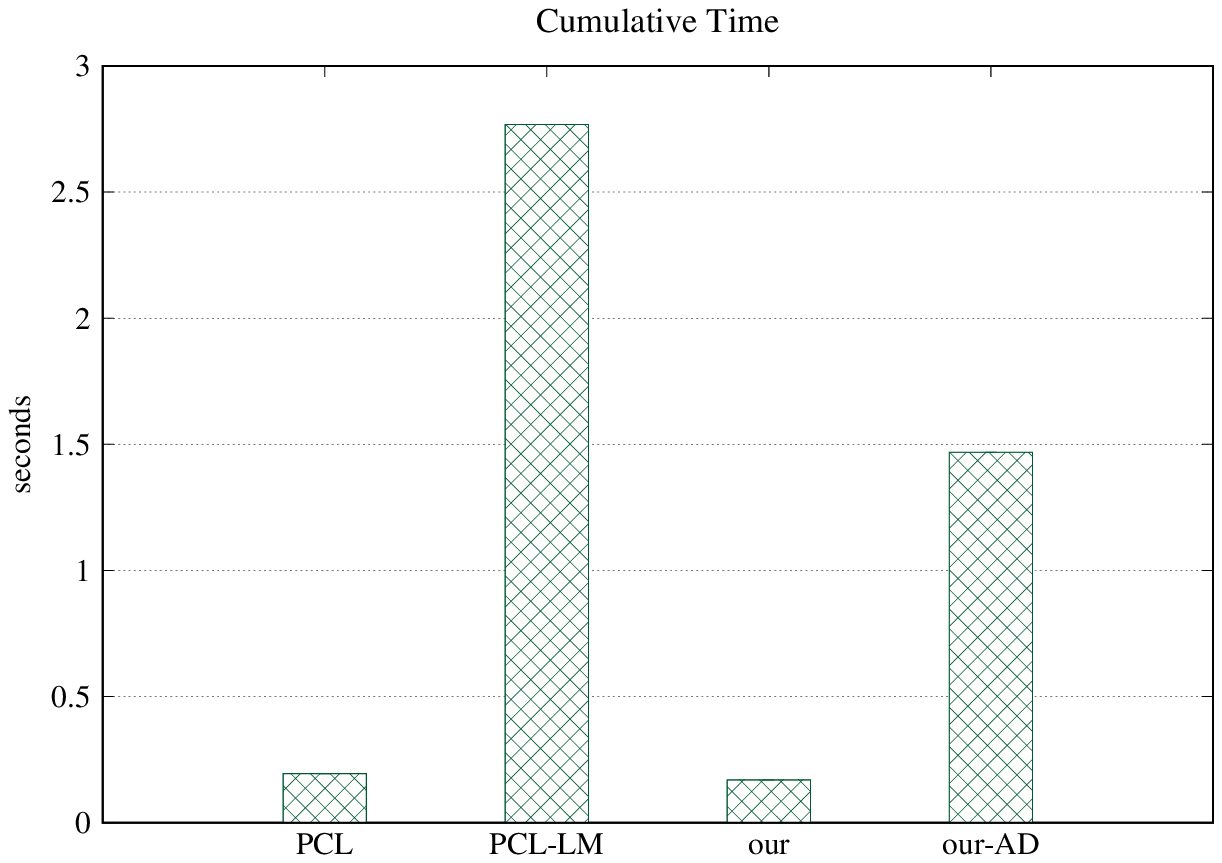}
    \subcaption{Cumulative time: \texttt{ETH-Hauptgebaude}}
    \label{fig:dense-timings-eth-haupt}
  \end{subfigure} \\ \vspace{5pt}
  \begin{subfigure}{0.48\linewidth}
    \centering
    \includegraphics[width=0.8\columnwidth]{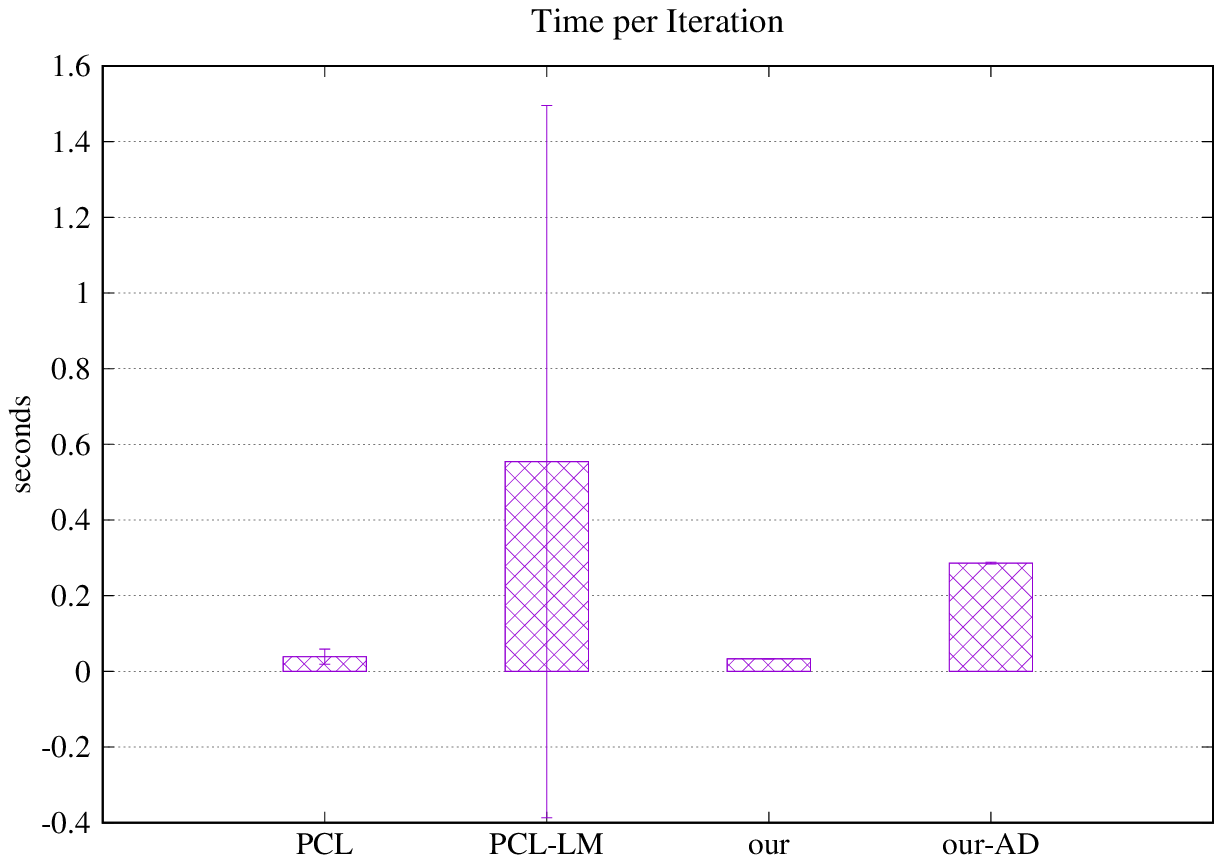}
    \subcaption{Iteration time: \texttt{ETH-Apartment}}
    \label{fig:dense-timings-it-eth-apartment}
  \end{subfigure}
  \begin{subfigure}{0.48\linewidth}
    \centering
    \includegraphics[width=0.8\columnwidth]{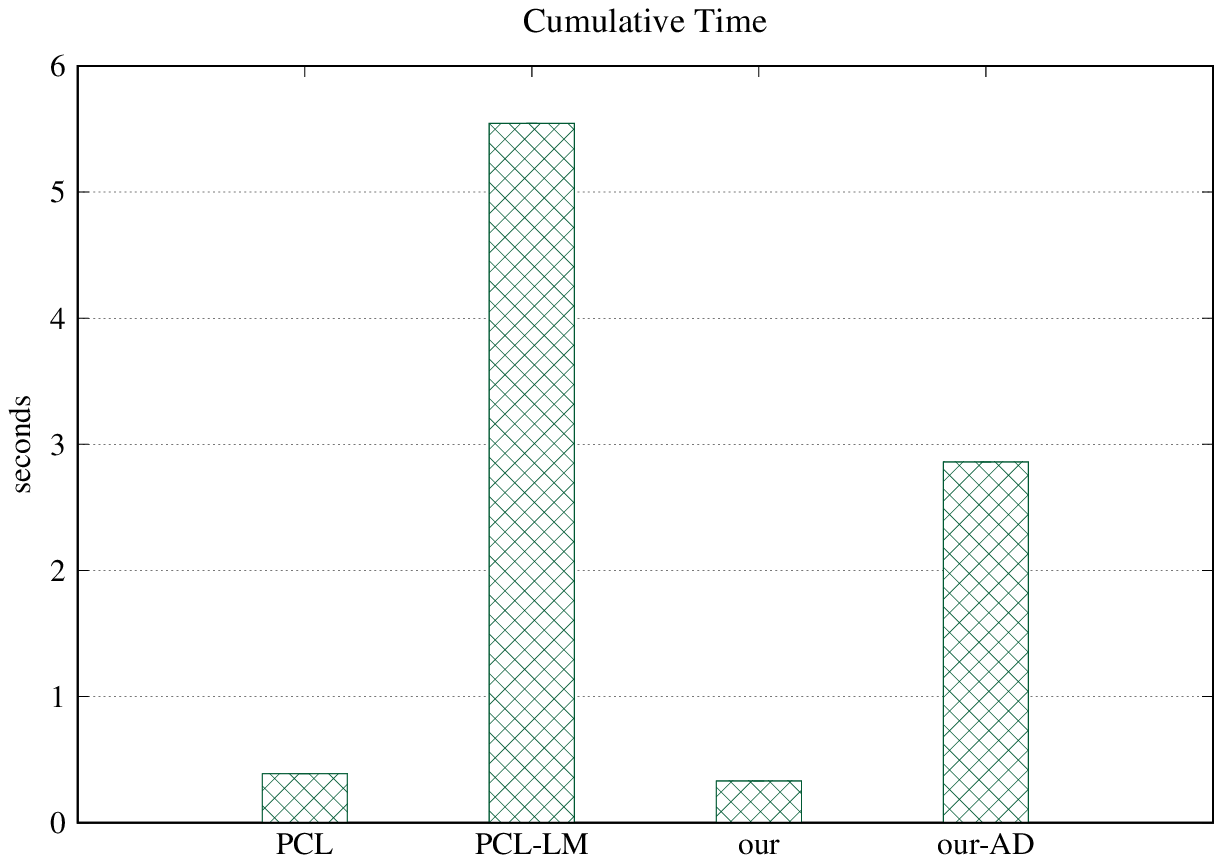}
    \subcaption{Cumulative time: \texttt{ETH-Apartment}}
    \label{fig:dense-timings-eth-apartment}
  \end{subfigure} \\ \vspace{5pt}
  \begin{subfigure}{0.48\linewidth}
    \centering
    \includegraphics[width=0.8\columnwidth]{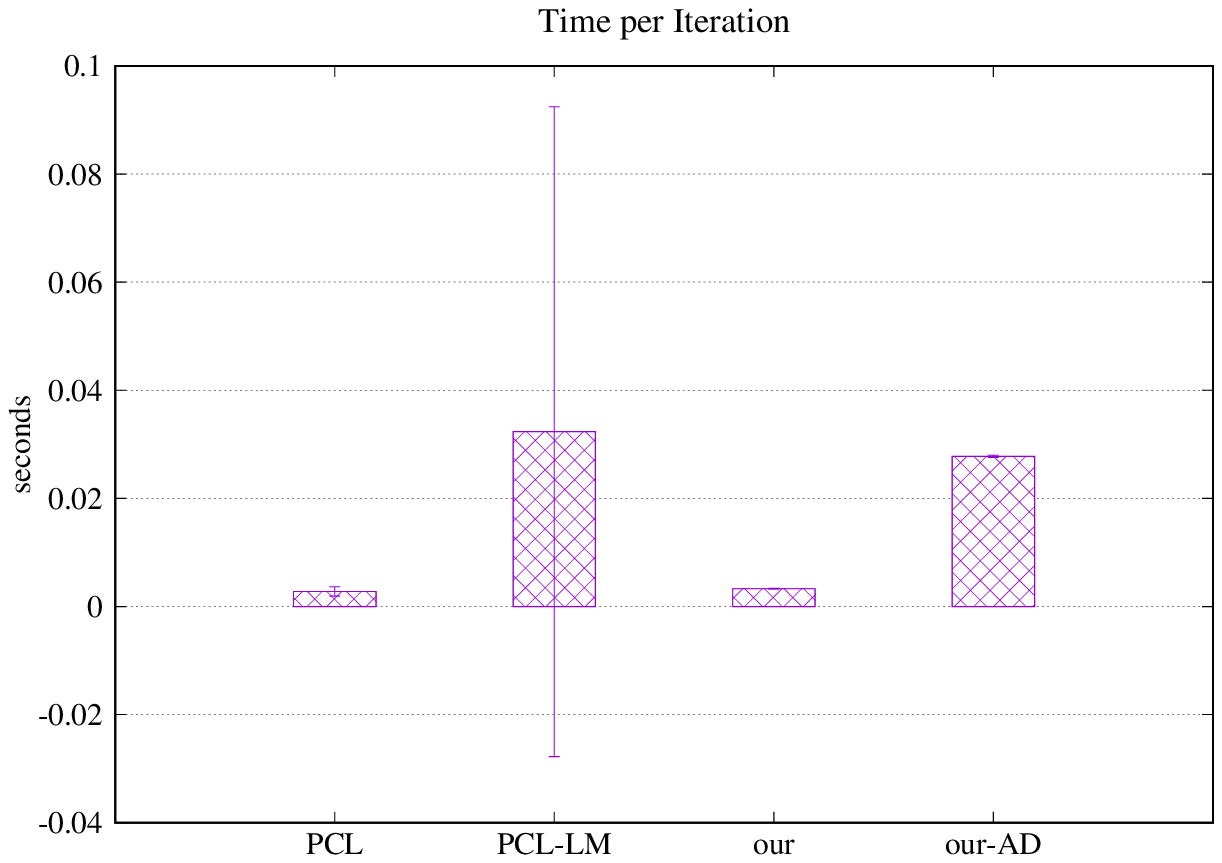}
    \subcaption{Iteration time: \texttt{Stanford-Bunny}}
    \label{fig:dense-timings-it-bunny}
  \end{subfigure}
  \begin{subfigure}{0.48\linewidth}
    \centering
    \includegraphics[width=0.8\columnwidth]{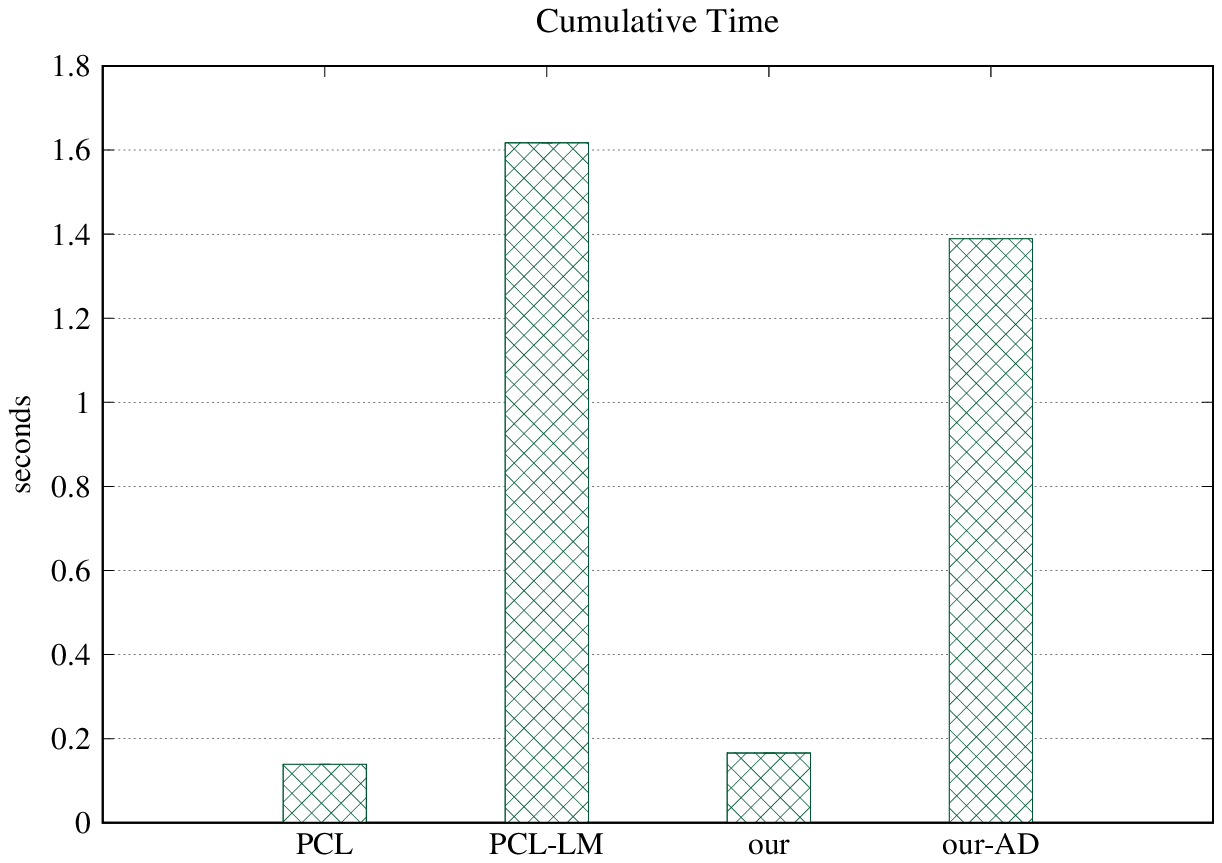}
    \subcaption{Cumulative time: \texttt{Stanford-Bunny}}
    \label{fig:dense-timings-bunny}
  \end{subfigure}
  \caption{Timing analysis of the \gls{ils} optimization. On the left column
    are reported the mean and standard deviation of a full \gls{ils} iteration -
    computed over 10 total iterations. On the right column, instead, the
    cumulative time to perform all 10 \gls{ils} iterations.}
  \label{fig:dense-timings}
\end{figure*}

\begin{figure*}[!t]
  \centering
  \begin{subfigure}{0.489\linewidth}
    \centering
    \includegraphics[width=\columnwidth]{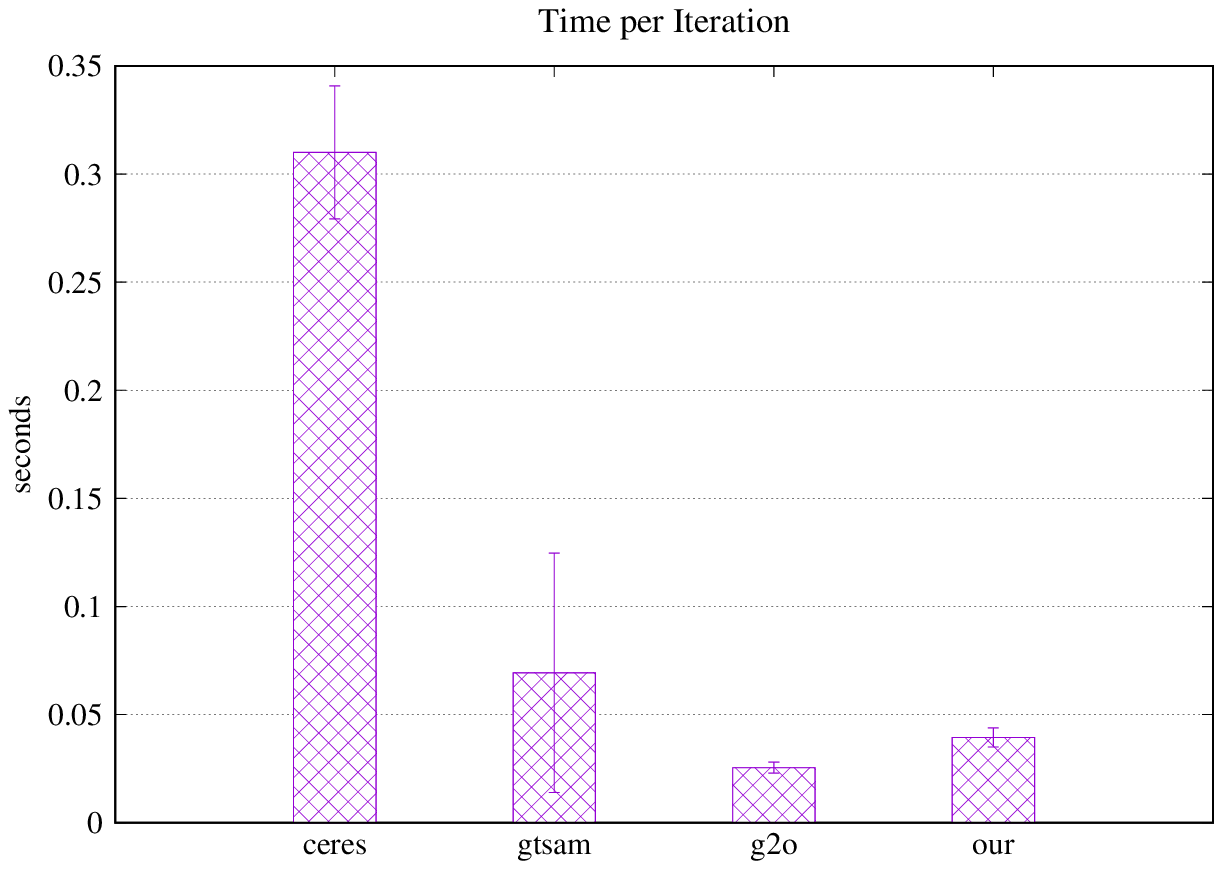}
    \subcaption{Iteration time: \texttt{kitti-00}.}
    \label{fig:it-time-kitti-00}
  \end{subfigure}
  \begin{subfigure}{0.489\linewidth}
    \centering
    \includegraphics[width=\columnwidth]{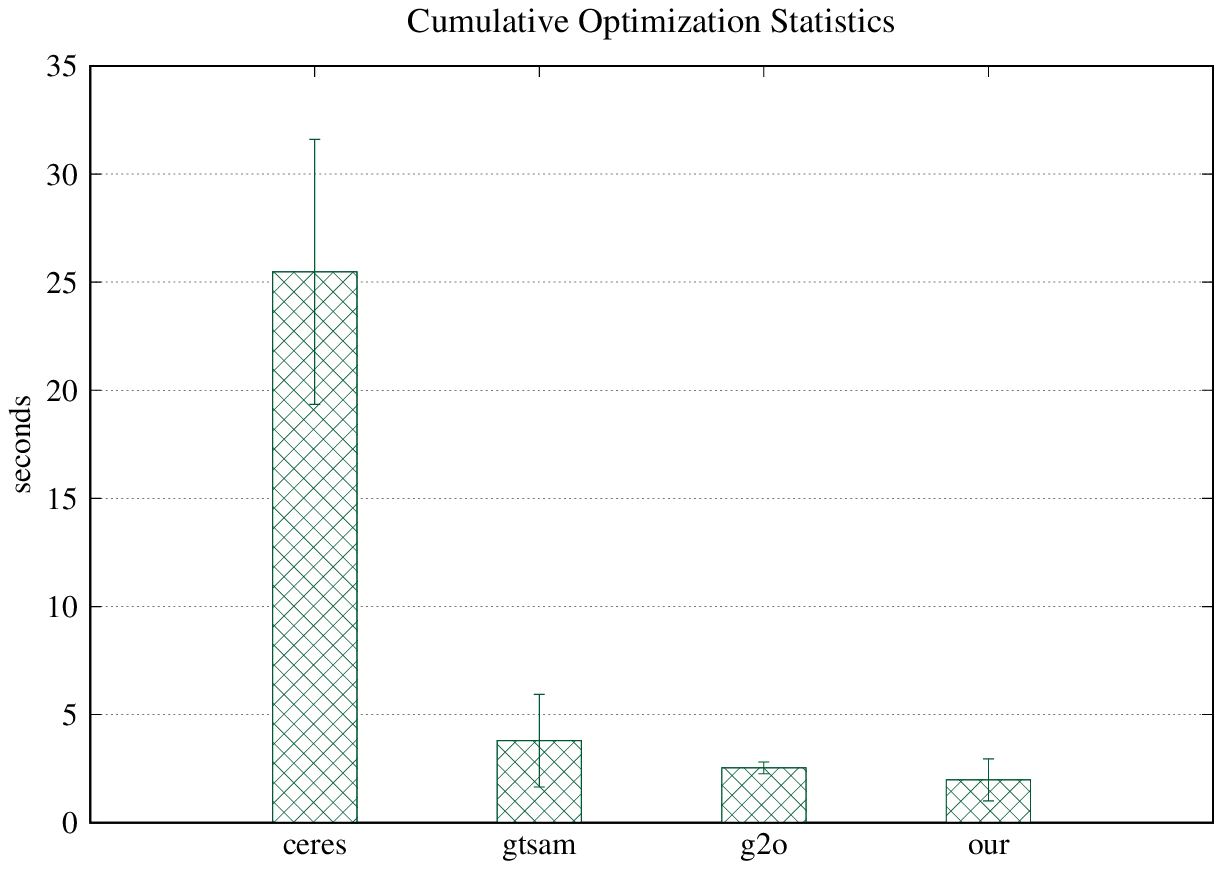}
    \subcaption{Cumulative time: \texttt{kitti-00}.}
    \label{fig:cum-time-kitti-00}
  \end{subfigure} \\ \vspace{10pt}
  \begin{subfigure}{0.489\linewidth}
    \centering
    \includegraphics[width=\columnwidth]{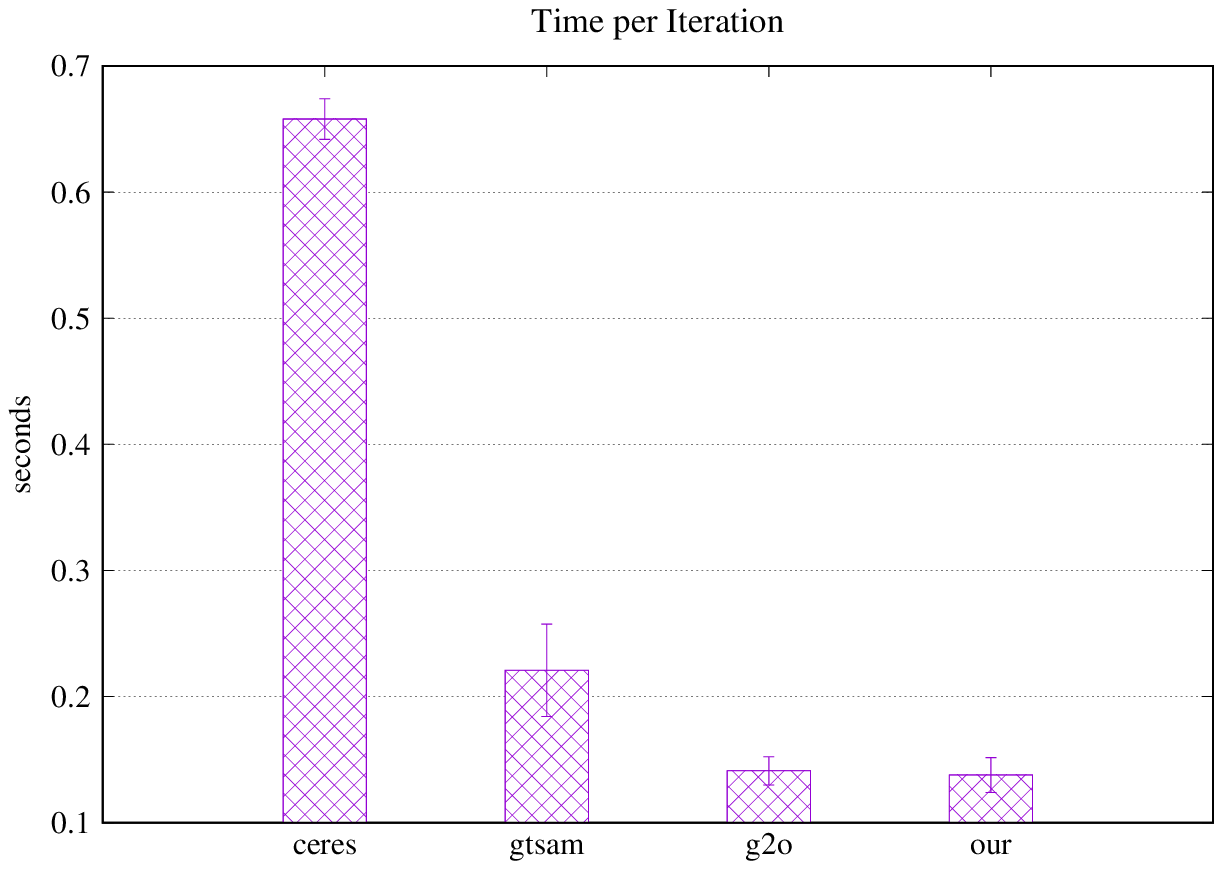}
    \subcaption{Iteration time: \texttt{sphere-b}.}
    \label{fig:it-time-sphere-b}
  \end{subfigure}
  \begin{subfigure}{0.489\linewidth}
    \centering
    \includegraphics[width=\columnwidth]{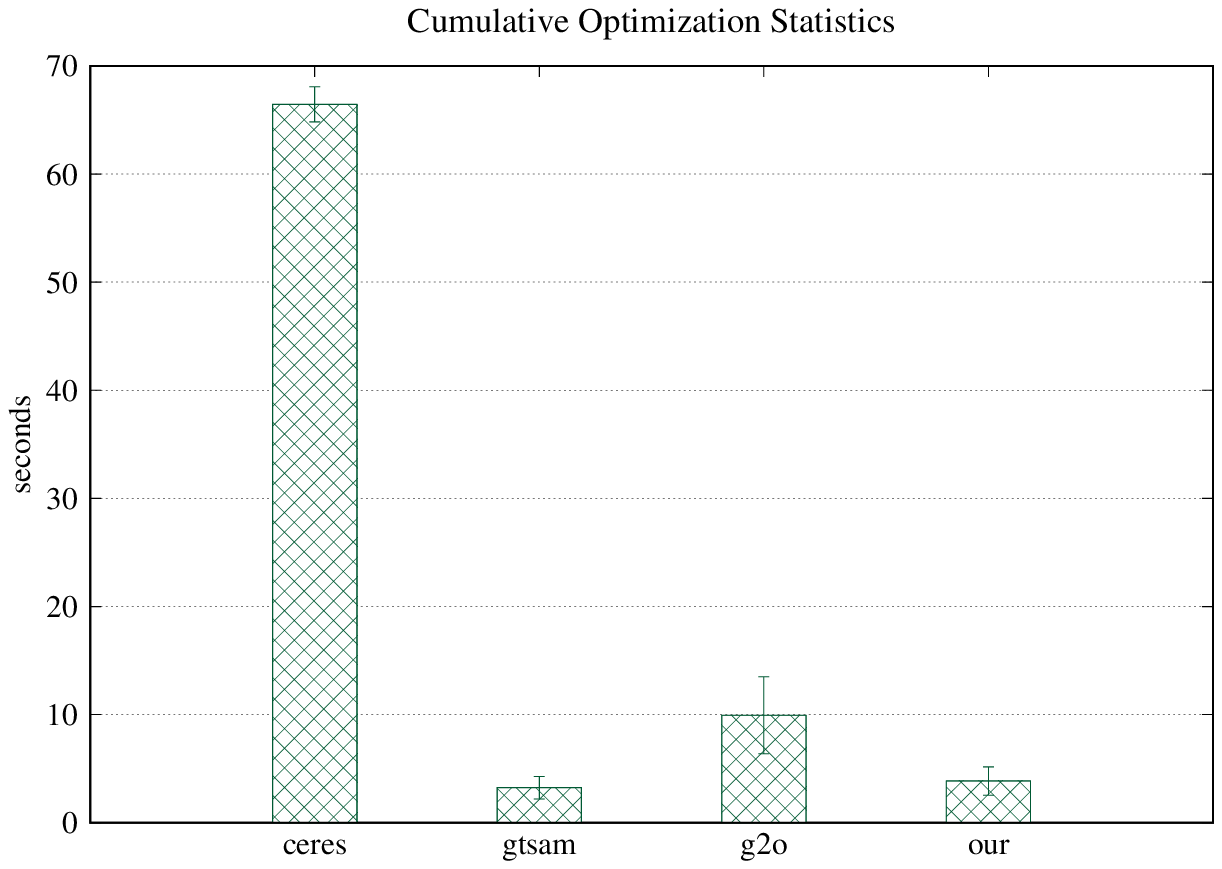}
    \subcaption{Cumulative time: \texttt{sphere-b}.}
    \label{fig:cum-time-sphere-b}
  \end{subfigure} \\ \vspace{10pt}
  \begin{subfigure}{0.489\linewidth}
    \centering
    \includegraphics[width=\columnwidth]{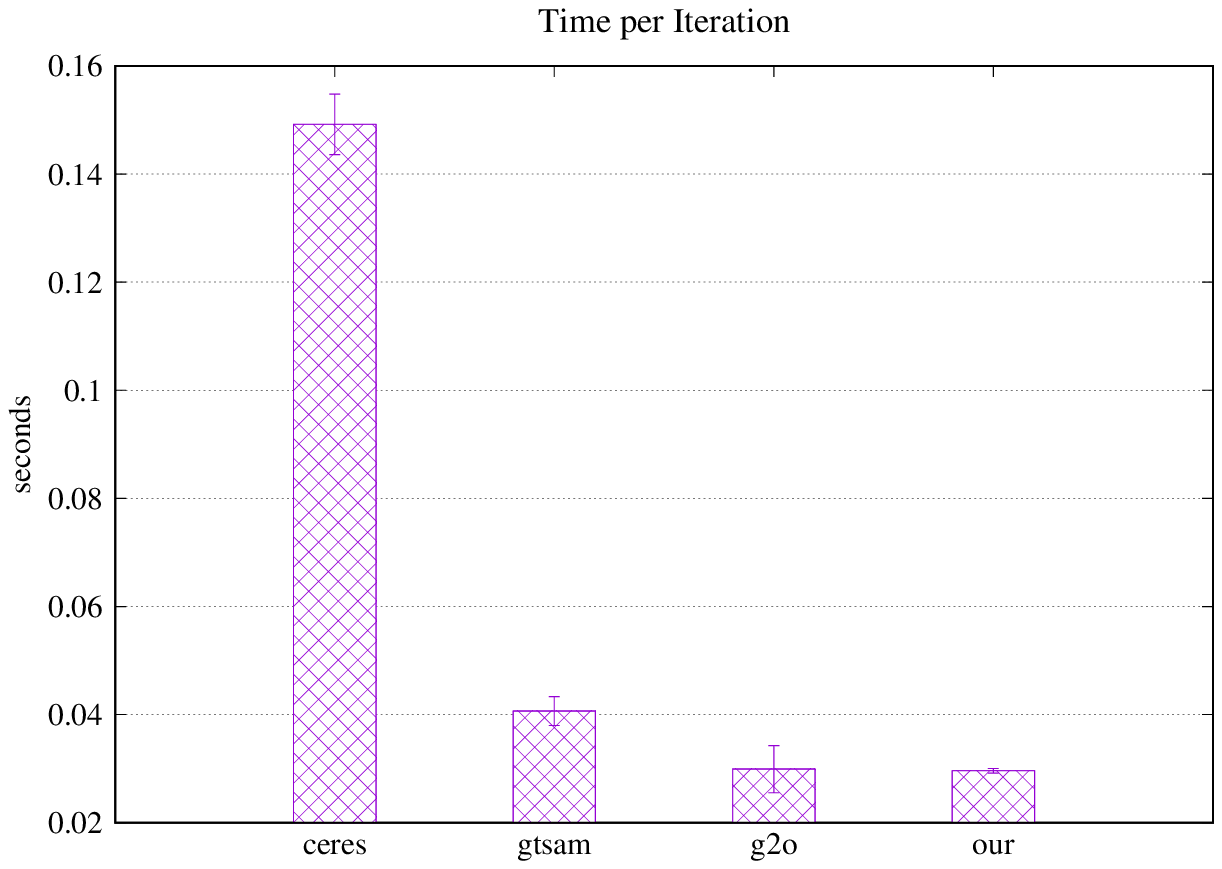}
    \subcaption{Iteration time: \texttt{torus-b}.}
    \label{fig:it-time-torus-b}
  \end{subfigure}
  \begin{subfigure}{0.489\linewidth}
    \centering
    \includegraphics[width=\columnwidth]{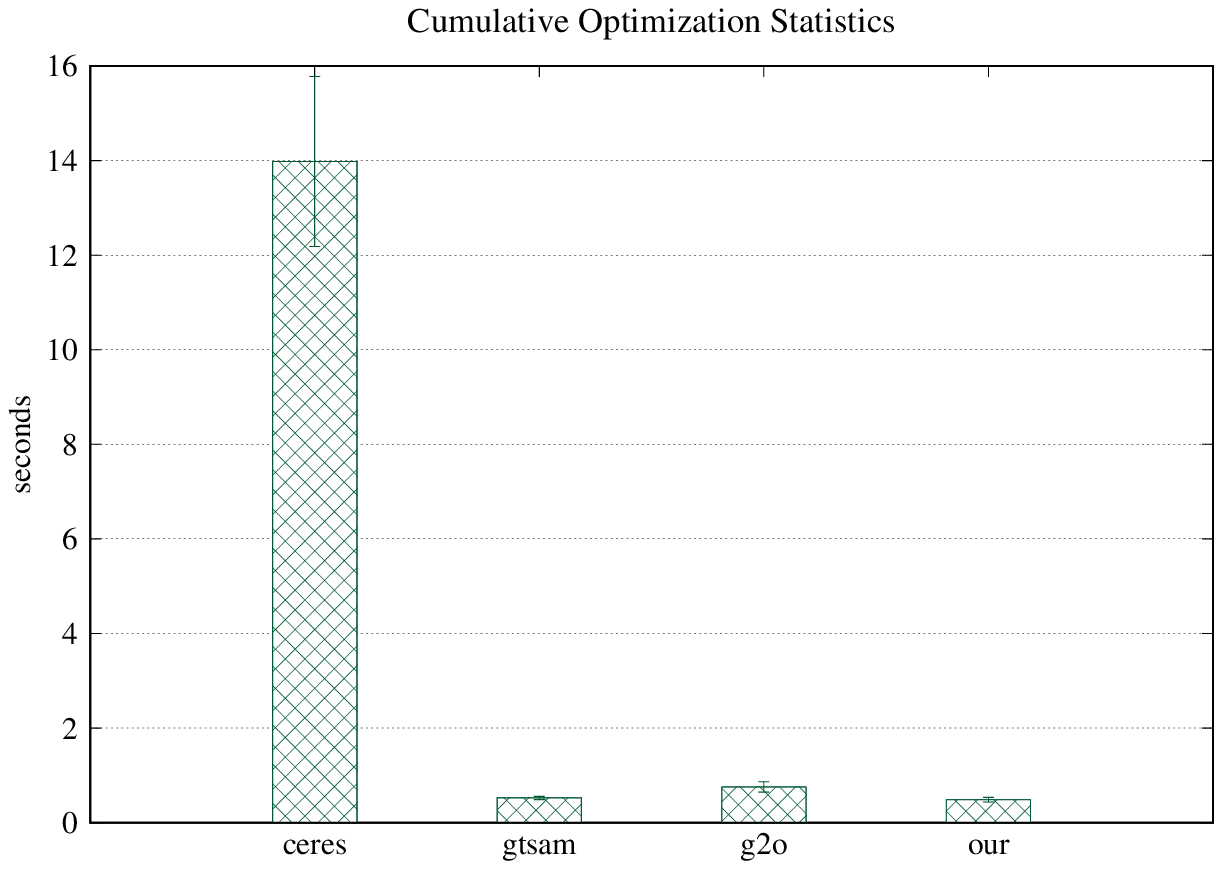}
    \subcaption{Cumulative time: \texttt{torus-b}.}
    \label{fig:cum-time-torus-b}
  \end{subfigure}
  \caption{Timing analysis of different optimization frameworks. The left
    column reports the mean and standard deviation of the time to perform a
    complete \gls{lm} iteration. The right column, instead, illustrates the 
    total
    time to reach convergence -
    mean and standard deviation.}
  \label{fig:pgo-timings}
\end{figure*}

In this section we propose several comparisons between our framework and other
state-of-the-art optimization system. The aim of these experiments is to
support the claims on the performance of our framework and, thus, we focused on
the accuracy of the computed solution and the time required to achieve it.
Experiments have been performed both on dense scenarios - such as \gls{icp} -
and sparse ones - \eg~\gls{pgo} and \gls{plgo}.


\subsection{Dense Problems}\label{sec:experiments-dense}
Many well-known \gls{slam} problems related to the front-end can be solved
exploiting the \gls{ils} formulation introduced before. In such scenarios - \eg
point-clouds registration - the number of variables is small compared to the
observations' one. Furthermore, at each registration step, the data-association
is usually recomputed to take advantage of the new estimate. In this sense, one
has to build the factor graph associated to the problem from scratch at each
step. In such contexts, the most time consuming part of the process is
represented by the construction of linear system
in~\eqref{eq:nonlinear-solution} and not its solution.

To perform dense experiments, we choose a well-known instance of this
kind of problems: \gls{icp}. We conducted multiple tests, comparing
our framework to the current state-of-the-art PCL
library~\cite{rusu2011pcl} on the standard registration datasets
summarized in~\tabref{tab:icp-dataset-specs}.
\begin{table}[!t]
  \centering
  \begin{tabular}{|c|c|c|c|}
    \hline
    \textsc{Dataset} &
    \textsc{Sensor} &
    \textsc{Variables} &
    \textsc{Factors} \\ \hline
    \texttt{ICL-NUIM}~\cite{handa2014benchmark} &
    RGB-D &
    1 &
    307200 \\ \hline
    \texttt{ETH-Hauptgebaude}~\cite{pomerleau2012challenging} &
    Laser-scanner &
    1 &
    189202 \\ \hline
    \texttt{ETH-Apartment}~\cite{pomerleau2012challenging} &
    Laser-scanner &
    1 &
    370276 \\ \hline
    \texttt{Stanford-Bunny}~\cite{curless1996volumetric} &
    3D digitalizer &
    1 &
    35947 \\ \hline
  \end{tabular}
  \caption{Specification of the datasets used to perform dense benchmarks.}
  \label{tab:icp-dataset-specs}
\end{table}
In all the cases, we setup a controlled benchmarking environment, to have a
precise ground-truth. In the \texttt{ETH-Hauptgebaude}, \texttt{ETH-Apartment}
and \texttt{Stanford-Bunny}, the raw data consists in a series of range scans.
Therefore, in such cases, we constructed the \gls{icp} problem as follows:%
\begin{itemize}
  \item[--] reading of the first raw scan and generate a point cloud
  \item[--] transformation of the point cloud according to a known isometry
  $\bT_{GT}$
  \item[--] generation of perfect association between the two clouds
  \item[--] registration starting from $\bT_{init} = \bI$.
\end{itemize}
Since our focus is on the \gls{ils} optimization, we used the same set
of data-associations and the same initial guess for all approaches. This
methodology renders the comparison fair and unbiased.
As for the \texttt{ICL-NUIM}
dataset, since obtained the raw point cloud unprojecting the range
image of the first reading of the \texttt{lr-0} scene.  After this
initial preprocessing, the benchmark flow is the same described
before.

In this context, we compared %
i) the accuracy of the solution obtained computing the translational and
rotational error of the estimate and %
ii) the time required to achieve that solution.
\begin{table*}[!t]
  \centering
  \begin{tabular}{cc|c|c|c|c|}
    \cline{3-6}
    &
    &
    \textsc{PCL} &
    \textsc{PCL-LM} &
    \textsc{Our} &
    \textsc{Our-AD}
    \\ \cline{1-6}
    \multicolumn{1}{ |c  }{\multirow{2}{*}{\texttt{ICL-NUIM-lr-0}} } &
    \multicolumn{1}{ |c| }{$\mathrm{e}_{pos} [m]$} &
    $6.525 \times 10^{-06}$ &
    $1.011 \times 10^{-04}$ &
    $1.390 \times 10^{-06}$ &
    $\mathbf{6.743 \times 10^{-07}}$ \\ \cline{2-6}
    \multicolumn{1}{ |c  }{}                        &
    \multicolumn{1}{ |c| }{$\mathrm{e}_{rot} [rad]$} &
    $1.294 \times 10^{-08}$ &
    $2.102 \times 10^{-05}$ &
    $\mathbf{1.227 \times 10^{-08}}$ &
    $9.510 \times 10^{-08}$ \\ \cline{1-6}
    \multicolumn{1}{ |c  }{\multirow{2}{*}{\texttt{ETH-Haupt}} } &
    \multicolumn{1}{ |c| }{$\mathrm{e}_{pos} [m]$} &
    $4.225 \times 10^{-06}$ &
    $2.662 \times 10^{-05}$ &
    $1.581 \times 10^{-06}$ &
    $\mathbf{2.384 \times 10^{-07}}$ \\ \cline{2-6}
    \multicolumn{1}{ |c  }{}                        &
    \multicolumn{1}{ |c| }{$\mathrm{e}_{rot} [rad]$} &
    $5.488 \times 10^{-08}$ &
    $8.183 \times 10^{-06}$ &
    $\mathbf{1.986 \times 10^{-08}}$ &
    $1.952 \times 10^{-07}$ \\ \cline{1-6}
    \multicolumn{1}{ |c  }{\multirow{2}{*}{\texttt{ETH-Apart}} } &
    \multicolumn{1}{ |c| }{$\mathrm{e}_{pos} [m]$} &
    $1.527 \times 10^{-06}$ &
    $5.252 \times 10^{-05}$ &
    $\mathbf{6.743 \times 10^{-07}}$ &
    $2.023 \times 10^{-06}$ \\ \cline{2-6}
    \multicolumn{1}{ |c  }{}                        &
    \multicolumn{1}{ |c| }{$\mathrm{e}_{rot} [rad]$} &
    $7.134 \times 10^{-08}$ &
    $1.125 \times 10^{-04}$ &
    $\mathbf{1.548 \times 10^{-08}}$ &
    $1.564 \times 10^{-07}$ \\ \cline{1-6}
    \multicolumn{1}{ |c  }{\multirow{2}{*}{\texttt{bunny}} } &
    \multicolumn{1}{ |c| }{$\mathrm{e}_{pos} [m]$} &
    $\mathbf{1.000 \times 10^{-12}}$ &
    $1.352 \times 10^{-05}$ &
    $1.284 \times 10^{-06}$ &
    $9.076 \times 10^{-06}$ \\ \cline{2-6}
    \multicolumn{1}{ |c  }{}                        &
    \multicolumn{1}{ |c| }{$\mathrm{e}_{rot} [rad]$} &
    $\mathbf{1.515 \times 10^{-07}}$ &
    $2.665 \times 10^{-04}$ &
    $1.269 \times 10^{-06}$ &
    $5.660 \times 10^{-07}$ \\ \cline{1-6}
  \end{tabular}
  \caption{Comparison of the final registration error of the optimization
  result.}
  \label{tab:dense-ate}
\end{table*}
We compared the recommended PCL registration suite - that uses the
Horn formulas - against our framework with and without \gls{ad}. Furthermore,
we also provide results obtained using PCL implementation of the \gls{lm}
optimization algorithm.

As reported in~\tabref{tab:dense-ate}, the final registration error is almost
negligible in all cases. Instead, in~\figref{fig:dense-timings} we document the
speed of each solver. When using the full potential of our framework -
\ie~using analytic Jacobians - it is able to achieve results in general equal
or better than the off-the-shelf PCL registration algorithm. Using \gls{ad} has
a great impact on the iteration time, however, our system is able to be faster
than the PCL implementation of \gls{lm} also in this case.

\subsection{Sparse Problems}\label{sec:experiments-sparse}
\begin{figure*}[!t]
  \centering
  \begin{subfigure}{0.489\linewidth}
    \centering
    \includegraphics[width=\columnwidth]{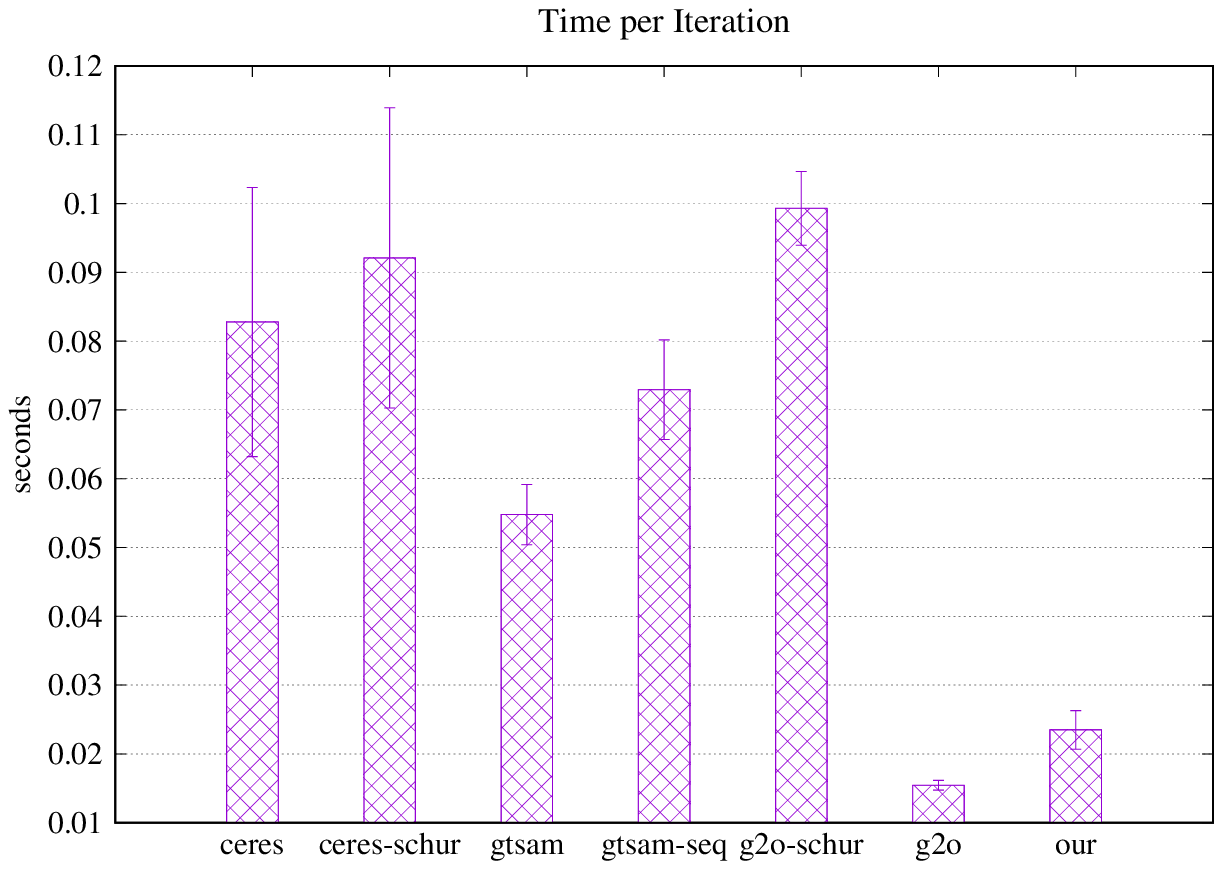}
    \subcaption{Iteration time: \texttt{victoria-park}.}
    \label{fig:it-time-victoria-park}
  \end{subfigure}
  \begin{subfigure}{0.489\linewidth}
    \centering
    \includegraphics[width=\columnwidth]{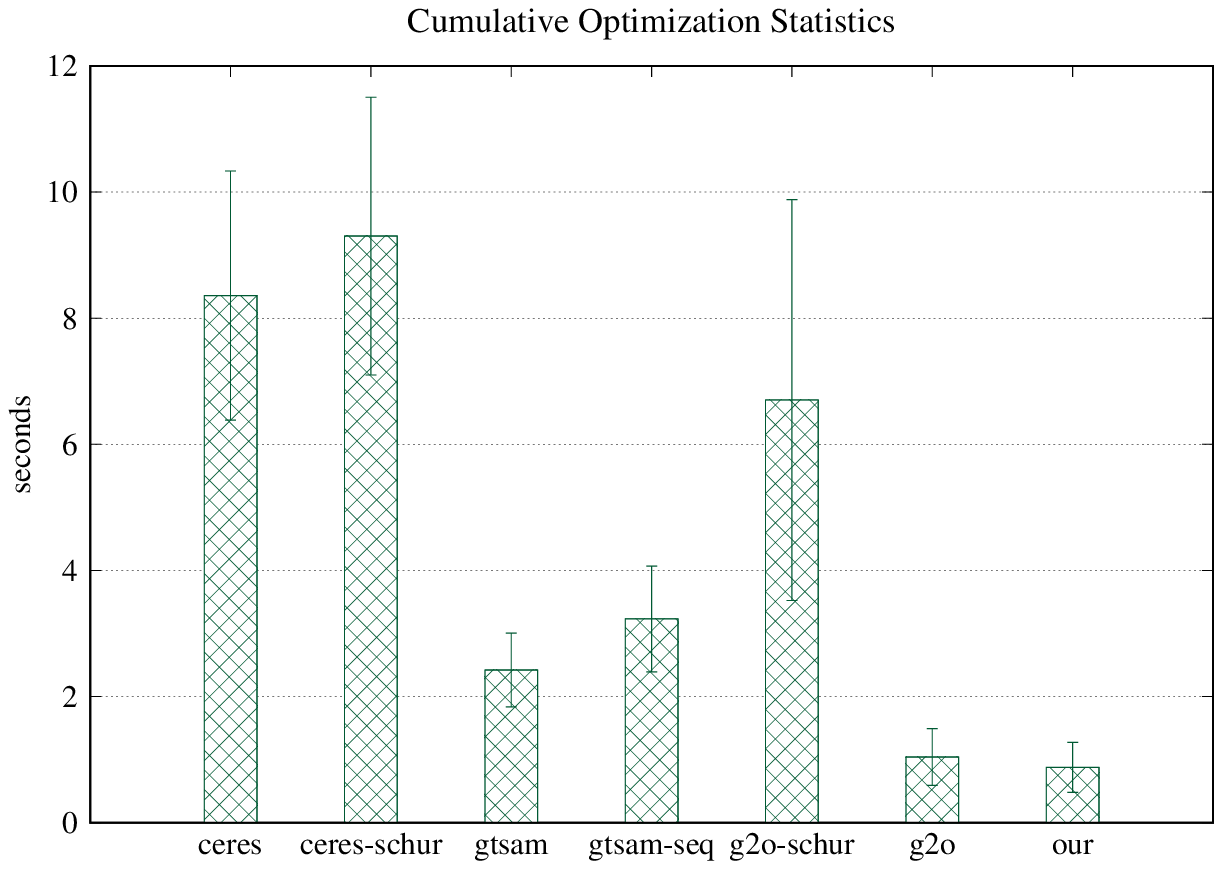}
    \subcaption{Cumulative time: \texttt{victoria-park}.}
    \label{fig:cum-time-victoria-park}
  \end{subfigure} \\ \vspace{10pt}
  \begin{subfigure}{0.489\linewidth}
    \centering
    \includegraphics[width=\columnwidth]{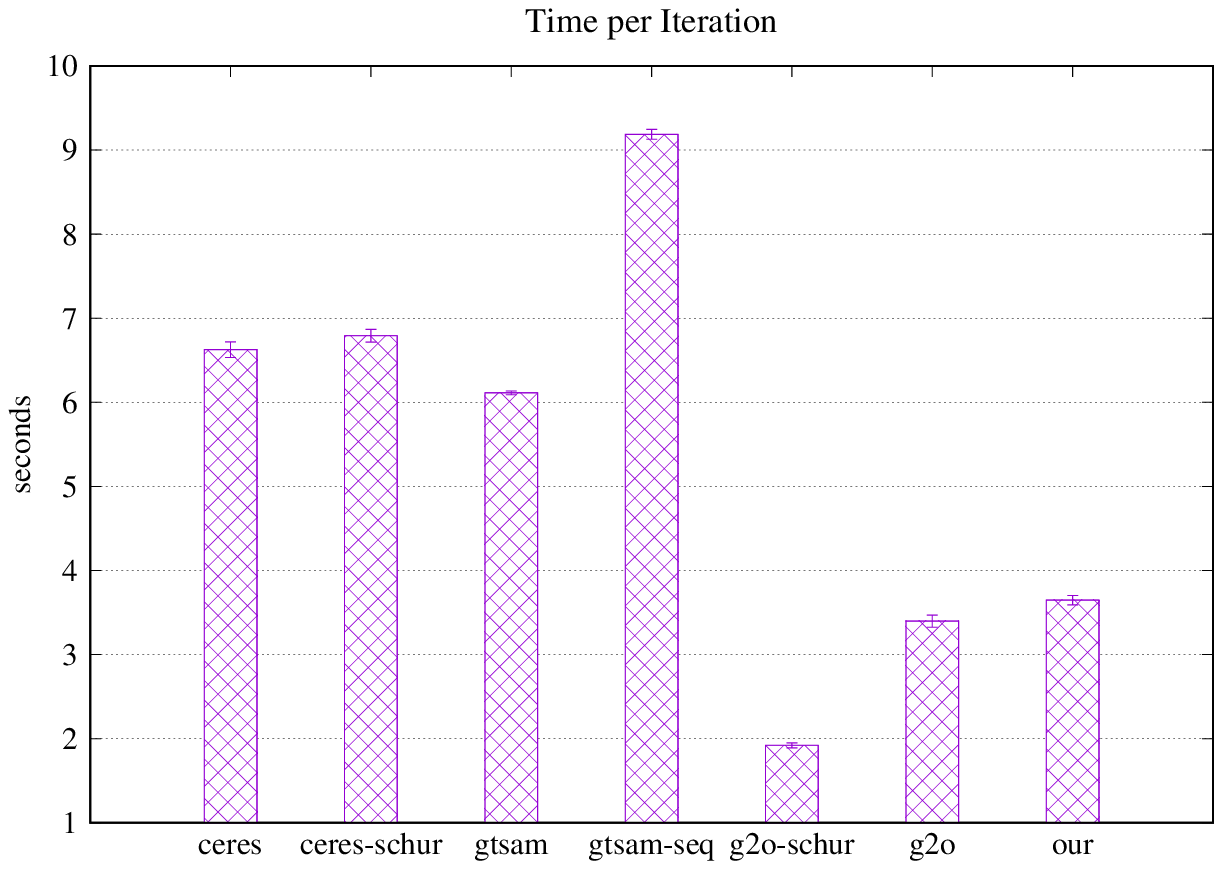}
    \subcaption{Iteration time: \texttt{kitti-00-full}.}
    \label{fig:it-time-kitti-00-full}
  \end{subfigure}
  \begin{subfigure}{0.489\linewidth}
    \centering
    \includegraphics[width=\columnwidth]{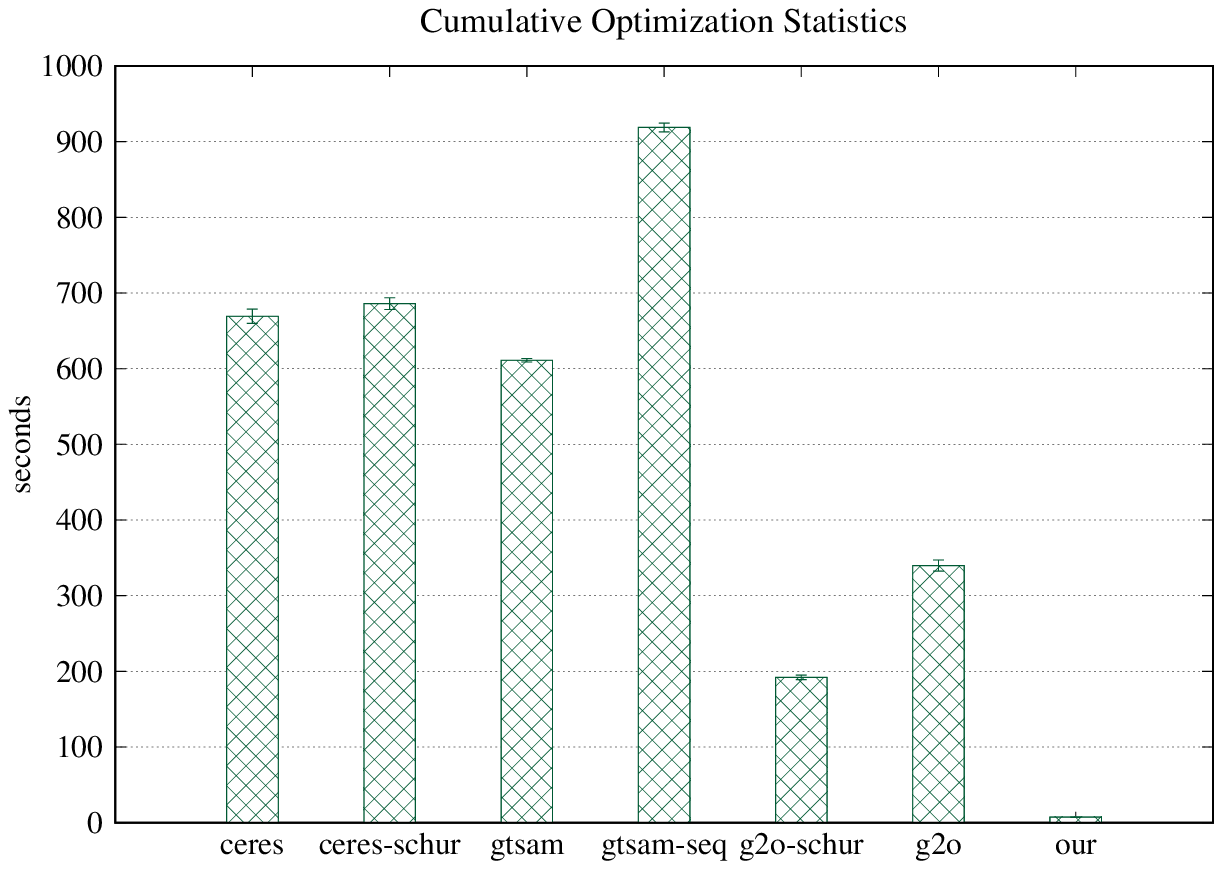}
    \subcaption{Cumulative time: \texttt{kitti-00-full}.}
    \label{fig:cum-time-kitti-00-full}
  \end{subfigure} 
  \caption{Timing analysis: the left column illustrates the time to perform a
    complete \gls{lm} iteration; the right column
    reports the total time to complete the optimization. All values are mean 
    and 
    standard deviation
    computed over 5 noise realizations.}
  \label{fig:plgo-timings}
\end{figure*}
Sparse problems are mostly represented by generic global optimization
scenarios, in which the graph has a large number of variables while each factor
connects a very small subset of those (typically two). In this kind of problems,
the graph remains unchanged during the iterations, therefore, the most
time-consuming part of the optimization is the solution of the
linear system not its construction. \gls{pgo} and \gls{plgo}
are two instances of this problem that are very common in the \gls{slam}
context and, therefore, we selected these two to perform comparative benchmarks.

\subsubsection{Pose-Graph Optimization}
\gls{pgo} represents the backbone of \gls{slam} systems and it has been well
investigated by the research community. For these experiments, we employed
standard 3D \gls{pgo} benchmark datasets - all publicly
available~\cite{aloise2019chordal}.
We added to the factors \gls{awgn} and we initialized the
graph using the breadth-first initialization.
\begin{table*}[!t]
  \centering
  \begin{tabular}{|c|c|c|c|c|}
    \hline
    \textsc{Dataset} &
    \textsc{Variables} &
    \textsc{Factors} &
    \textsc{Noise} $\bSigma_t \; [m]$&
    \textsc{Noise} $\bSigma_R \; [rad]$
    \\ \hline
    \texttt{kitti-00} &
    4541 &
    5595 &
    $\mathrm{diag}(0.05, 0.05, 0.05)$ &
    $\mathrm{diag}(0.01, 0.01, 0.01)$
    \\ \hline
    \texttt{sphere-b} &
    2500 &
    9799 &
    $\mathrm{diag}(0.10, 0.10, 0.10)$ &
    $\mathrm{diag}(0.05, 0.05, 0.05)$
    \\ \hline
    \texttt{torus-b} &
    1000 &
    1999 &
    $\mathrm{diag}(0.10, 0.10, 0.10)$ &
    $\mathrm{diag}(0.05, 0.05, 0.05)$
    \\ \hline
  \end{tabular}
  \caption{Specifications of \gls{pgo} datasets.}
  \label{tab:datasets}
\end{table*}
We report in \tabref{tab:datasets} the complete specifications of the datasets
employed together with the noise statistics used. Given the
probabilistic nature of the noise imposed on the factors, we performed
experiments over 10 noise realizations and we report here
the statistics of the results obtained - \ie~mean and standard deviation.
To avoid any bias in the comparison, we used the native \gls{lm}
implementation of each framework, since it was the only algorithm common to
all candidates. Furthermore, we imposed a maximum number of 100 \gls{lm}
iterations. Still, each framework has its own termination criterion active, so
that each one can detect when to stop the optimization. Finally, no robust
kernel has been employed in these experiments.

\begin{table*}[!t]
  \centering
  \begin{tabular}{cc|c|c|c|c|}
    \cline{3-6}
    &
    &
    \textsc{Ceres} &
    \textsc{\g2o} &
    \textsc{Gtsam} &
    \textsc{Our}
    \\ \cline{1-6}
    \multicolumn{1}{ |c  }{\multirow{2}{*}{\texttt{kitti-00}} } &
    \multicolumn{1}{ |c| }{$\mathrm{ATE}_{pos} [m]$} &
    $96.550 \pm 36.680$ &
    $94.370 \pm 39.590$ &
    $\mathbf{77.110 \pm 41.870}$ &
    $95.290 \pm 38.180$ \\ \cline{2-6}
    \multicolumn{1}{ |c  }{}                        &
    \multicolumn{1}{ |c| }{$\mathrm{ATE}_{rot} [rad]$} &
    $1.107 \pm 0.270$ &
    $0.726 \pm 0.220$ &
    $\mathbf{0.579 \pm 0.310}$ &
    $0.720 \pm 0.230$ \\ \cline{1-6}
    \multicolumn{1}{ |c  }{\multirow{2}{*}{\texttt{sphere-b}} } &
    \multicolumn{1}{ |c| }{$\mathrm{ATE}_{pos} [m]$} &
    $83.210 \pm 7.928$ &
    $\mathbf{9.775  \pm 4.003}$ &
    $55.890 \pm 12.180$ &
    $26.060 \pm 16.350$ \\ \cline{2-6}
    \multicolumn{1}{ |c  }{}                        &
    \multicolumn{1}{ |c| }{$\mathrm{ATE}_{rot} [rad]$} &
    $2.135 \pm 0.282$ &
    $\mathbf{0.150 \pm 0.160}$ &
    $0.861 \pm 0.170$ &
    $0.402 \pm 0.274$ \\ \cline{1-6}
    \multicolumn{1}{ |c  }{\multirow{2}{*}{\texttt{torus-b}} } &
    \multicolumn{1}{ |c| }{$\mathrm{ATE}_{pos} [m]$} &
    $14.130 \pm 1.727$ &
    $\mathbf{2.232  \pm 0.746}$ &
    $8.041  \pm 1.811$ &
    $3.691  \pm 1.128$ \\ \cline{2-6}
    \multicolumn{1}{ |c  }{}                        &
    \multicolumn{1}{ |c| }{$\mathrm{ATE}_{rot} [rad]$} &
    $2.209 \pm 0.3188$ &
    $\mathbf{0.121 \pm 0.0169}$ &
    $ 0.548  \pm 0.082 $ &
    $0.156 \pm 0.0305$ \\ \cline{1-6}
  \end{tabular}
  \caption{Comparison of the ATE (RMSE) of the optimization result - mean and
    standard deviation.}
  \label{tab:pgo-ate}
\end{table*}

\begin{table}[!t]
	\centering
	\begin{tabular}{|c|c|c|c|c|}
	\hline
		&
    \textsc{Ceres} &
\textsc{\g2o} &
\textsc{Gtsam} &
\textsc{Our}
		\\ \hline
		\texttt{kitti-00} &
		$81.70$ &
		$99.50$ &
		$69.40$ &
		$\mathbf{49.0}$
		\\ \hline
		\texttt{sphere-b} &
		$101.0$ &
		$70.90$ &
		$\mathbf{15.50}$ &
		$27.40$
		\\ \hline
		\texttt{torus-b} &
		$93.50$ &
		$12.90$ &
		$25.50$ &
		$\mathbf{16.40}$
		\\ \hline
	\end{tabular}
	\caption{Comparison of the number of \gls{lm} iterations to reach convergence 
	- mean values.}
	\label{tab:lm-pgo}
\end{table}

In~\tabref{tab:pgo-ate} we illustrate the \gls{ate} (RMSE)
computed on the optimized graph with respect to the ground truth. The values
reported refer to mean and standard deviation over all noise trials. As
expected, the result obtained are in line with all other methods.
\figref{fig:pgo-timings}, instead, reports a detailed timing analysis. The time
to perform a complete \gls{lm} iteration is always among the smallest, with a
very narrow standard deviation. Furthermore, since the specific implementation
of \gls{lm} is slightly different in each framework, we reported also the total
time to perform the full optimization, while the number of \gls{lm} iteration 
elapsed are shown in \tabref{tab:lm-pgo}.
Also in this case, our system is able to achieve state-of-the-art performances
that are better or equal to the other approaches.

\subsubsection{Pose-Landmark Graph Optimization}
\gls{plgo} is another common global optimization task in \gls{slam}. In this
case, the variables contain both robot (or camera) poses and landmarks'
position in the world. Factors, instead, embody spatial constraints between
either two poses or between a pose and a landmark. As a result, this kind of
factor graphs are the perfect representative of the \gls{slam} problem, since
they contain the robot trajectory \textit{and} the map of the environment.
\begin{table*}[!t]
  \centering
  \begin{tabular}{|c|c|c|c|c|c|}
    \hline
    \textsc{Dataset} &
    \textsc{Variables} &
    \textsc{Factors} &
    \textsc{Noise} $\bSigma_t \; [m]$ &
    \textsc{Noise} $\bSigma_R \; [rad]$ &
    \textsc{Noise} $\bSigma_{land} \; [m]$ \\ \hline
    \texttt{victoria-park}&
    7120 &
    10608 &
    $\mathrm{diag}(0.05, 0.05)$ &
    $0.01$ &
    $\mathrm{diag}(0.05, 0.05)$ \\ \hline
    \texttt{kitti-00-full}&
    123215 &
    911819 &
    $\mathrm{diag}(0.05, 0.05, 0.05)$ &
    $\mathrm{diag}(0.01, 0.01, 0.01)$ &
    $\mathrm{diag}(0.05, 0.05, 0.05)$ \\ \hline
  \end{tabular}
  \caption{Specification of \gls{plgo} datasets.}
  \label{tab:plgo-datasets}
\end{table*}
To perform the benchmarks we used two datasets: Victoria
Park~\cite{guivant2001optimization} and KITTI-00~\cite{geiger2012kitti}.
We obtained the last one running ProSLAM~\cite{schlegel2018proslam} on the
stereo data and saving the full output graph. We super-imposed to the factors
specific \gls{awgn} and we generated the initial guess through the
breadth-first initialization technique. \tabref{tab:plgo-datasets} summarizes
the specification of the datasets used in these experiments. Also in this case,
we sampled multiple noise trials (5 samples) and reported mean and standard
deviation of the results obtained. The configuration of the framework is the
same as the one used in \gls{pgo} experiments - \ie~100 \gls{lm} iterations at
most, with termination criterion active.

\begin{table*}[!t]
  \centering
  \begin{tabular}{cc|c|c|c|c|}
    \cline{3-6}
    &
    &
    \textsc{Ceres} &
    \textsc{\g2o} &
    \textsc{Gtsam} &
    \textsc{Our}
    \\ \cline{1-6}
    \multicolumn{1}{ |c  }{\multirow{2}{*}{\texttt{victoria-park}} } &
    \multicolumn{1}{ |c| }{$\mathrm{ATE}_{pos} [m]$} &
    $37.480 \pm 21.950$ &
    $29.160 \pm 37.070$ &
    $\mathbf{2.268 \pm 0.938}$ &
    $5.459 \pm 3.355$ \\ \cline{2-6}
    \multicolumn{1}{ |c  }{}                        &
    \multicolumn{1}{ |c| }{$\mathrm{ATE}_{rot} [rad]$} &
    $0.515 \pm 0.207$ &
    $0.401 \pm 0.461$ &
    $\mathbf{0.030 \pm 0.007}$ &
    $0.056 \pm 0.028$ \\ \cline{1-6}
    \multicolumn{1}{ |c  }{\multirow{2}{*}{\texttt{kitti-00-full}} } &
    \multicolumn{1}{ |c| }{$\mathrm{ATE}_{pos} [m]$} &
    $134.9 \pm 29.160$ &
    $31.14 \pm 27.730$ &
    $\mathbf{30.97 \pm 18.150}$ &
    $135.4 \pm 27.000$ \\ \cline{2-6}
    \multicolumn{1}{ |c  }{}                        &
    \multicolumn{1}{ |c| }{$\mathrm{ATE}_{rot} [rad]$} &
    $1.137 \pm 0.268$ &
    $\mathbf{0.173 \pm 0.157}$ &
    $0.174 \pm 0.104$ &
    $0.850 \pm 0.148$ \\ \cline{1-6}
  \end{tabular}
  \caption{Comparison of the ATE (RMSE) of the optimization result - mean and
    standard deviation.}
  \label{tab:plgo-ate}
\end{table*}
\begin{table*}[!t]
	\centering
	\begin{tabular}{|c|c|c|c|c|c|c|c|}
		\hline
		&
		\textsc{Ceres} &
		\textsc{Ceres-schur} &
		\textsc{\g2o} &
		\textsc{\g2o-schur} &
		\textsc{Gtsam} &
		\textsc{Gtsam-seq} &
		\textsc{Our}
		\\ \hline
		\texttt{victoria-pack} &
		$101.0$ &
		$101.0$ &
		$66.8$ &
		$66.0$ &
		$43.6$ &
		$43.6$ &
		$\mathbf{36.0}$ 
		\\ \hline
		\texttt{kitti-00-full} &
		$101.0$ &
		$101.0$ &
		$100.0$ &
		$100.0$ &
		$100.0$ &
		$100.0$ & 
		$\mathbf{2.0}$
		\\ \hline
	\end{tabular}
	\caption{Comparison of the number of \gls{lm} iterations to reach convergence 
	- mean values.}
	\label{tab:lm-pglo}
\end{table*}

As reported in~\tabref{tab:plgo-ate} the \gls{ate} (RMSE) that we obtain is
compatible with the one of the other frameworks. The higher error in the
\texttt{kitti-00-full} dataset is mainly due to the slow convergence of
\gls{lm}, that triggers too early the termination criterion, as shown in 
\tabref{tab:lm-pglo}. In such case, the
use of~\gls{gn} leads to better results, however, in order to not bias the
evaluation, we choose to not report results obtained with different \gls{ils}
algorithms.
As for the wall times to perform the optimization, the results are illustrated
in~\figref{fig:plgo-timings}. In \gls{plgo} scenarios, given the fact that
there are two types of factors, the linear system
in~\eqref{eq:nonlinear-solution} is can be rearranged as follows:
\begin{equation}
  \begin{pmatrix}
    \bH_{pp} & \bH_{pl} \\ \bH_{pl}^\top & \bH_{ll}
  \end{pmatrix}
  \begin{pmatrix}
    \bDeltax_{p} \\ \bDeltax_{l}
  \end{pmatrix} =
  \begin{pmatrix}
    -\bb_{p} \\ -\bb_{l}
  \end{pmatrix}.
  \label{eq:schur-system}
\end{equation}
A linear system with this structure can be solved more efficiently through the
Schur complement of the Hessian matrix~\cite{frese2005proof}, namely:
\begin{align}
  (\bH_{pp} - \bH_{pl}\bH_{ll}^{-1}\bH_{pl}^\top) \bDeltax_{p} &= -\bb_{p} +
  \bH_{pl} \bH_{ll}^{-1}\bb_{l} \label{eq:schur-poses} \\
  \bH_{ll} \bDeltax_{l} &= -\bb_{l} + \bH_{pl}^\top \bDeltax_{p}
  \label{eq:schur-landmarks}.
\end{align}
Ceres-Solver and \g2o can make use of the Schur complement to solve
this kind of special problem, therefore, we reported also the wall
times of the optimization when this technique is used. Obviously,
using the Schur complement leads to a major improvement in the
efficiency of the linear solver, leading to very low iteration
times. For completeness, we reported the results of GTSAM with two
different linear solvers: \texttt{cholesky\_multifrontal} and
\texttt{cholesky\_sequential}.  Our framework does not provide at the
moment any implementation of a Schur-complement-based linear solver,
still, the performance achieved are in line with all the non-Schur
methods, confirming our conjectures.

\section{Conclusions}\label{sec:conclusions}
In this work, we propose a generic overview on \gls{ils} optimization
for factor graphs in the fields of robotics and computer vision. Our
primary contribute is providing a unified and complete methodology to
design efficient solution to generic problems. This paper analyzes in
a probabilistic flavor the \emph{mathematical fundamentals} of
\gls{ils}, addressing also many important collateral aspects of the
problem such as dealing with non-Euclidean spaces and with outliers,
exploiting the sparsity or the density.  Then, we propose a set of
common use-cases that exploit the theoretic reasoning previously done.

In the second half of the work, we investigate how to \emph{design} an
efficient system that is able to work in all the possible scenarios
depicted before.  This analysis led us to the \emph{development} of a
novel \gls{ils} solver, focused on efficiency, flexibility and
compactness.  The system is developed in modern C++ and almost
entirely self-contained in less than 6000 lines of code.  Our system
can seamlessly deal with sparse/dense, static/dynamic problems with a
unified consistent interface. Furthermore, thanks to specific
implementation designs, it allows to easily prototype new factors and
variables or to intervene at low level when performances are critical.
Finally, we provide an extensive evaluation of the system's
performances, both in dense - \eg~\gls{icp} - and sparse - \eg~batch
global optimization - scenarios.  The evaluation shows that the
performances achieved are in line with contemporary state-of-the-art
frameworks, both in terms of accuracy and speed.

\appendices
\section{$\se3$ Mappings} \label{sec:appendix-se3}
In this section, we will assume that a
$\mathrm{SE}(3)$ variable $\bX$ is composed as follows:
\begin{equation}
  \bX \in \mathrm{SE}(3) = 
  \begin{bmatrix} 
  \bR & \bt\\\bZero_{1\times3} & 1
  \end{bmatrix}.
  \label{eq:isometry}
\end{equation}
A possible minimal representation for this object could be using 3 Cartesian 
coordinates for the position and the 3 Euler angles for the orientation, namely
\begin{equation}
  \bDelta x = 
  \begin{bmatrix}
  \Delta x &\Delta y &\Delta z &\Delta \phi &\Delta \gamma & \Delta \psi
  \end{bmatrix}^\top \eq \begin{bmatrix} \bDelta \bt & \bDelta 
  \theta\end{bmatrix}^\top
  \label{eq:minimal-euler}
\end{equation}
To pass from one representation to the other, we should define 
suitable mapping functions. In this case, we use the following notation:
\begin{align}
  \bDelta \bX &= \v2t(\bDeltax) \label{eq:v2t-euler} \\
  \bDeltax &= \mathrm{t2v}(\bDelta \bX) \label{eq:t2v-euler}.
\end{align}
More in detail, the function $\v2t$ computes the $\se3$ isometry reported 
in~\eqref{eq:isometry} from the 6-dimensional vector $\bDeltax$ 
in~\eqref{eq:minimal-euler}. While the translational component of $\bX$ can be 
recovered straightforwardly from $\bDeltax$, the rotational part requires to 
compose the rotation matrices around each axis, leading to the following 
relation:
\begin{equation}
  \bR(\bDelta \theta) = \bR(\Delta \phi,\Delta \gamma,\Delta \psi) = 
  \bR_x(\Delta 
  \phi) \; \bR_y(\Delta \gamma) \; \bR_z(\Delta \psi).
  \label{eq:composing-rotation}
\end{equation}
In~\eqref{eq:composing-rotation}, $\bR_x, \bR_y, \bR_z$ represent elementary 
rotations around the $x,y$ and $z$ axis. Summarizing, we can 
expand~\eqref{eq:v2t-euler} as:
\begin{equation*}
  \bDelta \bX \eq \v2t(\bDelta x) \eq \begin{bmatrix}
  \bR(\bDelta \theta) & \bDelta t \\
  \bZero_{1\times3} & 1
  \end{bmatrix}
\end{equation*}
Expanding~\eqref{eq:composing-rotation} and performing all the multiplications, 
the rotation matrix $\bR(\bDelta \theta)$ is computed as follows:
\begin{align}
  &\bR(\bDelta \theta) 
  \eq
  \begin{bmatrix}
  r_{11} & r_{12} & r_{13} \\
  r_{21} & r_{22} & r_{23} \\
  r_{31} & r_{32} & r_{33} \\
  \end{bmatrix} \nonumber \\
  &\eq
  \begin{bmatrix}
    c(\Delta \gamma)\,c(\Delta \psi) & 
    -c(\Delta \gamma)\,s(\Delta \psi) & 
    s(\Delta \gamma) \\
    a & 
    b & 
    -c(\Delta \gamma)\,s(\Delta \phi) \\
    c & 
    d & 
    c(\Delta \gamma)\,c(\Delta \phi)
  \end{bmatrix}
  \label{eq:rot-composition-expanded}
\end{align}
where $c(\cdot)$ and $s(\cdot)$ indicate the cosine and sine of an angle 
respectively, while
\begin{align*}
  a &= c(\Delta\phi) \, s(\Delta\psi) + s(\Delta\phi) \, c(\Delta\psi) \, 
  s(\Delta \gamma) \\
  b &= c(\Delta \phi) \, c(\Delta \psi) - s(\Delta \phi) \, s(\Delta \gamma) \, 
  s(\Delta \psi) \\
  c &= s(\Delta \phi) \, s(\Delta \psi) - c(\Delta \phi) \, c(\Delta \psi) \, 
  s(\Delta \gamma) \\
  d &= s(\Delta \phi) \, c(\Delta \psi) + c(\Delta \phi) \, s(\Delta \gamma) \, 
  s(\Delta \psi).
\end{align*}

On the contrary, through~\eqref{eq:t2v-euler} we compute the minimal 
parametrization $\bDeltax$ starting from $\bDelta \bX$.
Again, while the translational component of $\bDeltax$ can be retrieved easily 
from the isometry. The rotational component $\bDelta \theta$ - \ie~the 3 Euler 
angles - should be computed starting from the rotation matrix 
in~\eqref{eq:rot-composition-expanded}, in formul\ae:
\begin{align*}
  \Delta \phi &= \mathrm{atan2}\left(\frac{-r_{23}}{r_{33}}\right) \\
  \Delta \psi &= \mathrm{atan2}\left(\frac{-r_{12}}{r_{11}}\right) \\
  \Delta \gamma &= \mathrm{atan2}\left(\frac{r_{13}}{r_{11} \cdot 
    \frac{1}{c(\Delta \psi)}}\right).
\end{align*}

Other minimal parametrizations of $\se3$ can be used, and they
typically differ on how they represent the rotational component of the
$\se3$ object. Common alternatives to euler angles are unit
quaternions and axis-angle.  Clearly, changing the minimal
parametrization will affect the jacobians too, and thus the entire
optimization process.


\section{ICP Jacobian} \label{sec:appendix-icp}
In this section, we will provide the reader the full mathematical derivation of 
the Jacobian matrices reported in~\secref{sec:icp-factor}. To this end, we 
recall that the measurement function for the \gls{icp} problem is:
\begin{equation}
  \bh^\mathrm{icp}_k(\bX) \triangleq 
  \bX^{-1} \bp = \bR^\top (\bp - \bt).
  \label{eq:icp-prediction}
\end{equation}
If we apply a small state perturbation using the $\boxplus$ operator defined 
in~\eqref{eq-icp-boxplus-se3}, we obtain:
\begin{align}
  \bh^\mathrm{icp}(\bX \boxplus \bDelta x) &\eq (\v2t(\bDelta \bx) \cdot 
  \bX)^{-1} \bp \nonumber \\
  &\eq \bX^{-1} \cdot \v2t(\bDelta \bx)^{-1} \bp \nonumber \\
  &\eq \bR^\top 
  (\v2t(\bDelta \bx)^{-1}\bp - \bt).
  \label{eq:icp-perturbed-error}
\end{align}
To compute the Jacobian matrix $\bJ^\mathrm{icp}$ we should 
derive~\eqref{eq:icp-perturbed-error} with respect to $\bDeltax$.
Note that, the translation vector $\bt$ is constant with 
respect to the perturbation, so it will have no impact in the computation.
Furthermore, $\bR^\top$ represents a constant multiplicative factor. 
%
Since the state perturbation is local, its magnitude is small enough to 
make the following approximation hold:
\begin{equation}
  \bR(\bDelta \theta) \approx 
  \begin{bmatrix}
    1 & -\bDelta \psi & \bDelta \gamma\\
    \bDelta \psi & 1 & -\bDelta \phi\\
    -\bDelta \gamma & \bDelta \phi & 1 
  \end{bmatrix}.
\end{equation}
Finally, the Jacobian $\bJ^\mathrm{icp}$ is computed as follows:
\begin{align}
  &\bJ^\mathrm{icp}(\bX) =
  \frac{\partial \left(\bh^\mathrm{icp}(\bX \boxplus \bDelta \bx)\right)}
    {\partial \bDeltax} \Big\rvert_{\bDeltax = \bZero} \nonumber \\
  &\eq 
  \bR^\top\frac{\partial \left(\v2t(\bDelta \bx)^{-1}\bp\right)}
    {\partial \bDeltax} \Big\rvert_{\bDeltax = \bZero} \eq \nonumber \\
  &=
  -\bR^{\top}
  \begin{bmatrix}
    \frac{\partial \left(\v2t(\bDelta \bx)^{-1} \bp\right)}{\partial \bDelta 
      \bt}\Big\rvert_{\bDelta \bx = \bZero} &
    \frac{\partial \left(\v2t(\bDelta \bx)^{-1} \bp\right)}{\partial \bDelta 
      \theta}\Big\rvert_{\bDelta\bx = \bZero}
  \end{bmatrix} \nonumber \\
  &\eq
  -\bR^{\top} 
  \begin{bmatrix}
    \mathbf{I}_{3\times 3} & -\left[\bp\right]_{\times}
  \end{bmatrix}
  \label{eq:icp-complete-jacobian}
\end{align}
where $\left[\bp\right]_{\times}$ is the skew-symmetric matrix constructed out 
of $\bp$.
%

\section{Projective Registration Jacobian} \label{sec:appendix-pr}
In this section, we will provide the complete derivation of the Jacobians in 
the context of projective registration. From~\eqref{eq:pr-prediction}, we know 
that the prediction is computed as
\begin{equation*}
  \bh^\mathrm{reg}_k(\bX) = 
  \mathrm{hom}(\bK [\bh^{icp}(\bp^\mathrm{m}_{\mathrm{j}(k)})]).
\end{equation*}
Furthermore, as stated in~\eqref{eq:pr-jacobians-chain-rule}, we can compute 
the Jacobian for this factor through the chain rule, leading to the following 
relation:
\begin{align}
  \bJ^\mathrm{reg}(\bX) &=
    \frac
    {\partial \mathrm{hom}(\bv)}
    {\partial \bv}
    \Bigg\rvert_{\bv = \bp^\mathrm{cam}} \;\bK\; \bJ^{\mathrm{icp}} (\bX) 
    \nonumber \\
  &= \bJ^\mathrm{hom} (\bp^\mathrm{cam})\, \bK\, \bJ^{\mathrm{icp}} (\bX).
  \label{eq:pr-jacobian}
\end{align}
where $\bp^\mathrm{cam} = \bK \bX^{-1} \bp^\mathrm{m}$.
Since we already computed $\bJ^\mathrm{icp}$ 
in~\eqref{eq:icp-complete-jacobian}, in the remaining we will focus only on 
$\bJ^\mathrm{hom}$ - \ie~the contribution of the homogeneous division.
Given that the function $\hom(\cdot)$ is defined as
\begin{equation*}
  \hom([x \; y \; z]^\top) = 
  \begin{bmatrix}
    x / z \\  
    y / z
  \end{bmatrix}
\end{equation*}
the Jacobian $\bJ^\mathrm{hom}(\bp^{cam})$ is computed as follows:
\begin{equation}
  \bJ^\mathrm{hom} (\bp^\mathrm{cam}) \eq \begin{bmatrix}
  \frac{1}{p^{cam}_z} & 0 & \frac{-p^{\mathrm{cam}}_x}{ (p^{\mathrm{cam}}_z)^2} 
  \\
  0 & \frac{1}{p^{cam}_z} & \frac{-p^{\mathrm{cam}}_y}{ (p^{\mathrm{cam}}_z)^2}.
  \end{bmatrix}
  \label{eq:pr-complete-jacobian}
\end{equation}

\section{Bundle Adjustment Jacobian} \label{sec:appendix-ba}
In this section we address the computation of the Jacobians in the context of 
Bundle Adjustment. The scenario is the one described in~\secref{sec:sfm-ba}.
We recall that, in this case, each factor involves two state variables, namely 
a pose and a landmark. Therefore, $\bJ^{\mathrm{ba}}$ has the following pattern:
\begin{equation*}
  \bJ^\mathrm{ba}_k = 
  \begin{pmatrix}  
  \bZero & \cdot & \bZero &
  \bJ^\mathrm{ba}_{k,\mathrm{n}(k)} & 
  \bZero & \cdot & \bZero &
  \bJ^\mathrm{ba}_{k,\mathrm{m}(k)} & 
  \bZero & \cdot & \bZero
  \end{pmatrix}. 
\end{equation*}
More in detail, the two non-zero block embody the derivatives computed with 
respect to the two active variables, namely:
\begin{align*}
  \bJ^\mathrm{ba}_{pose} &= 
    \frac{\partial \left( \be_k(\bX^{cam} \boxplus \bDeltax^{cam}, 
      \bx^{land}) \right)}
      {\partial \bDeltax^{cam}}
  \\
  \bJ^\mathrm{ba}_{land} &= 
   \frac{\partial \left( \be_k(\bX^{cam}, \bx^{land}\boxplus \bDeltax^{land}) 
   \right)}
   {\partial \bDeltax^{land}}
\end{align*}
where $\be_k$ represents the error for the $k$-th factor, computed according 
to~\eqref{eq:ba-error}.

Unrolling the multiplications, we note that $\bJ^\mathrm{ba}_{pose}$ is the 
same as the one computed in~\eqref{eq:pr-complete-jacobian} - \ie~in the 
projective registration example.
The derivatives relative to the landmark - \ie~$\bJ^\mathrm{ba}_{pose}$ - can 
be straightforwardly computed from~\eqref{eq:icp-perturbed-error}, considering 
that the derivation is with respect to the landmark perturbation this time. In 
formlu\ae:
\begin{equation}
  \bJ^{\mathrm{ba}}_{land}(\bX) =
  \bJ^\mathrm{hom} (\bp^\mathrm{cam}) \bK \bR^\top.
  \label{eq:ba-jacobian-landmark}
\end{equation}
Summarizing, the complete Bundle Adjustment Jacobian is computed as:
\begin{align}
  \bJ^\mathrm{ba}_k &= 
    \begin{pmatrix}  
      \bZero & \cdot & \bZero &
      \bJ^{\mathrm{ba}}_{pose} & 
      \bZero & \cdot & \bZero &
      \bJ^{\mathrm{ba}}_{land} & 
      \bZero & \cdot & \bZero
    \end{pmatrix} \nonumber \\
    \bJ^{\mathrm{ba}}_{pose} &= 
      \bJ^\mathrm{hom} (\bp^\mathrm{cam}) \, \bK \, \bJ^{\mathrm{icp}} 
    \hspace{14pt}
    \bJ^{\mathrm{ba}}_{land} = 
      \bJ^\mathrm{hom} (\bp^\mathrm{cam}) \bK \bR^\top.
  \label{eq:ba-complete-jacobian}
\end{align}
%
%
%

\bibliographystyle{unsrt}

\balance

\end{document}